\documentclass[12pt]{article}

\usepackage{amssymb,amsmath}
\usepackage[T1]{fontenc}
\usepackage{algpseudocode,algorithm}
\usepackage{graphicx}

\newcommand{\Z}{\mathbf{0}}
\newcommand{\citep}{\cite}
\newcommand{\Vol}{\mathcal{V}}
\newcommand{\tra}{\mathrm{tr}}
\newcommand{\figaddr}[1]{#1}

\newcommand{\Bem}[1]{}

\newcommand{\naszLink}{\mbox{\url{https://github.com/ipipan-barstar/ASC.RSfEoSGC}}}

\usepackage{placeins}
\usepackage{url}

\begin{document}

\title{Rough Sets for {Explain}ability of Spectral Graph Clustering}

\newcommand{\orcidID}[1]{$^{#1}$ }
\author{
 Bartłomiej Starosta \orcidID{0000-0002-5554-4596}  
 \and
 Sławomir T. Wierzchoń \orcidID{0000-0001-8860-392X} 
\and 
Piotr Borkowski \orcidID{0000-0001-9188-5147}         \and  Dariusz Czerski \orcidID{0000-0002-3013-3483} 
\and Marcin Sydow $^*$
        \orcidID{0000-0001-6346-6324}
\and Eryk Laskowski 
        \orcidID{0000-0001-6346-6324}
        \and Mieczysław A. Kłopotek \orcidID{0000-0003-4685-7045}
        \\ 
        Institute of Computer Science of Polish Academy of Sciences\\ 
  ul. Jana Kazimierza 5, 01-248 Warszawa, Poland\\
  * Polish-Japanese Academy of Information Technology,\\ Koszykowa 86, 02-008 Warszawa, Poland
        }

\maketitle

\begin{abstract}
Graph Spectral Clustering methods (GSC)  allow representing clusters of diverse shapes, densities, etc. However, the results of such algorithms, when applied e.g. to text documents, are hard to explain to the user, especially due to embedding in the spectral
space which has no obvious relation to document contents. Furthermore, the presence of documents
without clear content meaning and the stochastic nature of the clustering algorithms deteriorate
explainability. 
This paper proposes an enhancement to the explanation methodology, proposed in an
earlier research of our team. 
It allows us to overcome the latter problems by taking inspiration from rough set theory.  
\end{abstract}


\section{Introduction}\label{sec:intro}

Explainability turns out to be crucial for the application of Machine Learning techniques, in particular clustering. From this perspective, Graph Spectral Clustering (GSC for short) methods have some serious drawbacks as they operate in a space hard to understand by humans, that is, in the Laplacian spectral domain, and this is why they are considered opaque models\footnote{
See also, for example \cite{Sabbatini:2022}, and  
\url{https://crunchingthedata.com/when-to-use-spectral-clustering/}.
}. 
{Model-unaware methods of textual clustering are known and could be considered an option, but they only describe groups and do not explain why documents are included. }
Recently explanatory clustering methods for Combinatorial and Normalized Laplacian-based GSC were proposed for the textual domain, as presented in {\citep{Plosone2025}}. Yet, two problems remain open: (1) in real-world settings, there exist objects that do not fit any cluster and are forced into some clusters just distorting them, and (2) clustering methods do not achieve the real optimal value of the quality function so that some cluster elements may also not fit the concepts and can distort them. In this paper, we want to contribute to solving both problems by applying concepts from rough set theory, where entities, e.g. documents, are divided into core and boundary objects.

The first problem is addressed by separating the data points into core and boundary elements. The core elements are only considered suitable for cluster explanation. The second issue can be addressed by applying various clustering techniques. We can achieve a more accurate clustering outcome by considering the intersection of results as the core part of the clustering and treating the remaining points as boundary points. For the first issue, we propose a similarity-based criterion to be applied to the data points. The solution to the second problem is only feasible if the alternative clustering methods are convergent, in the sense that they yield the same solution under a favorable prevalence of core points. We show empirically that both goals can be achieved using the same discernment-based approach.

In practice, however, this ideal is rarely achieved with real-world algorithms and/or real-world data, due to e.g. NP-hardness of the problem, sticking in a local optimum by the clustering algorithm, measurement errors/biases, etc.   {Therefore, in practice, various relaxations are applied, resulting in imprecise or questionable cluster membership. Although some approaches (such as fuzzy sets) focus on assigning some degree of cluster membership, this is not a good idea in practical applications. For instance, in business applications, clustering (or ``segmentation'') is performed to address e.g. some promotion to a segment/cluster of customers, proposing some incentives, discounts, etc.  The decision has to be discrete; there is no room to send a promotion ``to some extent''. So, the cluster assignment needs to be sharp. However, as mentioned earlier, there are cases where determining the intrinsic cluster membership of certain items with absolute certainty is challenging.
}

Thus, the approach introduced in the context of rough sets by \cite{Paw82} may prove fruitful here as it allows one to distinguish between cluster core elements and boundary elements.
This distinction can, for example, result from repeated runs of a particular probabilistic clustering algorithm. Elements that belong to the same cluster on each run are (with high probability) the core of a cluster, while the others can be considered to belong to the boundaries of different clusters.  
Another situation, though bearing some similarity, is when we apply different algorithms optimizing the same criterion, but each in an approximate way. The elements fitting together may be considered the core, with a high probability of fitting the (approximated) criterion, while the other will constitute boundaries of appropriate clusters.  However, another approach can be a robustness study, checking the impact on the optimality criterion of randomly shuffling parts of the established clusters. 

In each of these cases, the clustering methodology faces the currently popular challenge of explainability. Although the explainability of clustering results is a challenge in itself, the explainability of rough clustering results brings new expectations to the explanation process.  

The distinction between the core elements of a cluster and the boundary may be crucial for business. Orienting advertisements towards the cluster core might be significantly more efficient than focusing on the boundary. To determine this, it may be very important to explain the boundary contents in comparison to the core contents.

In this paper, we will address the problem of {explain}ability under clustering with different algorithms that optimize (in an approximate way) the same criterion. 

The structure of the paper is as follows.  
Section \ref{sec:prevworkGSA}  briefly introduces Graph Spectral Analysis (GSA).  Section \ref{sec:challenges} is dedicated to the challenges of real data analysis. Here, a short review of the literature on clustering with rough sets tools is given and our approach to this problem is explained. Section \ref{sec:experiments} reports numerical experiments and Section \ref{sec:conclusions} concludes the paper.

\section{Brief introduction to Graph Spectral Analysis}\label{sec:prevworkGSA}

The so-called Graph Spectral Analysis, encompassing Graph Spectral Clustering (GSC) and Graph Spectral Classification (GSL), stands for a more recent way of looking into relationships between data objects that are characterized by mutual similarity measures and hence can be best described by a graph with weights equal to those similarities. 

There are several graph clustering methods; see, for example, \citep{Schaeffer}, aimed at discovering clusters that meet related but slightly different criteria. The best known are the RCut criterion and the NCut criterion.

Assume that the matrix $S$, called the similarity matrix, represents similarities between pairs of $n$ objects with entries ranging from 0 to 1. $S$ is assumed to have a diagonal equal to zero, as self-loops are not allowed (see, e.g. \citep{TU:2022:3673}). 

The RCut clustering aims at splitting the dataset into $k$ clusters $\Gamma=\{C_1,\dots, C_k\}$ minimizing the following criterion:
\begin{equation}\label{eq:qRCut_def}
  Q^{[RCut]}(\Gamma) 
  =\sum_{j=1}^k \frac{1}{n_j}  \sum_{i\in C_j} \sum_{\ell\not\in C_j}  s_{i\ell}
\end{equation}
where $n_j=|C_j|$. A detailed discussion leading to the above formulation can be found, for example, in \citep{Plosone2025}.

The goal of NCut clustering is to split the data set into $k$ clusters $\Gamma=\{C_1,\dots, C_k\}$ while minimizing the following criterion:
\begin{equation}\label{eq:qNCut_def}
  Q^{[NCut]}(\Gamma) 
  =\sum_{j=1}^k \frac{1}{\Vol_j}  \sum_{i\in C_j} \sum_{\ell\not\in C_j}  s_{i\ell}
\end{equation}
where $\Vol_j=\sum_{i \in C_j} d_{ii}$ denotes the volume of the node~$j$, and  $d_{ii}=\sum_{j=1}^n s_{ij}$ stands for the degree of the node~$i$.  {Thus, we try to minimize the total weights of links joining nodes belonging to opposite clusters.  In the first case, this sum is divided by the cardinalities of the respective clusters, while in the second case - by the volumes of these clusters.
} Again, see \citep{Plosone2025} for a more in-depth discussion.

The NRCut clustering can be considered a compromise between RCut and NCut. It 
aims at splitting the dataset into $k$ clusters $\Gamma=\{C_1,\dots, C_k\}$   minimizing the following criterion:
\begin{equation}\label{eq:qNRCut_def}
  Q^{[NRCut]}(\Gamma) 
  =\sum_{j=1}^k \frac{1}{n_j+\Vol_j}  \sum_{i\in C_j} \sum_{\ell\not\in C_j}  s_{i\ell}
\end{equation}
with notation as above.  

Although finding clusters that meet the above criteria is NP-hard, GSC methods are designed to be a relaxation of cut-based graph clustering methods.

Whichever of these clustering methods is considered, one can reformulate them to minimize the square form of the product of a Laplacian and the so-called indicator vectors. An indicator vector is a vector of cluster membership of items represented by the nodes. Each cluster is characterized by an indicator vector. Non-zero entries of this vector indicate membership and zero entries indicate non-membership in a given cluster.  

Let us recall the definitions of selected Laplacians, suitable for handling the mentioned graph-cut tasks. 

A \emph{combinatorial Laplacian} $L$ corresponding to the matrix $S$ is defined as 
\begin{equation}\label{eq:combLapDef} L=D-S, \end{equation}
where $D$ is the degree matrix, i.e., a diagonal matrix whose elements are the degrees $d_{jj}$ of all nodes.
A \emph{normalized Laplacian}%
\footnote{
Other Laplacians are also used (see, \citep{STWMAKSpringer:2018}), e.g. the random walk Laplacian $\mathbb{L}$  of a graph, defined as \begin{equation}\mathbb{L}=LD^{-1}  = I -SD^{-1} \end{equation}
 }
$\mathcal{L}$ of the graph represented by~$S$ is defined as 
\begin{equation}\label{eq:normLapDef}\mathcal{L}=D^{-1/2}L D^{-1/2}= I -D^{-1/2}S D^{-1/2} .
\end{equation}%

Finding a partition into $k$ groups according to the RCut criterion can be reformulated in terms of finding (discrete) indicator vectors $\mathbf{y}_j$ of cluster membership to $C_j$, $j=1,...,k$, and such that the sum
$\sum_{j=1}^k \mathbf{y}_j^T L \mathbf{y}_j$ reaches its minimum.

Similarly, finding a partition into $k$ groups according to the NCut criterion can be reformulated in terms of finding (discrete) indicator vectors $\mathbf{y}_j$, $j=1,...,k$, such that the sum
$\sum_{j=1}^k \mathbf{y}_j^T \mathfrak{L} \mathbf{y}_j $ 
reaches its minimum.  Again, this is an NP-hard task, so under relaxation one can consider the indicator vectors as continuous, requiring only that such (pseudo) indicator vectors be of unit length and orthogonal to each other.

Based on the Rayleigh-Ritz theorem \citep{Gol96}, it is concluded that the sought pseudo-indicator vectors are the eigenvectors of the respective Laplacian, related to the lowest eigenvalues. If we cluster into $k$ clusters, we take the lowest $k$ eigenvectors (except the last, which is certain to be equal to zero). 
These pseudo-indicator vectors span the embedding space for the objects which are then clustered in this embedding using, e.g. $k$-means algorithm. For an explanation and further details, see, e.g.~\citep{vonLuxburg:2007} or~\citep{STWMAKSpringer:2018}. 
Afterward, note that $L$-based clustering means GSC based on combinatorial Laplacian $L$, and $N$-based clustering means GSC based on normalized Laplacian $\mathfrak{L}$.
Note that by projecting the original data onto the space spanned by $k$ eigenvectors of the corresponding Laplacian, we discover an interesting feature of such a procedure: it naturally combines clustering with (nonlinear) dimensionality reduction.

In the following sections, we explore additional significant embeddings. Other embeddings for objects have also been formulated in the past. 

\subsection{$K$-embedding}

The $K$-embedding 
{\citep{Plosone2025}}  
aims at placing objects $i,\ell$ at distances equal to $1-s_{i\ell}$ and then to perform $k$-means clustering in this embedding. The resulting clustering fits a criterion similar to that of $RCut$. 

This embedding is defined as follows. Let $A$ be a matrix of the form:
\begin{equation} \label{eq:defA}
    A= \mathbf{1}\mathbf{1}^T-I-S\;
\end{equation}
\noindent where $I$ is the identity matrix, and $\mathbf{1}$ is the (column) vector consisting of ones, both of appropriate dimensions%
, and $S$ is the already defined similarity matrix of $n$ objects. 
The matrix $A$ is non-negative and has a diagonal equal to zero, so it may be considered as a kind of (squared) pseudo-distance, needed by the Gower's embedding method \citep{Gower:1966}.  
Let $K$ be the $n \times n$ matrix of the (double centered) form \citep{Gower:1966}:
\begin{equation}
    K=-\frac12(I-\frac1{n}\mathbf{1}\mathbf{1}^T)A(I-\frac1{n}\mathbf{1}\mathbf{1}^T)\;
\end{equation} 

By the Spectral Theorem, it can be represented as
\[ K=V\Lambda V^T=V\Lambda^{1/2}\Lambda^{1/2} V^T=
(V\Lambda^{1/2})(V\Lambda^{1/2})^T \;\]
where $\Lambda$ is the diagonal matrix of the eigenvalues of $K$, and $V$ is the matrix whose columns are the corresponding eigenvectors (of unit length) of $K$. Denoting $Y=V\Lambda^{1/2}$, we obtain $K=YY^T$.
According to Theorem 3, claim 4 in \citep{MAK:2019:FI:Gower}, detailing the respective theorem of \cite{Gower:1966},
the rows of $Y$, hence the columns of $Y^T$ constitute an embedding, called \emph{$K$-embedding}, such that squared distances of embedded points are equal to squared distances, represented by the matrix $A$.%
\footnote{Note that the mentioned Theorem 3 was proven for Euclidean embeddings. For its validity under non-euclidean embeddings see \citep{RAKMAKSTW:2020:trick}.
}

Let $\mathbf{z}_i=\Lambda^{1/2} V^T_{i}$, where $V_i$ stands for $i$-th row of $V$, be the $i$th column of $Y^T$. According to the mentioned Theorem 3, 
$\|\mathbf{z}_i-\mathbf{z}_\ell\|^2=
 A_{i\ell}$.
The non-diagonal elements of the matrix $A$ in the equation \eqref{eq:defA} are equal to $1-s_{i\ell}$.  Therefore 
\begin{equation}    \label{eq:embeddist}
\|\mathbf{z}_i-\mathbf{z}_\ell\|^2=
 A_{i\ell}= 
1-s_{i\ell}
\end{equation}
for $i\ne \ell$. Hence upon performing $k$-means clustering in this space we \emph{de facto} try to maximize the average sum of similarities within a cluster, as the following criterion is minimized:
\begin{equation}\label{eq:Kminimization}
Q^{[Kbased]}(\Gamma)=\sum_{j=1}^k \sum_{i\in C_j} ||\mathbf{z}_i-\boldsymbol\mu(C_j)||^2   
\end{equation}
This differs slightly from the RCut criterion which seeks to minimize the average sum of similarities to elements outside of the cluster. 
The advantage is that it can be easily reformulated in terms of Term-Vector Space embedding so that cluster membership has a clear explanation in the textual domain, as shown in {\citep{Plosone2025}}.

\subsection{$\mathcal{M}$-embedding}

The $\mathcal{M}$-embedding,  {\citep{Plosone2025}}, aims at placing objects $i,j$ at distances equal to $1/d_{ii}+1/d_{\ell\ell} -s_{i\ell}/(d_{ii}d_{\ell\ell})$, and then to perform weighted  $k$-means clustering in this embedding. The resulting clustering fits a criterion identical to that of NCut.
Let $\mathcal{E}$ be a matrix of the form   
\begin{equation}\label{eq_E}
    \mathcal{E}= \mathbf{1}\mathbf{1}^T-I\;
\end{equation}
and $\mathcal{A}$ be the matrix
\begin{equation}
    \mathcal{A}= D^{-1}{(\mathcal{E}D+D\mathcal{E}-2S)}D^{-1}\;
\end{equation}
Let $\mathcal{M}$ be the matrix of the form:
\begin{equation}
    \mathcal{M}=-\frac12(I-\frac1{n}\mathbf{1}\mathbf{1}^T)\mathcal{A}(I-\frac1{n}\mathbf{1}\mathbf{1}^T)\;
\end{equation}
We proceed with $\mathcal{M}$ in a way similar to that of the $K$ matrix. 
Let $\Lambda$ be the diagonal matrix of eigenvalues of $\mathcal{M}$, and $V$ the matrix where the columns are the corresponding eigenvectors (of unit length) of $\mathcal{M}$. 
Then $\mathcal{M}=V\Lambda V^T$. 
Let $\boldsymbol{\zeta}_i=\Lambda^{1/2} V^T_{i}$, where $V_i$ stands for $i$-th row of $V$.  
Let $\boldsymbol{\zeta}_i,\boldsymbol{\zeta}_\ell$ be the {embedding}s of the items $i,\ell$, respectively. This embedding shall be called \emph{$\mathcal{M}$-embedding}.
Then 
\begin{equation}    \label{eq:embeddistfrak}
\|\boldsymbol{\zeta}_i-\boldsymbol{\zeta}_\ell\|^2= \mathcal{A}_{i\ell}= (d_{ii}+d_{\ell\ell}-2s_{i\ell})/(d_{ii}d_{\ell\ell})
\end{equation}
for $i\ne \ell$, and zero otherwise. 
For this embedding,  weighted $k$-means clustering on the vectors $\boldsymbol{\zeta}_i$ with weights equal to $\omega_{i}=d_{ii}$ respectively is performed. 
Clustering via weighted $k$-means with weights $\omega_{i}$ in the $\mathcal{M}$ embedding will optimize the following criterion
 \begin{equation} \label{eq:Mminimization}
 Q^{[\mathcal{M} based]}(\Gamma; \boldsymbol{\omega})=\sum_{j=1}^k \sum_{i\in C_j} \omega_i\|\boldsymbol{\zeta}_i-\boldsymbol\mu_{\boldsymbol{\omega}}(C_j)\|^2    
 \end{equation}
 whereby 
$$
 \boldsymbol\mu_{\boldsymbol{\omega}}(C_j)
 =
 \frac{
 \sum_{\in C_j} \omega_i\boldsymbol{\zeta}_i
 }
 {
 \sum_{\in C_j} {\omega}_i
 }
 =\frac{1}{\Vol_j}
 \sum_{\in C_j} \omega_i\boldsymbol{\zeta}_i
 $$
Both NCut clustering and $\mathcal{M}$-embedding clustering seek to minimize the average sum of similarities to elements outside of the cluster, whereby the averaging is not performed via dividing by cluster cardinality but rather by cluster volume. 
The advantage is that $\mathcal{M}$-embedding can be easily reformulated in terms of term-vector space embedding so that cluster membership has a clear explanation in textual domain, as proven in {\cite{Plosone2025}}.

\subsection{$\mathcal{B}$-embedding}
Both NRCut clustering and $\mathcal{B}$-embedding clustering, explained below, seek to minimize the average sum of similarities to elements outside of the cluster, whereby the averaging is not done by dividing by the number of clusters, but rather by the number of clusters and their volume.

Let us introduce the following notation:
$s'_{i\ell}=s_{i\ell}$ for $i\ne\ell$ and $s'_{ii}=1$,
$d'_{ii}=\sum_\ell s'_{i\ell}$,
${D'}$ be the matrix with diagonal $d'_{ii}$ and zeros elsewhere. 
$\omega'_{i}= d'_{ii}$, 
${\Vol}'_j=\sum_{i \in C_j} d'_{ii}$, 
$F_j=\sum_{i \in C_j} {d'}_{ii}^{-1}$, 
$F=\sum_{i \in \mathcal{D}} {d'}_{ii}^{-1}$, 
$\omega'_S=\sum_{i \in S} d'_{ii}$, 
$\omega'_{il}=\omega'_i \omega'_l$,

Define  the  $\mathcal{A}$ matrix as follows: 
\begin{equation}
    \mathcal{A}= \mathcal{E}{D'}^{-2} + {D'}^{-2}\mathcal{E} -  2{D'}^{-1}S{D'}^{-1}\;
\end{equation}
where $\mathcal{E}$ is defined in equation~\eqref{eq_E}, and  $D',S$ being defined as previously.  
Let $\mathcal{B}$ be the matrix of the form:
\begin{equation}
    \mathcal{B}=-\frac12(I-\frac1{n}\mathbf{1}\mathbf{1}^T)\mathcal{A}(I-\frac1{n}\mathbf{1}\mathbf{1}^T)\;
\end{equation}
We proceed with $\mathcal{B}$ in a similar way as with $K$ matrix. 
Note that $\mathbf{1}$ is an eigenvector of $\mathcal{B}$, with the corresponding eigenvalue equal to 0. 
All other eigenvectors must be orthogonal to it as $\mathcal{B}$ is real and symmetric, so for any other eigenvector $\mathbf{v}$ of $\mathcal{B}$ we have: $\mathbf{1}^T\mathbf{v}=0$.

Let $\Lambda$ be the diagonal matrix of eigenvalues of $\mathcal{B}$, and $V$ the matrix where the columns are the corresponding eigenvectors (unit length) of $\mathcal{B}$\Bem{$K$}. 
Then $\mathcal{B}=V\Lambda V^T$. 
Let $\boldsymbol{\zeta}_i=\Lambda^{1/2} V^T_{i}$, where $V_i$ stands for $i$-th row of $V$.  
Let $\boldsymbol{\zeta}_i,\boldsymbol{\zeta}_\ell$ be the {embedding}s of the items $i,\ell$, resp. 
This embedding shall be called \emph{$\mathcal{B}$-embedding}.
Then 
\begin{equation}    \label{eq:NRembeddistfrak}
\|\boldsymbol{\zeta}_i-\boldsymbol{\zeta}_\ell\|^2= \mathcal{A}_{i\ell}= 
\frac1{{d'}_{ii}^2}+\frac1{{d'}_{\ell\ell}^2}-2\frac{s_{i\ell}}{{d'}_{ii}{d'}_{\ell\ell}}
\end{equation}
Let us now discuss the performance of weighted $k$-means clustering on the vectors $\boldsymbol{\zeta}_i$ with weights equal to ${d'}_{ii}$, respectively.

 \begin{equation} \label{eq:RNKminimization}
 Q^{[\mathcal{B} based]}(\Gamma; \boldsymbol{\omega'})=\sum_{j=1}^k \sum_{i\in C_j} \omega'_i\|\boldsymbol{\zeta}_i-\boldsymbol\mu_{\boldsymbol{\omega'}}(C_j)\|^2    
 \end{equation}
 whereby 
$$
 \boldsymbol\mu_{\boldsymbol{\omega'}}(C_j)
 =
 \frac{
 \sum_{\in C_j} \omega'_i\boldsymbol{\zeta}_i
 }
 {
 \sum_{\in C_j} {\omega'}_i
 }
 =\frac{1}{{\Vol'}_j}
 \sum_{\in C_j} \omega'_i\boldsymbol{\zeta}_i
 $$
This  embedding can be easily reformulated in terms of term-vector space embedding so that cluster membership has a clear explanation in textual domain. For more details, see \cite{RozdzialvMonografia:2025}. 
  
\section{Coping with real datasets}\label{sec:challenges}

{
In our research, we concentrate on the domain of textual documents. 
We assume that the data is embedded in Term Vector Space and consequently, the features of each document are the terms contained therein (that is words). We remove, however, from consideration those terms that occur only once in the entire collection. If this operation leads to empty documents, we discard them. 
Document similarity is the cosine similarity of document vectors in this space.
}

The GSC methods perform very well when the similarity matrix fits the requirement of a block diagonal matrix, such that similarities of objects belonging to the same cluster (block) are on average higher than similarities between objects belonging to different clusters \citep{Fischer:2005}, \citep{Su:2020}.  
Under such circumstances, for a broad range of similarity matrices, the clustering results using either method are (nearly) identical. 
 
In practice, however, we often encounter situations far from this ideal. In such cases, the data set cannot be divided into clearly cut subsets.
Many objects do not have a clear preference for being similar to objects in a particular cluster. This can be compared to a situation where you have a ``sea of objects'' with some groups rising ``above sea level'' and forming ``islands'' that are   well-defined  clusters, while the rest are hard to  classify. This may be seen as  fitting into the broad view of rough sets. There exist ``core'' clusters and there exist ``boundary objects''. 

\subsection{Review of literature on handling boundary data in clustering via rough sets}

Rough set theory offers a range of tools to address uncertainty, vagueness, and incompleteness effectively. Among these tools, approximate clustering is an active area of research, see e.g. \citep{Bello17} for a discussion and extensive bibliography. An interesting overview of applications of rough set theory in big data analysis and processing is given in \citep{Pieta2021}. In particular, the authors discuss the advantages of combining this theory with other approaches, such as fuzzy sets, probabilistic concepts, and deep learning.  They note that this latter hybrid idea seems to be promising for developing new methods, related tools, and extensions. See also \citep{Peres:2012}, especially the chapter by Lingras and Peters on pages 23--50, as well as Part XIV of the proceedings~\citep{Babu2014}. An easy and accessible introduction into various aspects of rough clustering is provided by ~\cite{Lingras2011}. Particularly these authors describe also methods hybridizing rough clustering with other techniques such as genetic algorithms, Kohonen self-organizing networks, and support vector clustering. 
 
Since $k$-means clustering,  is one of the most popular and effective clustering algorithms, see e.g. \citep{Jain2010}, we will focus on its different variants in the following. 

In its original formulation, the algorithm, called HKM (Hard K Means) for short, tries to minimize the objective function
\begin{subequations}\label{eq:kmeans}
\begin{align}
J(X,\Gamma) &= \sum_{\ell=1}^k \sum_{i \in C_\ell} \|\mathbf{x}_i - \boldsymbol{\mu}_\ell\|^2 = \frac{1}{2}\sum_{\ell=1}^k \frac{1}{n_\ell}\sum_{i \in C_\ell}\sum_{j\in C_\ell}\|\mathbf{x}_i - \mathbf{x}_j\|^2\;\label{eq:kmeans1}\ \\
&= \sum_{\ell=1}^k \sum_{i =1}^n u_{i\ell}\|\mathbf{x}_i - \boldsymbol{\mu}_\ell\|^2 = \tra(XX^\texttt{T}) - \tra \big(H^\texttt{T} (XX^\texttt{T})H \big)\;\label{eq:kmeans2}
\end{align}
\end{subequations}
Here 
$X=(\mathbf{x}_1, \dots, \mathbf{x}_n)^{\mathrm{T}}$ 
is a set of $n$ objects described by real-valued vectors~$\mathbf{x}_i$, $i=1,\dots,n$,  
$\boldsymbol{\mu}_{\ell}= (1/n_{\ell})\sum_{i \in C_{\ell}}\mathbf{x}_i$ is the centroid of $\ell$-th cluster, and $n_\ell$ is its size. The number $u_{i\ell} \in \{0,1\}$ equals 1 if $i$-th object belongs to cluster $C_\ell$ and $u_{i\ell} = 0$ otherwise. $H \in \mathbb{R}^{n \times k}$ is a matrix with normalized degrees $h_{i\ell} = u_{i\ell}/\sqrt{n_\ell}$. Such a normalization ensures that $H$ is an orthonormal matrix, i.e. $H^\texttt{T}H=\mathbb{I}$. The first line of the above equation, \eqref{eq:kmeans1}, explains the nature of HKM: it returns compact clusters containing elements that are as similar to each other as possible. A drawback of this algorithm is that it only returns satisfactory results if the clusters are linearly separable. The second line of the above equation, \eqref{eq:kmeans2}, provides a solution. Noticing that $XX^\texttt{T}$ is a symmetric and semi-positive definite matrix (spd for short), we can replace $XX^\texttt{T}$ by any other spd matrix, $K$, which leads to a kernel $k$-mean, \citep{STC04}. This new variant allows us to identify the non-linear structures, making it more suitable for real data sets. Formally, the kernel clustering algorithm relies on transforming samples from the original $m$-dimensional Euclidean space (where $m$ is the size of the vector~$\mathbf{x}$) into a possibly infinite-dimensional feature space in which the differences between samples can be preserved and even enhanced. By using $k$-means for this new representation, we can get a better clustering performance. Of course, this requires choosing the right kernel function with the right parameters. 

In practice, e.g. in text document clustering, we face situations where data can belong to several clusters simultaneously. To deal with such cases, \cite{Bezdek81} generalized the objective function~\eqref{eq:kmeans} by introducing the following criterion
\begin{equation}\label{eq:FCM}
J_F(X,\Gamma) = \sum_{i=1}^n \sum_{\ell=1}^k \tilde{u}_{i\ell}^\alpha \|\mathbf{x}_i - \boldsymbol{\mu}_\ell\|^2\;   
\end{equation}
where $\tilde{u}_{i\ell} \in [0,1]$ is the degree to which element $\mathbf{x}_i$ belongs to cluster $C_{\ell}$, and $\alpha>1$ is a hyperparameter, called also fuzziness exponent. Typically,~$\alpha=2$ is assumed. If $\alpha \rightarrow 1$, the values of $\tilde{u}_{i\ell}$ converge to 0 or 1, and the algorithm becomes identical to the $k$-means algorithm. On the other hand, if $\alpha \rightarrow \infty$, then the values of $\tilde{u}_{i\ell}\rightarrow 1/k$, i.e. we get a completely fuzzy partition. The algorithm minimizing the objective function~\eqref{eq:FCM} under additional probabilistic constraint $\sum_{\ell=1}^k \tilde{u}_{i\ell}=1$ is known as the Fuzzy $c$ Means, or FCM algorithm. Using membership values results in a more flexible and interpretable description of the clusters. In addition to its undoubted advantages, the algorithm also has significant drawbacks. Like KHM, it is sensitive to the choice of initial parameters and prefers spherical clusters. It also performs poorly on data sets with unbalanced clusters and is sensitive to noise and outliers. Furthermore, the computational complexity of FCM is higher than that of $k$-means, \citep{axioms2024}. Various modifications that have been introduced to deal with these drawbacks are described, for example, in Section 3.3 of \citep{STWMAKSpringer:2018}. 

To broaden the applicability of FCM and improve the accuracy of clustering, two ``kernelized'' versions of FCM are developed: (a) centroids are constructed in the original $m$-dimensional Euclidean space, and (b) centroids are located in the (infinite-dimensional) feature space. Both approaches are briefly described by \cite{GP10}. In addition, these authors conducted a comprehensive comparative analysis of fuzzy clustering and these two variants of kernel-based FCM. They found that there was no statistically significant difference between the kernel-based algorithms and FCM and its modification proposed by 
\cite{Gustafson:1978}.
They also observed that kernel-based FCM algorithms are generally amenable to the choice of specific values of the kernel parameters.

FCM and HKM have another common feature: if the number of correctly grouped objects increases, the number of incorrectly grouped objects decreases. On the contrary, \cite{Pet2015}, \cite{Peters2013} noted that in rough clustering, the numbers of correctly and incorrectly clustered objects are no longer complementary, hence the number of incorrectly clustered objects can be minimized independently. This property is important in practical applications, where minimizing the number of incorrectly clustered objects is more important than maximizing the number of correctly clustered objects. 

One of the first variants of rough $k$-means, called hereafter RKM, was proposed by~\cite{Lingras2004}. The authors assumed that each cluster $C_\ell$ is described by the lower, $\underline{C}_\ell$, and upper, $\overline{C}_\ell$, approximation. It is known, that $\underline{C}_\ell$ contains objects that certainly belong to $C_\ell$, while $\overline{C}_\ell \supseteq \underline{C}_\ell$ contains objects that possibly belong to~$C_\ell$. Thus, the following properties must be satisfied:
\begin{itemize}
\item[(i)] each element $\mathbf{x}_i$ can belong to at most one lower approximation~$\underline{C}_\ell$,
\item[(ii)] if $\mathbf{x}_i \in \underline{C}_\ell$, then $\mathbf{x}_i \in \overline{C}_\ell$, and
\item[(iii)] if $\mathbf{x}_i$ does not belong to any lower approximation, then it is a member of at least two upper approximations. 
\end{itemize}
To decide about membership of any object $\mathbf{x}$ to a proper cluster approximation they introduced the set
\[T(\mathbf{x}) = \{t\colon \|\mathbf{x}-\boldsymbol{\mu}_t\| - \|\mathbf{x}-\boldsymbol{\mu}_{\ell^*}\| < \tau, \quad \&\quad t \ne \ell^*\}\; \]
where $\ell^*$ is the index of the centroid closest to $\mathbf{x}$, and $\tau>0$ is a user-defined threshold. Then, $\mathbf{x}$
is assigned to the upper approximations $\overline{C}_t$ for all $t \in T(\mathbf{x})$, and otherwise $\mathbf{x}$ is assigned to the lower approximation~$\underline{C}_{\ell^*}$. Now, the new approximation of the centroid is computed as the convex combination of the centers of the lower and boundary regions.  

\cite{Peters2006} has proposed some algorithm stability improvements. He modified the way of determining the sets $T(\mathbf{x})$ as well as the procedure for determining the cluster centroids. In this new formulation, these are computed as the convex combination of the centers of the lower and upper approximations, i.e. 
\begin{equation}\label{eq:muRKM}
\boldsymbol{\mu}_\ell = \left\{
\begin{array}{cl}
  w^L\underline{P}_\ell + w^U\overline{P}_\ell & \textrm{if $|\underline{C}_\ell| > 0$ and $|\overline{C}_\ell| > 0$}\\
  \underline{P}_\ell & \textrm{if $|\underline{C}_\ell| > 0$ and $|\delta({C}_\ell)| = 0$}\\
  \overline{P}_\ell & \textrm{if $|\underline{C}_\ell| = 0$ and $|\delta({C}_\ell)| > 0$}
\end{array}\;
\right.
\end{equation}
where
\[\underline{P}_\ell = \frac{1}{|\underline{C}_\ell|}\sum_{i \in \underline{C}_\ell} \mathbf{x}_i, \quad \overline{P}_\ell = \frac{1}{|\delta({C}_\ell)}\sum_{i \in \delta({C}_\ell)} \mathbf{x}_i\;\]
and $\delta(C_\ell) = \overline{C}_\ell\backslash \underline{C}_\ell$ is the boundary of $\ell$-th cluster, and the parameters $w^L$ and $w^B$, such that $w^L \ge w^B$, $w^L+w^B=1$, define the importance of the lower approximation and boundary area of the cluster. A unification of the Lingras and Peters approaches has been described in \citep{Ubukata2017}.

Another way to improve the quality of RKM by carefully choosing the hyperparameters $\tau$, $w^L$, and $w^B$ was proposed by \cite{MITRA2004}, who used a genetic algorithm to select appropriate values for these parameters. She assumed that $w^L \in (0.5,1.0)$ and $\tau \in (0,0.5)$. The Davies-Bouldin clustering validation index was used as the fitness function. 

\cite{Zhang2016} modified approximate $k$-means by assigning different weights $w_{i\ell}$ to the features. Their solution is similar to the one used to compute centroids in FCM except that the weights are computed using a Gaussian function. The authors also describe related approaches to approximate clustering.

Regarding the relationship between RKM and other variants of the $k$-means algorithm, \cite{Peters2013} said that this algorithm is a useful enrichment of HKM and that it establishes a different subfamily (by processing uncertainty due to missing or incorrect information) compared to fuzzy and possibilistic clustering (which operate by processing ambiguity based on similarity). This is in agreement with the known fact that fuzzy sets and rough sets are orthogonal concepts, \citep{Pawlak1985}. However, \cite{Dubois1990} proposed a different perspective on this problem. Accepting the fact that the two tools are designed for different tasks, they stated that their joint use can bring many advantages. This has given rise to new directions referred to as ``fuzzy rough'' and ``rough fuzzy''. In the case of clustering, we should mention the paper \citep{Mitra2006}, which proposed an extension of RKM to the rough fuzzy $c$-means, or RFCM for short. Here, each sample $\mathbf{x}_i$ occurring in the equation \eqref{eq:muRKM} is weighted by the fuzzy membership value ~$\tilde{u}_{i\ell}^\alpha$ ($\alpha=2$), similarly to the FCM algorithm. An additional value of this paper is that the authors described in detail how to calculate the rough and rough fuzzy variants of the Davies-Bouldin and Dunn indices used to assess clustering quality. The main focus of this article was the adaptation of a new clustering architecture, referred to as collaborative clustering and introduced by~\cite{Pedrycz2002}. Briefly, the method involves processing several subsets of patterns together to find a common structure for all of them, see, for example, ~\citep{Cornuejols2018} for a deeper explanation. 

Another collaborative approaches were proposed by \cite{HU:2017, Yue:2023}. \cite{HU:2017}  combine ensemble clustering with supportive random forest classifiers. \cite{Yue:2023} concentrate on cluster intersection of clusterings obtained using different base clustering methods. They define the similarity of objects as the counts of their co-occurrence in clusters in base clustering (co-association matrix consists of these counts). They then apply the standard Graph Spectral Clustering method to the co-association matrix.  

\cite{Maji2007} noted a weakness of the RFCM: each object has a distinct weight (its fuzzy membership value), resulting in shifting centroids from their desired locations. The algorithm is also sensitive to noise and outliers. The authors proposed the RFPCM (rough-fuzzy possibilistic $c$-means) algorithm to improve it. They postulate that all objects in a lower approximation should have an identical influence on the determination of their centers. This way they developed an algorithm with crisp lower approximations and fuzzy boundaries. Later, \cite{Paul2014}, proposed so-called robust rough-fuzzy $c$-means, in which each cluster is described by its centroid (prototype), a possibilistic fuzzy lower approximation, and a probabilistic fuzzy (upper) boundary. The centroid depends on the mentioned lower approximation and the upper boundary. The algorithm detects overlapping clusters with arbitrary shapes in the presence of noise.


Recently, \cite{Zhao2023} proposed a more sophisticated approach to this problem. First, the so-called transfer learning mechanism, see~\citep{Zhuang2021}, is introduced into rough clustering. Then the transfer rough clustering algorithm's objective function is optimized using the differential evolution algorithm.

\citep{Manish2010} carried out a comparative study of fuzzy and rough clustering and found out that though a conversion between the results of both is frequently possible, the rough set-based clustering has the advantage of better descriptiveness in high dimensions. The advantages of the rough set approach are exploited to improve the FCM algorithm by \citep{YU2024}, leading to creation of the  FRCM algorithm (fuzzy-rough $c$-means).



One of the methods used in the past to determine a set of objects that unambiguously  belong to a cluster and another larger set that possibly contains it was developed based on the so-called credal partitioning, or evidential clustering, \citep{Masson:2008:ECM}. Interestingly, this approach is suitable for adding partial a priori knowledge about cluster membership \citep{Antoine:2012:CECM}.

Another method was proposed by \cite{Whang:2015:nonexhaus}. 
It is based on an extension of weighted $k$-means to make it both overlapping (so that there are boundary objects) and non-exhaustive (to take care of outliers). Parameters control the maximum overlapping level and the share of outliers. It turns out to be applicable to graph cut approximations.  

While there seems to be no direct study on the explainability of rough set-based clustering, explainability itself was a part of various rough set-based prediction methodologies like that of \cite{Cao:2022}. 
\cite{GrzegorowskiJSMS:2023} used the clustering and rough set methods to obtain human-readable cluster descriptions, emphasizing each cluster’s most discernible characteristics. \cite{Singh2017} propose to use reducts (an important notion used in rough set theory) to explain the existence of clusters and to clarify why two clusters are different.

\cite{WANG:201859} apply the graph spectral clustering to non-textual data with the intent of semantic cluster content description based on fuzzy methodology. The authors concentrate on appropriate feature selection  to get the best possible cluster separation.   
\cite{CEKIK2020113691} apply rough set theory to find the best features in short textual documents in the context of classification. 

We already know that replacing the Euclidean distance with kernel-induced metric makes it possible to cluster the objects that belong to non-convex clusters. This gave birth to rough-kernel clustering algorithms. Let us mention a few of them: \citep{Zhu-2008}, \citep{Das-2009}, \citep{hu2010kernelized}, and \citep{Tripathy-2012}. 




\subsection{Our approach to handling boundary data in clustering inspired by rough sets}\label{sec:ourtheory}

The difference of our approach to the problems with distortions of clusterings lies essentially in insisting that the clustering results need to be explainable in the textual domain.
As we already wrote in {\cite{Plosone2025}}, there is a big difference between describing a cluster with a vector of most frequent words occurring therein and the explanation of the cluster. The explanation means that the cluster membership of documents in a cluster is justified by its cosine similarity to cluster prototype (centroid) in the term-vector space. And these two issues are not equivalent.  
In {\cite{Plosone2025}}  we demonstrated that the traditional $L$-based and $N$-based GSC methods can be translated via our proprietary $K$-embedding and $\mathcal{M}$-embedding can be approximately equivalently translated to term-vector space representation and thus be explainable textually by term vectors. None of the aforementioned publications seems to care about it. Only cluster descriptions are considered. 

What is more, even if some methodologies are based on GSC, like \cite{Yue:2023}, and their intermediate results rely on cosine similarity, their approach does not support the explainability, as they rely on GSC clustering co-occurrences in clusterings from various clustering methods rather than on derivation of clusters that are explainable,  which makes the connection to the original document similarity is lost. 

In our approach the idea is to eliminate noisy documents and cluster the rest via the GSC as mentioned earlier methods with inherent explainability property.
This approach is inspired by the rough set theory. The noisy documents are considered as boundary documents. The remaining ones are more likely to be core (lower approximation) documents that is they represent the intrinsic underlying distinct concepts.

\subsection{Theoretical Background}

Let us consider an idealized case in which objects have nonzero similarities within clusters and zero similarities between clusters. 
Suppose also that the objects are arranged such that the similarity matrix has the block-diagonal structure, i.e.  

\begin{equation}
     S =
\begin{pmatrix}  
  S_{[[1]]}	& \Z & \Z& \dots & \Z	 \\
  \Z  & S_{[[2]]}	& \Z& \dots & \Z		\\
  \Z & \Z & S_{[[3]]}  & \dots & \Z		\\
  \dots & \dots & \dots & \dots & \dots \\
  \Z &\Z &\Z & \dots & S_{[[k]]}		
\end{pmatrix},   
\end{equation}
whereby the symmetric matrices $S_{[[i]]}$, $i=1,\dots,k$ are the similarity matrices within the clusters $C_1,\dots,C_k$ (with zero diagonals), with the number of objects within the cluster $C_i$ equal to $n_i>0$.
Let further $N_0=0$ and $N_i=N_{i-1}+n_i$ for $i=1,\dots,k$.

Let $D(S)$ be the diagonal matrix such that $d_{j,j}=\sum_{\ell=1}^{n_k} s_{j,\ell}$ and $L(S)$ be the combinatorial Laplacian $L(S)=D(S)-S$.

\begin{equation}
     L(S) =
\begin{pmatrix}  
  L(S_{[[1]]})	& \Z & \Z& \dots & \Z	 \\
  \Z  & L(S_{[[2]]})	& \Z& \dots & \Z		\\
  \Z & \Z & L(S_{[[3]]})  & \dots & \Z		\\
  \dots & \dots & \dots & \dots & \dots \\
  \Z &\Z &\Z & \dots & L(S_{[[k]]})		
\end{pmatrix},   
\end{equation}

To perform GSC clustering, one takes $k$ eigenvectors of  $L(S)$ assigned to the lowest eigenvalues, whereby any eigenvalue of  $L(S)$ is provably non-negative as $L(S)$ is a Laplacian. 
Let us denote with $\mathbf{v^{(i)}}$ a vector with all elements equal to $\frac{1}{\sqrt{n_i}}$.
It is of unit length and it is an eigenvector of  $L(S_i)$ which corresponds to the lowest eigenvalue of it (that is 0). 
What is more, as observed in \cite{Ng:2001}, due to block-diagonality,   
 our $L(S)$ has $k$ eigenvalues equal to zero with corresponding mutually orthogonal eigenvectors
 $\mathbf{v}_i=(v_{i,1},\dots,v_{i,n_k})^T$, $i \in 1:k$ such  
 that 
 $v_{i,\ell}={v^{(i)}}_{\ell- N_{i-1}}$ for $\ell\in [N_{i-1}+1:N_i]$ and 
  $v_{i,\ell}=0$ for $\ell\not\in [N_{i-1}+1:N_i]$. 

Under these circumstances, the application of $k$-means to 
objects $o_j$, $j \in 1:n_k$ with coordinates $o_j=(v_{1,j}, v_{2,j},\dots,v_{k,j})^T$ will cluster the data correctly. 

Let us however assume the presence of a disturbing object 
$o_{N_k+1}$ with zero similarity to any other object. Then we have the similarity matrix $S'$

\begin{equation}
     S^* =
\begin{pmatrix}  
S & \Z \\
\Z & 0
\end{pmatrix}   
=
\begin{pmatrix}  
    S_{[[1]]}	& \Z & \Z& \dots & \Z	&\Z \\
  \Z  & S_{[[2]]}	& \Z& \dots & \Z		&\Z	\\
  \Z & \Z & S_{[[3]]}  & \dots & \Z		&\Z	\\
  \dots & \dots & \dots & \dots & \dots 	&\dots \\
  \Z &\Z &\Z & \dots & S_{[[k]]}	&\Z\\
    \Z &\Z &\Z & \dots & \Z & 0
\end{pmatrix},   
\end{equation}
Then

\begin{equation}
    L( S^*) =L
\begin{pmatrix}  
S & \Z \\
\Z & 0
\end{pmatrix}   
=
\begin{pmatrix}  
   L( S_{[[1]]})	& \Z & \Z& \dots & \Z	&\Z \\
  \Z  & L(S_{[[2]]})	& \Z& \dots & \Z		&\Z	\\
  \Z & \Z & L(S_{[[3]]})  & \dots & \Z		&\Z	\\
  \dots & \dots & \dots & \dots & \dots 	&\dots \\
  \Z &\Z &\Z & \dots & L(S_{[[k]]})	&\Z\\
    \Z &\Z &\Z & \dots & \Z & 0
\end{pmatrix},   
\end{equation}

Now $L(S')$ has $k+1$ eigenvalues equal to 0 and same corresponding eigenvalues as $L(S)$ plus the eigenvector 
$\mathbf{v}_{k+1}=(0,0,\dots,0,1)^T$. 
We have the problem. GSC can pick up the right eigenvectors (the first $k$) and we are lucky, or the $k+1$st eigenvector will be taken for clustering and then we get a problem. The last object may have coordinates close enough to an intrinsic cluster center and then we are lucky, or it may be too far away so that the clustering process singles it out as one separate cluster while collapsing two intrinsic clusters to a single one. 

This may happen as follows: 
Consider two datapoints $j$ and $j'$ belonging to distinct intrinsic  clusters $C_i$ and $C_{i'}$. 
The vector difference between them is equal  $o_j-o_{j'}=(0,\dots,1/\sqrt{n_i},0,\dots,-1/\sqrt{n_{i'}},0,\dots,0)^T$.
Then length of $o_j-o_{j'}$ will be equal to $\sqrt{\frac1{n_{i'}}+\frac1{n_i}}$.
On the other hand, 
the vector difference $o_j-o_{N_k+1}$ them is equal  $o_j-o_{N_k+1}=(0,\dots,1/\sqrt{n_i},0,\dots,-1,0,\dots,0)^T$.
Then length of $o_j-o_{N_k+1}$ will be equal to $\sqrt{1+\frac1{n_i}}$.
The table \ref{tab:evGSCexampleComb} presents example eigenvectors obtained when adding 1 or 4 unrelated documents to a document collection with cleanly separated 4 clusters of 2,3,4,5 documents resp.  

With increasing cluster sizes the distance between true clusters $\|o_j-o_{j'}\|$  will decrease down to zero, while the dissimilar element's distance to objects of any cluster $\|o_j-o_{N_k+1}\|$  will get closer to 1. 
At a point it will be more advantageous for $k$-means to separate the dissimilar element as a single cluster and to collapse two intrinsic clusters. 
With increasing number of elements dissimilar to all the other elements, the situation will get worse. 

This misfortune can be prevented with our rough set based elimination process. 

In real settings, this situation of pure block-diagonality and of a pure dissimilar object may happen rarely, but instead low similarity objects may exist so that the behavior of the clustering process will approximate the mentioned one. 
This is easily seen from Theorem 2 formulated in   \cite{Ng:2001}   
implying that given slight noise, the eigenvectors and hence the clustering behave in a similar manner.
\footnote{A deeper analysis of such noisy conditions may be found in 
\cite{Peng:2017}
as well as in 
\cite{Macgregor:2022}.
}
Hence our proposed method will work also.  

To check this, we generated a similarity matrix for 4 clusters with cluster sizes 8 12 16 20.  
Then added noise in the range from=0., to=0.000001. 
$k$-means (with tenfold restart) applied to the space spanned by 4 eigenvectors of the combinatorial Laplacian related to 4 lowest eigenvalues retrieved correctly  4 clusters with sizes 8 12 16 20. 

However, for the same design, if one unrelated document was added, one got cluster sizes 1, 16, 28, 12. With two unrelated documents, cluster sizes were 8 12 16 20 . With 4 unrelated documents the cluster sizes were 17, 21, 21, 1. 

For true clusters of sizes 16 24 32 40 the returned clusters, when 4 unrelated documents were added,  were: 
 97, 2, 1, 16

\begin{table}
 \footnotesize
    \begin{tabular}{|r|l|l|l|l|l|}
  \hline
$\mathbf{o}$  &$\mathbf{v}_1$ &$\mathbf{v}_2$ &$\mathbf{v}_3$&$\mathbf{v}_5$ &$\mathbf{v}_4$  \\
     \hline
1 &  0 &  0 &  0.707 &  0 &  0 \\ 
2 &  0 &  0 &  0.707 &  0 &  0 \\ 
3 &  0 &  0 &  0 &  0 &  -0.577 \\ 
4 &  0 &  0 &  0 &  0 &  -0.577 \\ 
5 &  0 &  0 &  0 &  0 &  -0.577 \\ 
6 &  0 &  -0.5 &  0 &  0 &  0 \\ 
7 &  0 &  -0.5 &  0 &  0 &  0 \\ 
8 &  0 &  -0.5 &  0 &  0 &  0 \\ 
9 &  0 &  -0.5 &  0 &  0 &  0 \\ 
10 &  -0.447 &  0 &  0 &  0 &  0 \\ 
11 &  -0.447 &  0 &  0 &  0 &  0 \\ 
12 &  -0.447 &  0 &  0 &  0 &  0 \\ 
13 &  -0.447 &  0 &  0 &  0 &  0 \\ 
14 &  -0.447 &  0 &  0 &  0 &  0 \\ 
15 &  0 &  0 &  0 &  1 &  0 \\ 
\hline
  \end{tabular}%
  \begin{tabular}{|r|l|l|l|l|l|l|l|l|}
  \hline
$\mathbf{o}$  &$\mathbf{v}_1$ &$\mathbf{v}_2$ &$\mathbf{v}_3$  &$\mathbf{v}_5$ &$\mathbf{v}_6$ &$\mathbf{v}_7$ &$\mathbf{v}_8$ &$\mathbf{v}_4$\\
     \hline
1 &  0 &  0 &  0.707 &  0 &  0 &  0 &  0 &  0 \\ 
2 &  0 &  0 &  0.707 &  0 &  0 &  0 &  0 &  0 \\ 
3 &  0 &  0 &  0 &  0 &  0 &  0 &  0 &  -0.577 \\ 
4 &  0 &  0 &  0 &  0 &  0 &  0 &  0 &  -0.577 \\ 
5 &  0 &  0 &  0 &  0 &  0 &  0 &  0 &  -0.577 \\ 
6 &  0 &  -0.5 &  0 &  0 &  0 &  0 &  0 &  0 \\ 
7 &  0 &  -0.5 &  0 &  0 &  0 &  0 &  0 &  0 \\ 
8 &  0 &  -0.5 &  0 &  0 &  0 &  0 &  0 &  0 \\ 
9 &  0 &  -0.5 &  0 &  0 &  0 &  0 &  0 &  0 \\ 
10 &  -0.447 &  0 &  0 &  0 &  0 &  0 &  0 &  0 \\ 
11 &  -0.447 &  0 &  0 &  0 &  0 &  0 &  0 &  0 \\ 
12 &  -0.447 &  0 &  0 &  0 &  0 &  0 &  0 &  0 \\ 
13 &  -0.447 &  0 &  0 &  0 &  0 &  0 &  0 &  0 \\ 
14 &  -0.447 &  0 &  0 &  0 &  0 &  0 &  0 &  0 \\ 
15 &  0 &  0 &  0 &  0 &  0 &  0 &  1 &  0 \\ 
16 &  0 &  0 &  0 &  0 &  0 &  1 &  0 &  0 \\ 
17 &  0 &  0 &  0 &  0 &  1 &  0 &  0 &  0 \\ 
18 &  0 &  0 &  0 &  1 &  0 &  0 &  0 &  0 \\ 
\hline
  \end{tabular}%
 
  \caption{ Example of combinatorial Laplacian eigenvectors related to zero eigenvalue for a dataset consisting of 4 separated groups of documents (of sizes 2,3,4,5) 
  with one document (left table) or four documents (right table) unrelated to anything  } 
\label{tab:evGSCexampleComb}
\end{table}

Let us mention also issues with normalized Laplacian 
$\mathcal{L}(S)= D(S)^{-1/2}L(S)D(S)^{-1/2}$. 
For the $S'$ matrix described above, $\mathcal{L}(S')$ cannot be computed as one of elements of $D(S')$ is equal to zero, so the inverted square root cannot be computed. 
Therefore we need to investigate other scenarios. 
First, note that

\begin{equation}
     \mathcal{L}(S) =
\begin{pmatrix}  
  \mathcal{L}(S_{[[1]]})	& \Z & \Z& \dots & \Z	 \\
  \Z  & \mathcal{L}(S_{[[2]]})	& \Z& \dots & \Z		\\
  \Z & \Z & \mathcal{L}(S_{[[3]]})  & \dots & \Z		\\
  \dots & \dots & \dots & \dots & \dots \\
  \Z &\Z &\Z & \dots & \mathcal{L}(S_{[[k]]})		
\end{pmatrix},   
\end{equation}

Consider now the following similarity matrix

\begin{equation}
     S^{**} =
\begin{pmatrix}  
S & \Z & \Z \\
\Z & 0 &\epsilon \\
\Z  &\epsilon  &0
\end{pmatrix}   
=
\begin{pmatrix}  
    S_{[[1]]}	& \Z & \Z& \dots & \Z	&\Z &\Z \\
  \Z  & S_{[[2]]}	& \Z& \dots & \Z		&\Z &\Z	\\
  \Z & \Z & S_{[[3]]}  & \dots & \Z		&\Z &\Z	\\
  \dots & \dots & \dots & \dots & \dots 	&\dots &\dots\\
  \Z &\Z &\Z & \dots & S_{[[k]]}	&\Z&\Z \\
    \Z &\Z &\Z & \dots & \Z & 0 &\epsilon \\
    \Z &\Z &\Z & \dots & \Z  &\epsilon & 0
\end{pmatrix},   
\end{equation}
Then

\begin{equation}
    \mathcal{L}( S^{**)} =\mathcal{L}
\begin{pmatrix}  
S & \Z & \Z \\
\Z & 0 &\epsilon \\
\Z  &\epsilon  &0
\end{pmatrix}   
=
\begin{pmatrix}  
    \mathcal{L}(S_{[[1]]})	& \Z & \Z& \dots & \Z	&\Z &\Z \\
  \Z  & \mathcal{L}(S_{[[2]]})	& \Z& \dots & \Z		&\Z &\Z	\\
  \Z & \Z & S_{[[3]]}  & \dots & \Z		&\Z &\Z	\\
  \dots & \dots & \dots & \dots & \dots 	&\dots &\dots\\
  \Z &\Z &\Z & \dots & \mathcal{L}(S_{[[k]]})	&\Z&\Z \\
    \Z &\Z &\Z & \dots & \Z & \frac{\epsilon}{\sqrt{\epsilon\epsilon}} &-\frac{\epsilon}{\sqrt{\epsilon\epsilon}} \\
    \Z &\Z &\Z & \dots & \Z  &-\frac{\epsilon}{\sqrt{\epsilon\epsilon}} & \frac{\epsilon}{\sqrt{\epsilon\epsilon}}
\end{pmatrix},   
\end{equation}

$k+1$ eigenvalues of $\mathcal{L}( S^{**)} $ are equal zero whereby 
the first $k$ corresponding  eigenvectors are those of $L(S)$ and the last one is of the form 
$(\Z, \frac{1}{sqrt{2}},  \frac{1}{sqrt{2}})^T$.
The argument of the problem with such unrelated documents goes here also along the already outlined argument for combinatorial Laplacian, but with higher cardinalities of clusters.  
The table \ref{tab:evGSCexampleNorm} presents example eigenvectors obtained when adding 2 or 4 unrelated documents to a document collection with cleanly separated 4 clusters of 4,6,8,10 documents resp.

\begin{table}
 \tiny
  \begin{tabular}{|r|l|l|l|l|l|}
  \hline
$\mathbf{o}$ &$\mathbf{v}_1$ &$\mathbf{v}_5$  &$\mathbf{v}_2$ &$\mathbf{v}_3$ &$\mathbf{v}_4$  \\
     \hline
1 &  0 &  0 &  0 &  0.53 &  0 \\ 
2 &  0 &  0 &  0 &  0.471 &  0 \\ 
3 &  0 &  0 &  0 &  0.504 &  0 \\ 
4 &  0 &  0 &  0 &  0.493 &  0 \\ 
5 &  0 &  0 &  -0.411 &  0 &  0 \\ 
6 &  0 &  0 &  -0.391 &  0 &  0 \\ 
7 &  0 &  0 &  -0.41 &  0 &  0 \\ 
8 &  0 &  0 &  -0.415 &  0 &  0 \\ 
9 &  0 &  0 &  -0.422 &  0 &  0 \\ 
10 &  0 &  0 &  -0.4 &  0 &  0 \\ 
11 &  -0.352 &  0 &  0 &  0 &  0 \\ 
12 &  -0.355 &  0 &  0 &  0 &  0 \\ 
13 &  -0.359 &  0 &  0 &  0 &  0 \\ 
14 &  -0.354 &  0 &  0 &  0 &  0 \\ 
15 &  -0.351 &  0 &  0 &  0 &  0 \\ 
16 &  -0.349 &  0 &  0 &  0 &  0 \\ 
17 &  -0.358 &  0 &  0 &  0 &  0 \\ 
18 &  -0.35 &  0 &  0 &  0 &  0 \\ 
19 &  0 &  0 &  0 &  0 &  -0.322 \\ 
20 &  0 &  0 &  0 &  0 &  -0.318 \\ 
21 &  0 &  0 &  0 &  0 &  -0.314 \\ 
22 &  0 &  0 &  0 &  0 &  -0.316 \\ 
23 &  0 &  0 &  0 &  0 &  -0.308 \\ 
24 &  0 &  0 &  0 &  0 &  -0.324 \\ 
25 &  0 &  0 &  0 &  0 &  -0.315 \\ 
26 &  0 &  0 &  0 &  0 &  -0.309 \\ 
27 &  0 &  0 &  0 &  0 &  -0.322 \\ 
28 &  0 &  0 &  0 &  0 &  -0.314 \\ 
29 &  0 &  0.707 &  0 &  0 &  0 \\ 
30 &  0 &  0.707 &  0 &  0 &  0 \\ 
\hline
  \end{tabular}%
  \begin{tabular}{|r|l|l|l|l|l|l|}
  \hline
$\mathbf{o}$  &$\mathbf{v}_1$ &$\mathbf{v}_5$ &$\mathbf{v}_6$ &$\mathbf{v}_2$ &$\mathbf{v}_3$ &$\mathbf{v}_4$ \\
     \hline
1 &  0 &  0 &  0 &  -0.523 &  0 &  0 \\ 
2 &  0 &  0 &  0 &  -0.501 &  0 &  0 \\ 
3 &  0 &  0 &  0 &  -0.49 &  0 &  0 \\ 
4 &  0 &  0 &  0 &  -0.486 &  0 &  0 \\ 
5 &  0 &  0 &  0 &  0 &  -0.414 &  0 \\ 
6 &  0 &  0 &  0 &  0 &  -0.419 &  0 \\ 
7 &  0 &  0 &  0 &  0 &  -0.408 &  0 \\ 
8 &  0 &  0 &  0 &  0 &  -0.387 &  0 \\ 
9 &  0 &  0 &  0 &  0 &  -0.416 &  0 \\ 
10 &  0 &  0 &  0 &  0 &  -0.406 &  0 \\ 
11 &  -0.338 &  0 &  0 &  0 &  0 &  0 \\ 
12 &  -0.345 &  0 &  0 &  0 &  0 &  0 \\ 
13 &  -0.347 &  0 &  0 &  0 &  0 &  0 \\ 
14 &  -0.362 &  0 &  0 &  0 &  0 &  0 \\ 
15 &  -0.359 &  0 &  0 &  0 &  0 &  0 \\ 
16 &  -0.358 &  0 &  0 &  0 &  0 &  0 \\ 
17 &  -0.36 &  0 &  0 &  0 &  0 &  0 \\ 
18 &  -0.359 &  0 &  0 &  0 &  0 &  0 \\ 
19 &  0 &  0 &  0 &  0 &  0 &  -0.314 \\ 
20 &  0 &  0 &  0 &  0 &  0 &  -0.317 \\ 
21 &  0 &  0 &  0 &  0 &  0 &  -0.316 \\ 
22 &  0 &  0 &  0 &  0 &  0 &  -0.324 \\ 
23 &  0 &  0 &  0 &  0 &  0 &  -0.319 \\ 
24 &  0 &  0 &  0 &  0 &  0 &  -0.323 \\ 
25 &  0 &  0 &  0 &  0 &  0 &  -0.308 \\ 
26 &  0 &  0 &  0 &  0 &  0 &  -0.313 \\ 
27 &  0 &  0 &  0 &  0 &  0 &  -0.312 \\ 
28 &  0 &  0 &  0 &  0 &  0 &  -0.314 \\ 
29 &  0 &  0 &  0.707 &  0 &  0 &  0 \\ 
30 &  0 &  0 &  0.707 &  0 &  0 &  0 \\ 
31 &  0 &  0.707 &  0 &  0 &  0 &  0 \\ 
32 &  0 &  0.707 &  0 &  0 &  0 &  0 \\ 
\hline
  \end{tabular}%

  \caption{ Example of normalized Laplacian eigenvectors related to zero eigenvalue for a dataset consisting of 4 separated groups of documents (of sizes 4,6,8,10) 
  with one pair of documents (left table) or two pairs of  documents (right table) unrelated to anything  } 
\label{tab:evGSCexampleNorm}
\end{table}

To check the impact of noise, we generated a similarity matrix for 4 clusters with cluster sizes 8 12 16 20.  
Then added noise in the range from=0., to=0.000001. 
$k$-means (with 200fold restart) applied to the space spanned by 4 eigenvectors of the combinatorial Laplacian related to 4 lowest eigenvalues retrieved correctly  4 clusters with sizes 8 12 16 20. 

However, for the same design, if one pair of unrelated documents was added, one got cluster sizes   20, 28, 8, 2. 

For true clusters of sizes 16 24 32 40 the returned clusters, when 2 pairs of unrelated documents were added,  were: 
 2, 66, 32, 16.

 In all these examples we see that tiny one-doc or two-doc start to impact the final clustering badlyu. Their impct could be elevated if our method of removal of unwanted (unrelated) documents would be applied.

\subsection{Practical Implications}

\begin{figure}
\begin{center}
\includegraphics[trim={0mm 0mm 0mm 18mm},clip,width=0.7\textwidth]{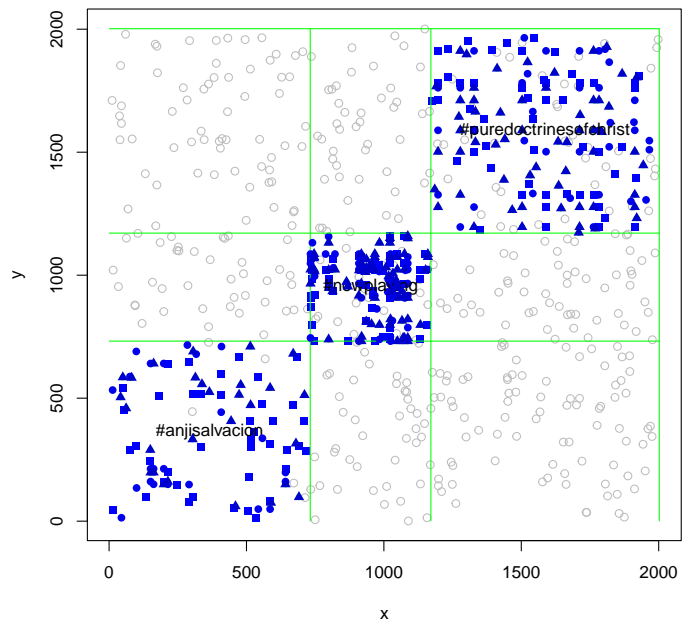}
 \end{center}
\caption{The positions of top 400  similarities between objects (tweets) in blue {(triangles, disks and squares)} and 400 lowest similarities in grey (circles). 
The X axis and Y axis represent internal identifiers of documents (tweets). The documents were 
 sorted by hashtag membership. 
Dataset: TWT.3 (see Sec.\ref{sec:nonblockexperiments}). 
}\label{fig:TWT3ht_mxlinks}
\end{figure}

\begin{figure}
\begin{center}
\includegraphics[trim={0mm 0mm 0mm 18mm},clip,width=0.51\textwidth]{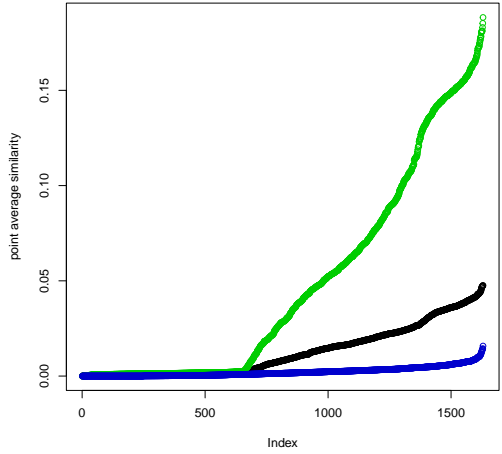} %
 \end{center}
\caption{
Average similarity of a document: 
general - black, 
within a true cluster - green,
between true clusters - blue.  
``True cluster'' means the assigned hashtag. 
Dataset: TWT.3 (see Sec. \ref{sec:nonblockexperiments}). 
}\label{fig:siminout}
\end{figure}

To  understand our idea of handling noisy documents, it may be helpful to look at Figure \ref{fig:TWT3ht_mxlinks}. One can see that that in real data, there exist documents with low similarity to documents within their own clusters. This point is still better illustrated by Figure \ref{fig:siminout}. There may occur in fact a  considerable amount of documents for which their similarities within the same hashtag group and outside do not differ significantly.

\begin{figure}
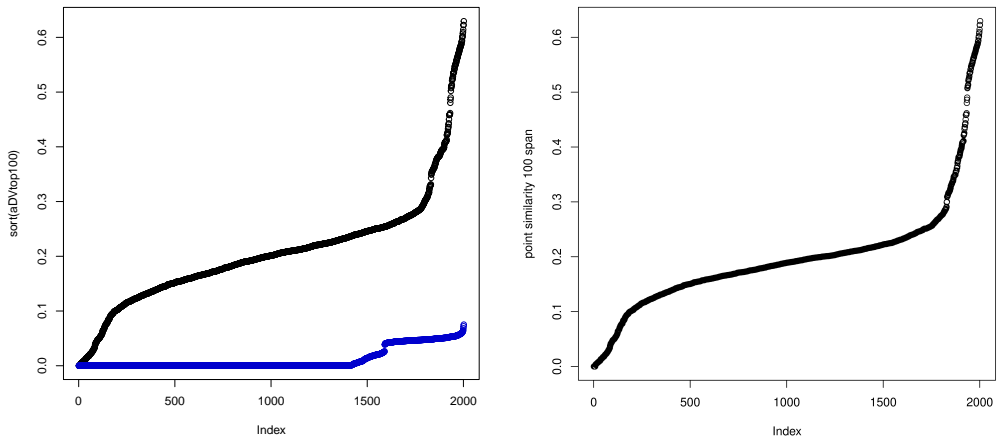

\begin{center}
\includegraphics[trim={0mm 0mm 0mm 18mm},clip,width=0.49\textwidth]{\figaddr{TWT3ht_toplowsim}} %
\includegraphics[trim={0mm 0mm 0mm 18mm},clip,width=0.49\textwidth]{\figaddr{TWT3ht_toplowdiff}} %
 \end{center}
\caption{Left: Objects sorted by increasing average similarity among its top 5\% similarities (black line) and by 
increasing average similarity among its lowest 5\% similarities (green line).
Right: 
Objects sorted by increasing difference between average similarity among its top 5\% similarities and  average similarity among its lowest 5\% similarities.
Dataset: TWT.3 (see Sec.\ref{sec:nonblockexperiments}). 
}\label{fig:TWT3ht_toplowsim}
\end{figure}

Figure \ref{fig:TWT3ht_toplowsim} (left) illustrates this point from another perspective. 
Average of 5\% of top similarities were computed for each document (black line); 
Also average of 5\% of bottom similarities was computed for each document (blue line).
One can see that numerous documents have not a big span between top and bottom similarities (see also Figure \ref{fig:TWT3ht_toplowsim}, right). So their cluster membership may be deemed questionable and their assignment to concrete clusters a bit random. 
 
These observations led us to the following approach: remove documents for which 
average of 5\% of top similarities and average of 5\% of bottom similarities lies below some threshold.  These documents can be considered as spurious, not belonging truly to any cluster.
If they were subject to clustering, they would constitute the boundary region of the cluster they were assigned to. They may be assigned to emerging clusters but they shall not influence their description. 
Under this assumption, the application of any GSC algorithm to the collection after removal of spurious documents will lead to an explainable set of clusters, due to the results mentioned above. 

Our approach to the issues related to approximate nature of GSC algorithms is to exploit the fact that various algorithms approximate the same clustering criterion. Wherever their results intersect, we can be assured of finding good clusters, while disagreements would indicate the cases of doubt. 
Let us take the RCut clustering criterion \eqref{eq:qRCut_def}. 
RCut clustering is NP-hard, while their approximations via $L$-based clustering, equation \eqref{eq:combLapDef}, or $K$-based clustering, equation \eqref{eq:Kminimization}, are not. Both types approximate the same target function but from different perspectives and with differing precision. Therefore, as experiments show, the clusters provided by both of them differ. However, one can be more convinced about true cluster membership when both $L$ and $K$-embedding intersect. 

A similar statement can be made about NCut clustering criterion  \eqref{eq:qNCut_def}. NCut clustering is NP-hard, while their approximations via $N$-based clustering, equation \eqref{eq:normLapDef}, or $\mathfrak{M}$-based clustering, equation \eqref{eq:Mminimization}, are not. Both types approximate the same target function but from different perspectives and with differing precision. Again, as experiments show, the clusters provided by both of them differ. 
Note also that the RCut and NCut may be considered (approximately) identical if cluster volumes are proportional to cluster cardinality. 

Surprisingly enough, under ideal settings of similarity matrices being (approximately) block matrices, the intrinsic clustering is discovered by all approaches (that is, the clustering pre-defined by an artificial data generator, as one can see in our experiments below). 

This supports the previously described idea to filter out subsets of the data that match the characteristics of the block matrix. Our approach to this goal is heuristic: we consider a subset of data to be good enough if the average of some percentage of the highest similarities is sufficiently different from the average of some percentage of the lowest similarities for a given data item.

We perform experiments to investigate how well these hypotheses work.

Note that, for purposes of our experiments, we used pre-defined thresholds (0.1, 0.2) on the difference between 5\% of top and bottom similarities of each document. 
For the purpose of application of our methodology, one should  proceed as as follows: 
(1) For each document $j$, compute $avg\_top_j$ - the average of top 5\% similarities of $d_j$ to other documents
(2)  Compute $avg\_bot_j$ - the average of bottom 5\% similarities of $d_j$ to other documents
(3)  Compute $avg\_diff_j=avg\_top_j-avg\_bot_j$.  
(4) Then sort all documents  $avg\_diff_j$. 
(5)  Visualize this sorted sequence (see e.g. Fig.\ref{fig:TWT3ht_toplowsim} right)
(6)  By visual inspection choose the threshold $t$ (criteria remove a few documents as possible, do not exceed a long plateau)
(7) Remove documents $d_j$ for which $avg\_diff_j<t$

\FloatBarrier

\section{Experiments}\label{sec:experiments}

{
In our research, we focus predominantly on textual documents, like tweets and other. 
We assume that the documents are embedded in the Term Vector Space and that their similarities are cosine similarities between document embedding vectors. These assumptions lead in a natural way to have similarity values ranging from 0 to 1 (as the angles between the vectors in this space belong to the range}
{$[0,\frac{\pi}{2}]$}{).}

As mentioned in the introduction, we have developed explanation methods for selected GSC algorithms for the textual domain.   
However, these explanations may be distorted if there exist noise documents. 
Thus we need two components to handle the problem: a method to remove the noisy data points and a method to check that there are fewer noisy points. 
Our approach here is as follows: we need (1) a method for removal of noisy points that is independent of the clustering methods, (2) distinct clustering methods that converge to the (same) correct clustering under noise removal. 

The removal method that we apply is simple, and clustering method independent: there must be a gap between top and lowest similarities of a document to the other ones. 
The clustering methods that we investigate are the ones described in detail in this paper, $L$-based, $K$-based, $N$-based and $\mathfrak{B}$-based, as well as other listed in Table~\ref{tab:GSCmethods}. they all belong to the GSC family. 

First, we demonstrate in Section 
\ref{sec:blockexperiments} 
that in fact GSC  four clustering methods $L$-based, $K$-based, $N$-based and $\mathfrak{B}$-based, discussed in the paper converge to the same clustering under GSC favourable conditions  that is block-similarity matrices. 
Then in Section \ref{sec:nonblockexperiments} we show the effects of application of our noisy-point-removal data on some real data for the same algorithms. 
And finally in Section \ref{sec:experimentsvarioushashtags} we show that the same works for diverse other methods of GSC.

We have performed experiments\footnote{%
See \naszLink\ for source code and data.
} 
for both synthetic (Sec. \ref{sec:blockexperiments} 
) and real data (Sec. \ref{sec:nonblockexperiments}, \ref{sec:experimentsvarioushashtags}) in order to see if and how the removal of suspected boundary elements contributes to better clustering results.

\subsection{Artificial data block-like similarity matrices}\label{sec:blockexperiments}

The synthetic data sets, used in this section, were generated by creating an empty matrix $S$ and then filling it in with similarities according to the following principles: 
\begin{itemize}
 \item the main diagonal is set to 0,   
 \item the following parameters were used by the generator:
 \begin{itemize}
     \item  $n$ - the intended sample size,  
     \item $m$ - the number of clusters,
     \item $props$ -  a vector (of length $m$) denoting the proportions of the size of the individual clusters, with an entry for each cluster, any numbers allowed, proportions are important,
     \item according to proportions $props$, each of the $n$ the elements is assigned to one of the clusters $1\colon m$ ;  
     \item  $min\_in$ - a vector denoting the minimal similarity within each cluster, with an entry for each cluster, ranging from 0 to 1;  
     \item  $max\_in$ - a vector denoting the maximal similarity within each cluster, with an entry for each cluster, ranging from 0 to 1;  
     \item  $max\_out$ - a vector denoting the   maximal similarity between cluster elements and elements outside the cluster (multiplied by $m$), with an entry for each cluster, ranging from 0 to $m$;  
 \end{itemize}
\item for elements $j,l$ of $i$th cluster, their similarity entry in $S_{j,l}=S_{l,j}$ is sampled from uniform distribution over the interval  
${min\_in}_i$ to ${max\_in}_i$,
\item for elements $j$ from cluster $i$ and $l$ from cluster  $k$,    their similarity entry in $S_{j,l}=S_{l,j}$ is sampled from uniform distribution over the interval  
$0$ to $\frac{{max\_out}_i+{max\_out}_k}{2m}$.
\end{itemize}


\subsubsection{Dataset 1}\label{sec:blockexperiments0}

The first synthetic dataset, obtained according to the above description, consists of four relatively easily discernible clusters with cardinalities:  491, 257, 512, and  240. 
Generator parameters were:  
the size of the sample: n=1500; 
proportions between clusters: props=$(1,1/2,1,1/2)$, 
minimum similarity within each cluster min\_in=$(0.3,0.3,0.3,0.3)$ (here in each cluster the same);  maximum similarity within each cluster max\_in=$(0.7,0.7,0.7,0.7)$ (i.e. identical in each cluster); maximum similarity between elements of different clusters, max\_out=$(0.6,0.6,0.6,0.6)$. 

All four methods, $L$-based, $K$-based, $N$-based and $\mathfrak{B}$-based,    produced correct clustering. 

\begin{figure}
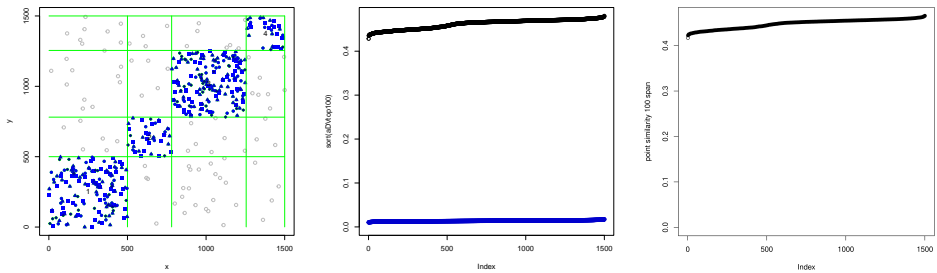

\begin{center}
\includegraphics[trim={0mm 0mm 0mm 18mm},clip,width=0.30\textwidth]{\figaddr{S0S_gen_0_mxlinks}} %
\includegraphics[trim={0mm 0mm 0mm 18mm},clip,width=0.30\textwidth]{\figaddr{S0S_gen_0_toplowsim}} %
\includegraphics[trim={0mm 0mm 0mm 18mm},clip,width=0.30\textwidth]{\figaddr{S0S_gen_0_toplowdiff}} %
 \end{center}
\caption{Left: The positions of top 400  similarities between objects  in blue and 400 lowest similarities in grey. 
Objects sorted by true cluster membership. 
Center: Objects sorted by increasing average similarity among its top 5\% similarities (black line) and by 
increasing average similarity among its lowest 5\% similarities (green line).
Right: 
Objects sorted by increasing difference between average similarity among its top 5\% similarities and  average similarity among its lowest 5\% similarities.
Synthetic dataset~1.
}\label{fig:S_gen_0_mxlinks}
\end{figure}

\begin{figure}
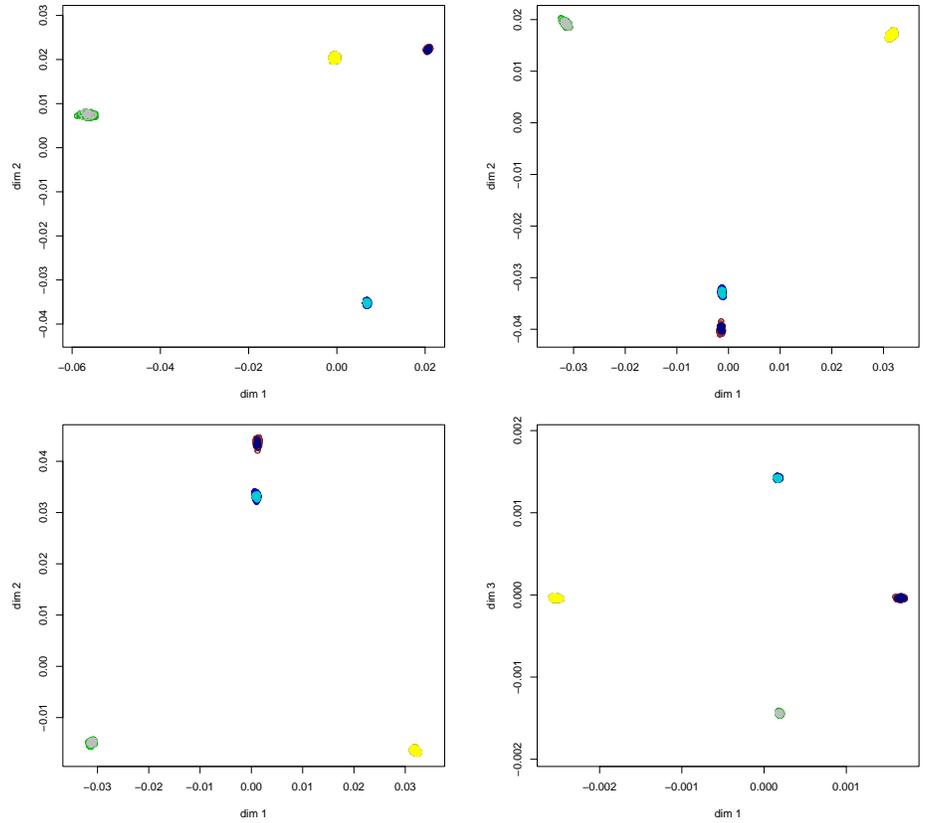

\begin{center}
\includegraphics[trim={0mm 0mm 0mm 18mm},clip,width=0.45\textwidth]{\figaddr{S0S_gen_0_Lemb_12}} %
\includegraphics[trim={0mm 0mm 0mm 18mm},clip,width=0.45\textwidth]{\figaddr{S0S_gen_0_Kemb_12}} %
\includegraphics[trim={0mm 0mm 0mm 18mm},clip,width=0.45\textwidth]{\figaddr{S0S_gen_0_Nemb_12}} %
\includegraphics[trim={0mm 0mm 0mm 18mm},clip,width=0.45\textwidth]{\figaddr{S0S_gen_0_Bemb_13}} %
 \end{center}
\caption{A two-dimensional view at data in the $L$-embedding (top left),  the $K$-embedding (top right), $N$-embedding (bottom left) and $\mathcal{B}$-embedding (bottom right). Different colors reflect the true clustering.  
Synthetic dataset~1.
}\label{fig:S_gen_0_LembTrue}
\end{figure}

Relevant figures showing various aspects are:
\ref{fig:S_gen_0_mxlinks},
\ref{fig:S_gen_0_LembTrue}.

{
In Figure \ref{fig:S_gen_0_LembTrue} and in similar ones, 
dim 1, dim 2, dim 3 etc. are the coordinates of the objects in the respective embedding (In  Figure \ref{fig:S_gen_0_LembTrue} the L-embedding is meant.) 
}

Figure \ref{fig:S_gen_0_mxlinks}, on the left,  illustrates the distribution of the highest and lowest similarities between the documents within the intrinsic clusters and outside of them. 
As should be expected in the case of typical block-diagonal similarity matrices, the highest similarities occur within clusters, while the lowest --- outside of them. 
Figure \ref{fig:S_gen_0_mxlinks}, in the middle, presents the average of top similarities (the upper line) and the average of the lowest similarities (lower line) for each document.  In the case of block-diagonal similarity matrices, there should be a big gap between these two lines. This should be obvious as the top similarities are expected to be similarities within the same cluster, while the lowest are the similarities between clusters. If we assume that there should be not too many clusters, then this will be true if we take 5\% of the highest and 5\% of the lowest similarities. 
Figure \ref{fig:S_gen_0_mxlinks}, to the right, shows the differences between these two means of top and bottom similarities just mentioned. One can expect good clustering properties if this difference is high. 

Figure \ref{fig:S_gen_0_LembTrue} shows that in fact, the clustering methods discussed in this paper detect very well the clusters. The embeddings related to each of the clustering methods separate excellently the clusters. 
Thus, from rough set point of view, there will be no points not belonging to lower approximations (cores) of any cluster.

\FloatBarrier
\subsubsection{Dataset 2}\label{sec:blockexperiments1}
We created four clusters with cardinalities
$497, 251, 496, 256$ respectively, using the generator described at the beginning of Section \ref{sec:blockexperiments}.
Generator parameters were:  
n=1500; 
props=$(1,1/2,1,1/2)$,
min\_in=$(0.3,0.35,0.4,0.45)$,
max\_in=$(0.7,0.65,0.6,0.55)$,
max\_out=$(0.5,0.6,0.7,0.8)$. 

This parameterization led to clusters with differing inner-cluster similarities compared to the previous experiments. Also, the distribution of similarities between clusters varied. 

\begin{figure}
\begin{center}
\includegraphics[trim={0mm 0mm 0mm 18mm},clip,width=0.49\textwidth]{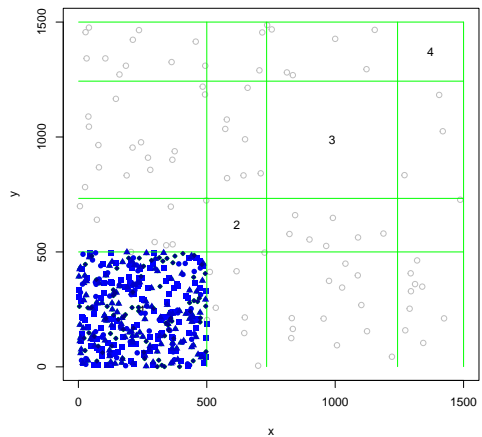} %
 \end{center}
\caption{The positions of top 400  similarities between objects  in blue and 400 lowest similarities in grey. 
Objects sorted by true cluster membership. 
Synthetic dataset~2.
}\label{fig:S_gen_1_mxlinks}
\end{figure}

\begin{figure}
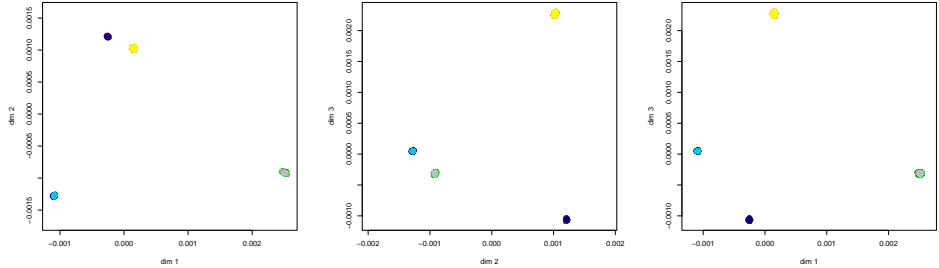

\begin{center}
\includegraphics[trim={0mm 0mm 0mm 18mm},clip,width=0.3\textwidth]{\figaddr{S1S_gen_1_BembTrue_12}} %
\includegraphics[trim={0mm 0mm 0mm 18mm},clip,width=0.3\textwidth]{\figaddr{S1S_gen_1_BembTrue_23}} %
\includegraphics[trim={0mm 0mm 0mm 18mm},clip,width=0.3\textwidth]{\figaddr{S1S_gen_1_BembTrue_13}} %
 \end{center}
\caption{A glance at the $\mathcal{B}$-embedding space from three different perspectives (axes 1,2, or 2,3 or 1,3). Different colors reflect the true clustering. Synthetic dataset~2.  
}\label{fig:S_gen_1_BembTrue}
\end{figure}

All four methods produced correct clustering. 
Relevant figures showing various aspects are:
\ref{fig:S_gen_1_mxlinks},
\ref{fig:S_gen_1_toplowsim},
\ref{fig:S_gen_1_KembTrue},
\ref{fig:S_gen_1_LembTrue},
\ref{fig:S_gen_1_NembTrue},
\ref{fig:S_gen_1_BembTrue}. 

The ideas behind all these figures are identical as in the previous subsection. 

\begin{figure}
\begin{center}
\includegraphics[trim={0mm 0mm 0mm 18mm},clip,width=0.49\textwidth]{\figaddr{S1S_gen_1_toplowsim}} %
\includegraphics[trim={0mm 0mm 0mm 18mm},clip,width=0.49\textwidth]{\figaddr{S1S_gen_1_toplowdiff}} %
 \end{center}
\caption{Left: Objects sorted by increasing average similarity among its top 5\% similarities (black line) and by 
increasing average similarity among its lowest 5\% similarities (green line).
Right: 
Objects sorted by increasing difference between average similarity among its top 5\% similarities and  average similarity among its lowest 5\% similarities.
Synthetic dataset~2.
}\label{fig:S_gen_1_toplowsim}
\end{figure}

\begin{figure}
\begin{center}
\includegraphics[trim={0mm 0mm 0mm 18mm},clip,width=0.3\textwidth]{\figaddr{S1S_gen_1_Lemb_12}} %
\includegraphics[trim={0mm 0mm 0mm 18mm},clip,width=0.3\textwidth]{\figaddr{S1S_gen_1_Lemb_23}} %
\includegraphics[trim={0mm 0mm 0mm 18mm},clip,width=0.3\textwidth]{\figaddr{S1S_gen_1_Lemb_13}} %
 \end{center}
\caption{A glance at the $L$-embedding space from three different perspectives (axes 1,2, or 2,3 or 1,3). Different colors reflect the true clustering.  
Synthetic dataset~2.
}\label{fig:S_gen_1_LembTrue}
\end{figure}

\begin{figure}
\begin{center}
\includegraphics[trim={0mm 0mm 0mm 18mm},clip,width=0.3\textwidth]{\figaddr{S1S_gen_1_Kemb_12}} %
\includegraphics[trim={0mm 0mm 0mm 18mm},clip,width=0.3\textwidth]{\figaddr{S1S_gen_1_Kemb_23}} %
\includegraphics[trim={0mm 0mm 0mm 18mm},clip,width=0.3\textwidth]{\figaddr{S1S_gen_1_Kemb_13}} %
 \end{center}
\caption{A glance at the $K$-embedding space from three different perspectives (axes 1,2, or 2,3 or 1,3). Different colors reflect the true clustering.
Synthetic dataset~2.
}\label{fig:S_gen_1_KembTrue}
\end{figure}

\begin{figure}
\begin{center}
\includegraphics[trim={0mm 0mm 0mm 18mm},clip,width=0.3\textwidth]{\figaddr{S1S_gen_1_NembTrue_12}} %
\includegraphics[trim={0mm 0mm 0mm 18mm},clip,width=0.3\textwidth]{\figaddr{S1S_gen_1_NembTrue_23}} %
\includegraphics[trim={0mm 0mm 0mm 18mm},clip,width=0.3\textwidth]{\figaddr{S1S_gen_1_NembTrue_13}} %
 \end{center}
\caption{A glance at the $N$-embedding space from three different perspectives (axes 1,2, or 2,3 or 1,3). Different colors reflect the true clustering.
Synthetic dataset~2.
}\label{fig:S_gen_1_NembTrue}
\end{figure}

\FloatBarrier
\subsubsection{Dataset 3}\label{sec:blockexperiments2}

We obtained four clusters with cardinalities
$1012,  325,  125,   38$, respectively, using the generator described at the beginning of Section \ref{sec:blockexperiments}.
Generator parameters were:  
props=$(1,1/3,1/3^2,1/3^3)$,
min\_in=$(0.3,0.35,0.4,0.45)$,
max\_in=$(0.7,0.65,0.6,0.55)$,
max\_out=$(0.6,0.7,0.8,0.9)$
and n=1500.

The cluster sizes differed strongly (by factor 27). 
The inner-cluster similarity was bigger for smaller clusters. 

\begin{figure}
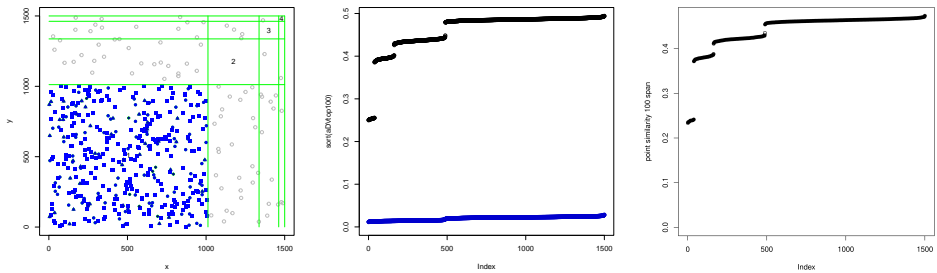

\begin{center}
\includegraphics[trim={0mm 0mm 0mm 18mm},clip,width=0.30\textwidth]{\figaddr{S2S_gen_2_mxlinks}} %
\includegraphics[trim={0mm 0mm 0mm 18mm},clip,width=0.30\textwidth]{\figaddr{S2S_gen_2_toplowsim}} %
\includegraphics[trim={0mm 0mm 0mm 18mm},clip,width=0.30\textwidth]{\figaddr{S2S_gen_2_toplowdiff}} %
 \end{center}
\caption{Left: The positions of top 400  similarities between objects  in blue and 400 lowest similarities in grey. 
Objects sorted by true cluster membership. 
Center: Objects sorted by increasing average similarity among its top 5\% similarities (black line) and by 
increasing average similarity among its lowest 5\% similarities (green line).
Right: 
Objects sorted by increasing difference between average similarity among its top 5\% similarities and  average similarity among its lowest 5\% similarities.
Synthetic dataset 3.
}\label{fig:S_gen_2_mxlinks}
\end{figure}

\begin{figure}
\begin{center}
\includegraphics[trim={0mm 0mm 0mm 18mm},clip,width=0.45\textwidth]{\figaddr{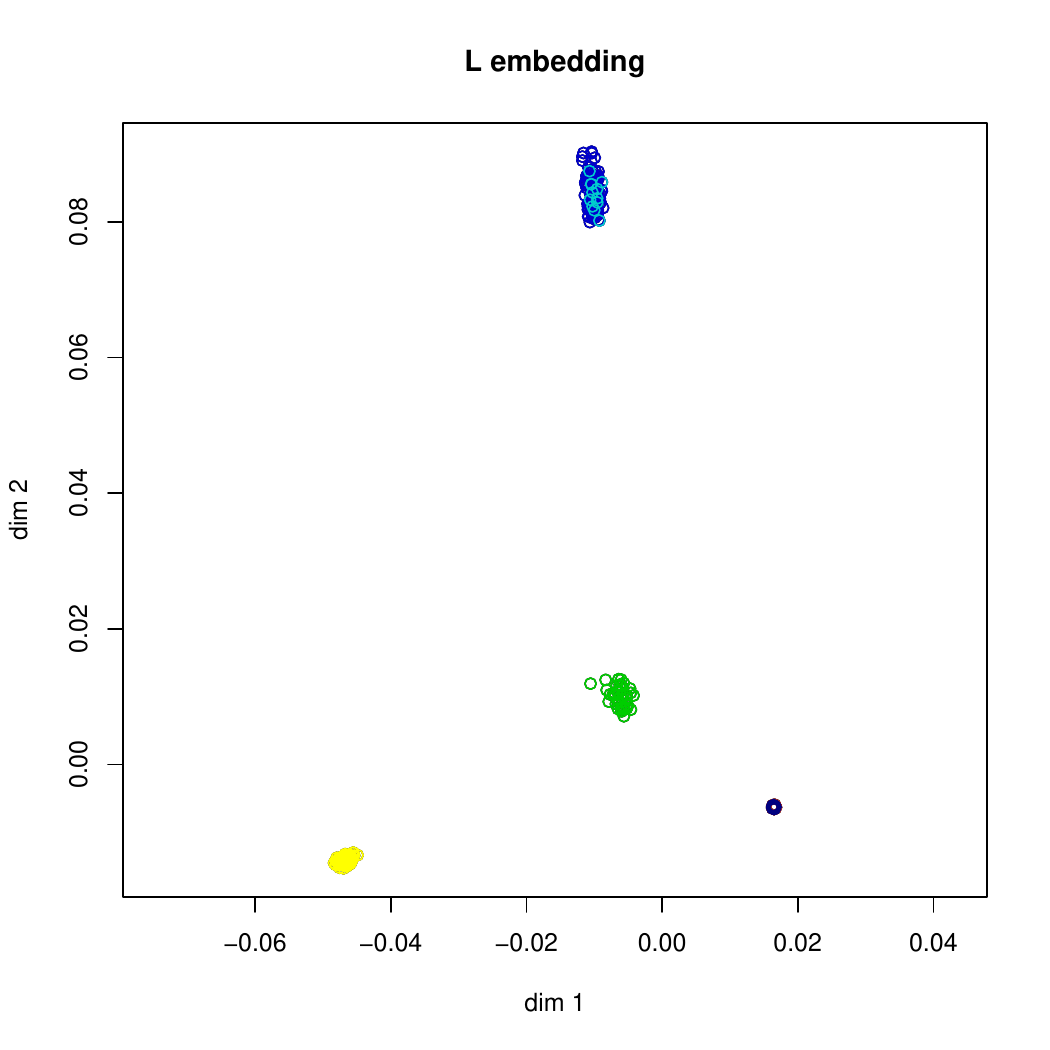}} %
\includegraphics[trim={0mm 0mm 0mm 18mm},clip,width=0.45\textwidth]{\figaddr{S2S_gen_2_Kemb_12}} %
\includegraphics[trim={0mm 0mm 0mm 18mm},clip,width=0.45\textwidth]{\figaddr{S2S_gen_2_Nemb_12}} %
\includegraphics[trim={0mm 0mm 0mm 18mm},clip,width=0.45\textwidth]{\figaddr{S2S_gen_2_Bemb_12}} %
 \end{center}
\caption{A two-dimensional view at data in the $L$-embedding (top left),  the $K$-embedding (top right), $N$-embedding (bottom left) and $\mathcal{B}$-embedding (bottom right). Different colors reflect the true clustering.  
Synthetic dataset~3.
}\label{fig:S_gen_2_LembTrue}
\end{figure}

All four methods produced correct clustering. 
Relevant figures showing various aspects are:
\ref{fig:S_gen_2_mxlinks},
\ref{fig:S_gen_2_LembTrue}.

The ideas behind all these figures are identical as in the previous subsections.

\FloatBarrier
\subsubsection{Dataset 4}\label{sec:blockexperiments3}

We crated four clusters  with cardinalities
$36,  108,  348, 1008$, respectively, using the generator described at the beginning of Section \ref{sec:blockexperiments}.
Generator parameters were:  
n=1500, 
props=($1,3,3^2,3^3$),
min\_in=$(0.3,0.35,0.4,0.45)$,
max\_in=$(0.7,0.65,0.6,0.55)$,
max\_out=$(0.6,0.7,0.8,0.9)$.

As in the preceding experiment, the cluster cardinalities differed extremely. However, this time the smaller the cluster, the smaller the similarities within it. 
Due to this change, the $L$-based clustering failed, as visible in Table~\ref{tab:trueL_S_Gen_3}. If one would like to predict the true clustering from $L$-based clustering, a 23\% error would be committed. 

\begin{figure}
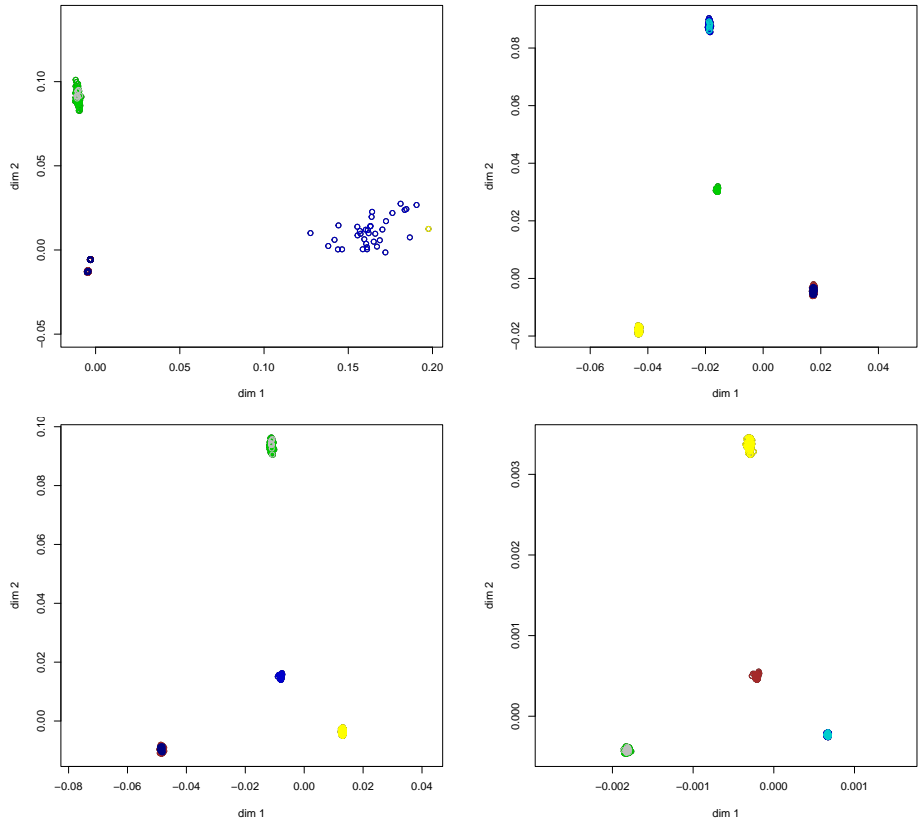

\begin{center}
\includegraphics[trim={0mm 0mm 0mm 18mm},clip,width=0.45\textwidth]{\figaddr{S3S_gen_3_Lemb_12}} %
\includegraphics[trim={0mm 0mm 0mm 18mm},clip,width=0.45\textwidth]{\figaddr{S3S_gen_3_Kemb_12}} %
\includegraphics[trim={0mm 0mm 0mm 18mm},clip,width=0.45\textwidth]{\figaddr{S3S_gen_3_NembTrue_12}} %
\includegraphics[trim={0mm 0mm 0mm 18mm},clip,width=0.45\textwidth]{\figaddr{S3S_gen_3_Bemb_12}} %
 \end{center}
\caption{A two-dimensional view at data in the $L$-embedding (top left),  the $K$-embedding (top right), $N$-embedding (bottom left) and $\mathcal{B}$-embedding (bottom right). Different colors reflect the true clustering.  
Synthetic dataset~4.
}\label{fig:S_gen_3_LembTrue}
\end{figure}

Relevant figures showing various aspects are:
\ref{fig:S_gen_3_mxlinks},
\ref{fig:S_gen_3_LembTrue}. 

The ideas behind all these figures are identical to those in the previous subsections. 
One can see in Fig. \ref{fig:S_gen_3_LembTrue}, that $L$-embedding based clustering encounters problems as the data is not well separated. This is surely caused by the fact that for some data points $L$-embedding-based between the similarities within and outside of a cluster is not that sharp, as visible in Figure \ref{fig:S_gen_3_mxlinks}, on the right. 

\begin{table} 
\centering
\begin{tabular}{|r|r|r|r|r|}
\hline TRUE/PRED& 1& 2& 3& 4\\
\hline  1& 35& 0& 0& 1\\
\hline  2& 0& 108& 0& 0\\
\hline  3& 0& 0& 348& 0\\
\hline  4& 0& 0& 1008& 0\\
\hline
\end{tabular}
\caption{Relationship between true clustering and the result of   $L$-based GSC;  errors:  23.2\%. 
Synthetic dataset 4. }
\label{tab:trueL_S_Gen_3}

\end{table} 

\begin{figure}
\begin{center}
\includegraphics[trim={0mm 0mm 0mm 18mm},clip,width=0.30\textwidth]{\figaddr{S3S_gen_3_mxlinks}} %
\includegraphics[trim={0mm 0mm 0mm 18mm},clip,width=0.30\textwidth]{\figaddr{S3S_gen_3_toplowsim}} %
\includegraphics[trim={0mm 0mm 0mm 18mm},clip,width=0.30\textwidth]{\figaddr{S3S_gen_3_toplowdiff}} %
 \end{center}
\caption{Left: The positions of top 400  similarities between objects  in blue and 400 lowest similarities in grey. 
Objects sorted by true cluster membership. 
Center: Objects sorted by increasing average similarity among its top 5\% similarities (black line) and by 
increasing average similarity among its lowest 5\% similarities (green line).
Right: 
Objects sorted by increasing difference between average similarity among its top 5\% similarities and  average similarity among its lowest 5\% similarities.
Synthetic dataset 4. 
}\label{fig:S_gen_3_mxlinks}
\end{figure}

\FloatBarrier

\subsection{Why different GSC yield similar results for the artificial data}

Assume that average similarities between objects in distinct clusters are small amounting to $s_0$. Within any cluster $C_j$ of cardinality $n_j$ let the average object similarity $s_j$ be higher than $s_0$. 
Consider two clusters $C_j,C_{j'}$. 
If the similarity between objects within $j$ is greater than between objects from $C_j$ and $C_{j'}$, that is   $s_j> s_0$, 
then to ensure that 
\begin{multline}
\frac{s_j}{\sqrt{(n_j-1)s_{j}+(n-n_j)s_0}\sqrt{(n_j-1)s_{j}+(n-n_j)s_0}} \\
>\frac{s_0}{\sqrt{(n_j-1)s_{j}+(n-n_j)s_0}\sqrt{(n_{j'}-1)s_{j'}+(n-n_j)s_0}} 
\end{multline}

it is required that: 
$$ \frac{s_j^2}{s_0^2}
> \frac{(n_j-1)s_{j}+(n-n_j)s_0}{(n_{j'}-1)s_{j'}+(n-n_j)s_0}$$
This inequity holds for various conditions e.g. when true cluster sizes are the same and within-cluster are the same. But also various restrictions can be derived when varying both. Note that we speak here about approximations. 
If the above inequity holds then also the combinatorial Laplacian and normalized Laplacian are approximately of the same mathematical form so that the clustering solutions are also similar. 
Though an in-depth analysis would be necessary to identify the exact conditions for clustering equivalence, one sees that untypical structures are responsible for differences between the clustering results. 

\subsection{Real data}\label{sec:nonblockexperiments}

\begin{table}[ht]
\centering
\begin{tabular}{|r|l|c|}
\hline
    No.  & hashtag & count \\
    \hline
  0& 90dayfiance & 316\\
	 1& tejran & 345\\
	 2& ukraine & 352\\
	 3& tejasswiprakash & 372\\
	 4& nowplaying & 439\\
  \hline
\end{tabular}
\begin{tabular}{|r|l|c|}
\hline
    No.  & hashtag & count \\
    \hline
	 5& anjisalvacion & 732\\
	 6& puredoctrinesofchrist & 831\\
	 7& 1 & 1105\\
	 8& lolinginlove & 1258\\
	 9& bbnaija & 1405\\
  \hline
\end{tabular}
\caption{TWT.10 dataset - hashtags and cardinalities of the set of related tweets  used in the experiments}\label{tab:twt10set}
\end{table}

For the experiments with real data we used the following collections:
\begin{itemize}
  \item TWT.10 -- the set of random tweets published on Twitter (currently X) between 2019 and 2023, 
    of length greater than 150 characters each, containing exactly one of the hashtags listed in Table~\ref{tab:twt10set}.\footnote{The dataset will be made available upon publication.}
  \item TWT.3 -- a subset of TWT.10 consisting of tweets containing the hashtags: 
    \texttt{\#anjisalvacion}, \texttt{\#nowplaying} and \texttt{\#puredoctrinesofchrist}. 
\end{itemize}

We demonstrate the issues based on a collection of tweets containing exactly one of the three hashtags: \texttt{\#anjisalvacion}, 
\texttt{\#puredoctrinesofchrist}, 
\texttt{\#nowplaying} of cardinalities  732, 831, 439 resp. 
This dataset was characterized by the following cut criterion values:
 RCut criterion for true = 22.53161,
 NCut criterion for true = 0.4605525,
 NRCut criterion for true = 0.4496038,
 RCut criterion for random = 124.3468,
 NCut criterion for random = 2.003028,
 NRCut criterion for random = 1.971268,
 whereby ``true'' means the value for cluster assignment according to hashtag values, while random means random cluster assignment (cluster cardinalities as in true clustering).

Our first question is: is there any grid structure in the data? 
Figure \ref{fig:TWT3ht_mxlinks} visualizes the top 0.01\% of similarities between objects. Objects (sorted by true hashtags) are represented by the X and Y axes, and the dots point at the pair of objects which belongs to this top similarity set. 
One can see an emerging block similarity structure.
From rough set theory point of view, predominant number of elements lies within lower approximations (cores) of clusters, while only a limited number of them belongs to the boundary region.

{
Figure \ref{fig:TWT3ht_toplowsim} gives a deeper insight in the similarity structure. 
It shows that some 100 objects for which the difference between top 5\% and bottom 5\% similarities is small. The clustering is more likely to be successful if the within-cluster similarities are significantly greater than similarities between elements of different clusters. 
But if we know that a cluster encompasses more than 5\% of objects, then small top-bottom similarities indicate problems with clustering. We know this even without actually having access to the clustering. 
}

This fact causes the $L$-based and $K$-based and $\mathcal{B}$-based clustering algorithms to perform poorly, see Table~\ref{tab:trueL}, \ref{tab:trueK}, \ref{tab:trueB}. Compared to them, the $N$-based algorithm performs quite well, see Table~\ref{tab:trueN}. 

{
The choice of 5\% boundary is based on two types of considerations: On the one hand we do not expect clusters that consist of fewer than 5\% of objects. On the other hand, the 5\% margins are used as places where tolocate outliers (1\%, or 0.1\% is more adequate for outliers, but 5\% is more reasonable for our sample sizes). So the idea here is: If the extreme similarities do not differ much then the object is most likely to be an outlier itself, as it will not belong really to any cluster and will blur the results. 
}

\begin{table} 
\centering
\begin{tabular}{|r|r|r|r|}
\hline TRUE/PRED& 1& 2& 3\\
\hline  \texttt{\#anjisalvacion} & 729& 3& 0\\
\hline  \texttt{\#nowplaying} & 364& 4& 71\\
\hline  \texttt{\#puredoctrinesofchrist} & 470& 361& 0\\
\hline
\end{tabular}
\caption{$K$-based prediction of cluster membership errors:  42.0\% }
\label{tab:trueK}
\end{table} 

\begin{figure}
\begin{center}
\includegraphics[trim={0mm 0mm 0mm 18mm},clip,width=0.3\textwidth]{\figaddr{TWT3ht_KembTrue_12}} %
\includegraphics[trim={0mm 0mm 0mm 18mm},clip,width=0.3\textwidth]{\figaddr{TWT3ht_KembTrue_23}} %
\includegraphics[trim={0mm 0mm 0mm 18mm},clip,width=0.3\textwidth]{\figaddr{TWT3ht_KembTrue_13}} %
 \end{center}
\caption{A glance at the $K$-embedding space from three different perspectives (axes 1,2, or 2,3 or 1,3). Different colors reflect the true clustering.  
Dataset TWT.3. 
}\label{fig:TWT3ht_KembTrue}
\end{figure}

\begin{figure}
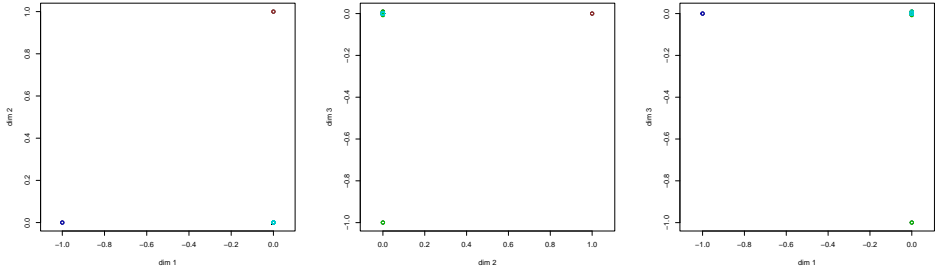

\begin{center}
\includegraphics[trim={0mm 0mm 0mm 18mm},clip,width=0.3\textwidth]{\figaddr{TWT3ht_Lemb_12}} %
\includegraphics[trim={0mm 0mm 0mm 18mm},clip,width=0.3\textwidth]{\figaddr{TWT3ht_Lemb_23}} %
\includegraphics[trim={0mm 0mm 0mm 18mm},clip,width=0.3\textwidth]{\figaddr{TWT3ht_lemb_13}} %
 \end{center}
\caption{A glance at the $L$-embedding space from three different perspectives (axes 1,2, or 2,3 or 1,3). Apparently, there exist points distant from the rest, skewing the clustering.   
Dataset TWT.3
}\label{fig:TWT3ht_LembTrue}
\end{figure}

\begin{table} 
\centering
\begin{tabular}{|r|r|r|r|}
\hline TRUE/PRED& 1& 2& 3\\
\hline  \texttt{\#anjisalvacion}& 0& 0& 732\\
\hline  \texttt{\#nowplaying} & 1& 1& 437\\
\hline  \texttt{\#puredoctrinesofchrist} & 0& 0& 831\\
\hline
\end{tabular}
\caption{$L$-based prediction of cluster membership errors:  58.3\% }
\label{tab:trueL}

\end{table} 

\begin{table} 
\centering
\begin{tabular}{|r|r|r|r|}
\hline TRUE/PRED& 1& 2& 3\\
\hline  \texttt{\#anjisalvacion} & 9& 0& 723\\
\hline  \texttt{\#nowplaying} & 162& 270& 5\\
\hline  \texttt{\#puredoctrinesofchrist} & 829& 1& 1\\
\hline
\end{tabular}
\caption{$N$-based prediction of cluster membership errors:  8.9\% }
\label{tab:trueN}

\end{table} 

\begin{figure}
\begin{center}
\includegraphics[trim={0mm 0mm 0mm 18mm},clip,width=0.3\textwidth]{\figaddr{TWT3ht_NembTrue_12}} %
\includegraphics[trim={0mm 0mm 0mm 18mm},clip,width=0.3\textwidth]{\figaddr{TWT3ht_NembTrue_23}} %
\includegraphics[trim={0mm 0mm 0mm 18mm},clip,width=0.3\textwidth]{\figaddr{TWT3ht_NembTrue_13}} %
 \end{center}
\caption{A glance at the $N$-embedding space from three different perspectives (axes 1,2, or 2,3 or 1,3). Different colors reflect the true clustering.  
Dataset TWT.3
}\label{fig:TWT3ht_NembTrue}
\end{figure}

\begin{figure}
\begin{center}
\includegraphics[trim={0mm 0mm 0mm 18mm},clip,width=0.3\textwidth]{\figaddr{TWT3ht_Bemb_12}} %
\includegraphics[trim={0mm 0mm 0mm 18mm},clip,width=0.3\textwidth]{\figaddr{TWT3ht_Bemb_23}} %
\includegraphics[trim={0mm 0mm 0mm 18mm},clip,width=0.3\textwidth]{\figaddr{TWT3ht_Bemb_13}} %
 \end{center}
\caption{A glance at the $\mathcal{B}$-embedding space from three different perspectives (axes 1,2, or 2,3 or 1,3). Different colors reflect the true clustering. Dataset TWT.3. 
}\label{fig:TWT3ht_BembTrue}
\end{figure}

\begin{table} %
\centering
\begin{tabular}{|r|r|r|r|}
\hline TRUE/PRED& 1& 2& 3\\
\hline  1& 998& 1& 1\\
\hline  2& 271& 0& 0\\
\hline  3& 729& 0& 0\\
\hline
\end{tabular}
\caption{$\mathcal{B}$-based prediction of cluster membership errors: 50.0\% }
\label{tab:trueB}
\end{table} %

\FloatBarrier


\begin{figure}
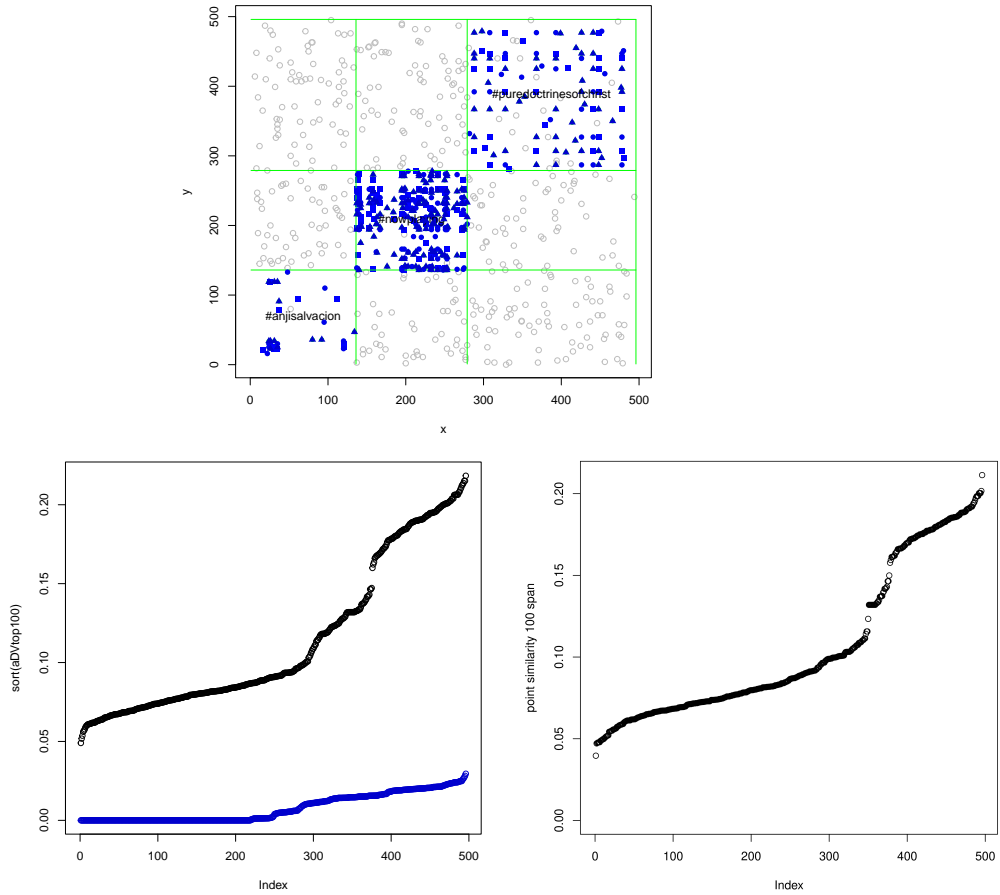

\begin{center}
\includegraphics[trim={0mm 0mm 0mm 18mm},clip,width=0.49\textwidth]{\figaddr{0.2_TWT3ht_mxlinks}} \newline%
\includegraphics[trim={0mm 0mm 0mm 18mm},clip,width=0.49\textwidth]{\figaddr{0.2_TWT3ht_toplowsim}} %
\includegraphics[trim={0mm 0mm 0mm 18mm},clip,width=0.49\textwidth]{\figaddr{0.2_TWT3ht_toplowdiff}} %
 \end{center}
\caption{Left: Objects sorted by increasing average similarity among its top 5\% similarities (black line) and by 
increasing average similarity among its lowest 5\% similarities (green line).
Right: 
Objects sorted by increasing difference between average similarity among its top 5\% similarities and  average similarity among its lowest 5\% similarities.
Dataset TWT.3 after removing boundary data at specified threshold.
}\label{fig:TWT3ht_toplowsim0.2}
\end{figure}

In the second stage of the experiment, 
all tweets were removed for which the difference between the average of  top 5\% of similarities and the average of the bottom 5\% similarities was below 0.2, and then once again 0.1.  
This led to a different statistics of hashtags:
\texttt{\#anjisalvacion}, \texttt{\#nowplaying},  \texttt{\#puredoctrinesofchrist}
having respective cardinalities of 136, 143, and 217.

\begin{figure}
\begin{center}
\includegraphics[trim={0mm 0mm 0mm 18mm},clip,width=0.3\textwidth]{\figaddr{0.2_TWT3ht_KembTrue_12}} %
\includegraphics[trim={0mm 0mm 0mm 18mm},clip,width=0.3\textwidth]{\figaddr{0.2_TWT3ht_KembTrue_23}} %
\includegraphics[trim={0mm 0mm 0mm 18mm},clip,width=0.3\textwidth]{\figaddr{0.2_TWT3ht_KembTrue_13}} %
 \end{center}
\caption{A glance at the $K$-embedding space from three different perspectives (axes 1,2, or 2,3 or 1,3). Different colors reflect the true clustering.  
Dataset TWT.3 after removing boundary data at specified threshold.
}\label{fig:TWT3ht_KembTrue0.2}
\end{figure}

\begin{figure}
\begin{center}
\includegraphics[trim={0mm 0mm 0mm 18mm},clip,width=0.3\textwidth]{\figaddr{0.2_TWT3ht_LembTrue_12}} %
\includegraphics[trim={0mm 0mm 0mm 18mm},clip,width=0.3\textwidth]{\figaddr{0.2_TWT3ht_LembTrue_23}} %
\includegraphics[trim={0mm 0mm 0mm 18mm},clip,width=0.3\textwidth]{\figaddr{0.2_TWT3ht_LembTrue_13}} %
 \end{center}
\caption{A glance at the $L$-embedding space from three different perspectives (axes 1,2, or 2,3 or 1,3). Different colors reflect the true clustering.  
Dataset TWT.3 after removing boundary data at specified threshold.
}\label{fig:TWT3ht_LembTrue0.2}
\end{figure}

\begin{figure}
\begin{center}
\includegraphics[trim={0mm 0mm 0mm 18mm},clip,width=0.3\textwidth]{\figaddr{0.2_TWT3ht_NembTrue_12}} %
\includegraphics[trim={0mm 0mm 0mm 18mm},clip,width=0.3\textwidth]{\figaddr{0.2_TWT3ht_NembTrue_23}} %
\includegraphics[trim={0mm 0mm 0mm 18mm},clip,width=0.3\textwidth]{\figaddr{0.2_TWT3ht_NembTrue_13}} %
 \end{center}
\caption{A glance at the $N$-embedding space from three different perspectives (axes 1,2, or 2,3 or 1,3). Different colors reflect the true clustering.  
Dataset TWT.3 after removing boundary data at specified threshold.
}\label{fig:TWT3ht_NembTrue0.2}
\end{figure}

The tables \ref{tab:trueL0.2}, \ref{tab:trueK0.2}, \ref{tab:trueN0.2}, \ref{tab:trueB0.2}   show the impact of this filtering. 
In each case one can see an improvement of the performance.
 
\begin{table} 
\centering
\begin{tabular}{|r|r|r|r|} 
\hline TRUE/PRED& 1& 2& 3\\
\hline  \#anjisalvacion& 136& 0& 0\\
\hline  \#nowplaying& 0& 143& 0\\
\hline  \#puredoctrinesofchrist& 0& 0& 217\\
\hline
\end{tabular}
\caption{$L$-based prediction of cluster membership errors:  0\% (reduced data)}
\label{tab:trueL0.2}

\end{table} 

\begin{table} 
\centering
\begin{tabular}{|r|r|r|r|}
\hline TRUE/PRED& 1& 2& 3\\
\hline  \texttt{\#anjisalvacion} & 0& 136& 0\\
\hline  \texttt{\#nowplaying} & 71& 23& 49\\
\hline  \texttt{\#puredoctrinesofchrist} & 0& 217& 0\\
\hline
\end{tabular}
\caption{
$K$-based prediction of cluster membership errors:  32.0\% (reduced data)}
\label{tab:trueK0.2}
\end{table} 

\begin{table} 
\centering
\begin{tabular}{|r|r|r|r|}
\hline TRUE/PRED& 1& 2& 3\\
\hline  \texttt{\#anjisalvacion} & 0& 136& 0\\
\hline  \texttt{\#nowplaying} & 127& 0& 16\\
\hline  \texttt{\#puredoctrinesofchrist} & 217& 0& 0\\
\hline
\end{tabular}
\caption{$\mathcal{B}$-based prediction of cluster membership errors:  25.6\% (reduced data)}
\label{tab:trueB0.2}
\end{table} 

\begin{table} 
\centering
\begin{tabular}{|r|r|r|r|}
\hline TRUE/PRED& 1& 2& 3\\
\hline  \texttt{\#anjisalvacion} & 0& 0& 345\\
\hline  \texttt{\#nowplaying} & 0& 160 & 0\\
\hline  \texttt{\#puredoctrinesofchrist} & 316& 0& 0\\
\hline
\end{tabular}
\caption{$N$-based prediction of cluster membership errors:  0.0\% (reduced data)}
\label{tab:trueN0.2}

\end{table} 

\begin{figure}
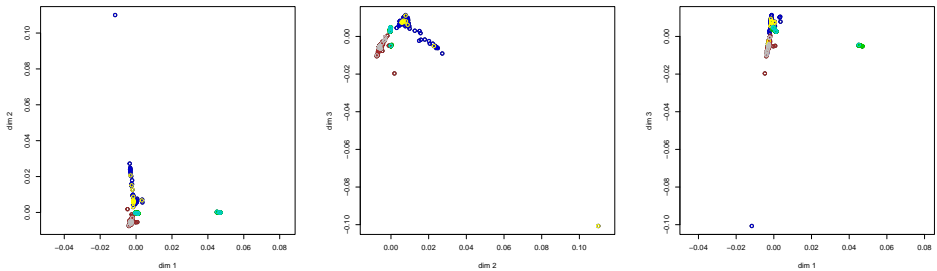

\begin{center}
\includegraphics[trim={0mm 0mm 0mm 18mm},clip,width=0.3\textwidth]{\figaddr{0.2_TWT3ht_BembTrue_12}} %
\includegraphics[trim={0mm 0mm 0mm 18mm},clip,width=0.3\textwidth]{\figaddr{0.2_TWT3ht_BembTrue_23}} %
\includegraphics[trim={0mm 0mm 0mm 18mm},clip,width=0.3\textwidth]{\figaddr{0.2_TWT3ht_BembTrue_13}} %
 \end{center}
\caption{A glance at the $\mathcal{B}$-embedding space from three different perspectives (axes 1,2, or 2,3 or 1,3). Different colors reflect the true clustering. Dataset TWT.3 after removing boundary data at specified threshold.
}\label{fig:TWT3ht_BembTrue0.2}
\end{figure}

\FloatBarrier

\subsection{Cluster Explanations}\label{sec:expl}

Though this paper does not deal with the textual explanation per se (an interested Reader shall refer to {\cite{Plosone2025}}), nonetheless it is worthy to have a look at changes induced by the spurious data filtering technique applied. 
Note that the keywords presented below are not the most frequent ones, but rather are selected by an explanation process of cluster centroids.
We observe, for example, that after the application of the technique, the $L$-based cluster explanations appear to be more reasonable. 

\subsubsection{Before removing non-distinguishing tweets}
 
\paragraph{True clusters top words}
Cluster \texttt{\#anjisalvacion}:  
  \textit{anjisalvacion},   \textit{anji},            \textit{dalampasigan},   
  \textit{life},            \textit{happy},           \textit{feelstheconcert}, 
  \textit{birthday},        \textit{people},          \textit{salvacion},      
 \textit{mv}.
Cluster \texttt{\#nowplaying}: 
  \textit{radio},     \textit{unknown},   \textit{app},       \textit{live},      \textit{listen},   
  \textit{hand},      \textit{android},   \textit{apple},     \textit{available}, \textit{download}. 
Cluster \texttt{\#puredoctrinesofchrist}:
  \textit{king},     \textit{god},      \textit{proverbs}, \textit{esv},      \textit{lord},    
  \textit{christ},   \textit{eli},      \textit{let},      \textit{jesus},    \textit{man}.     
  
\paragraph{L-based clusterings top words}
Cluster  2: 
  \textit{anjisalvacion}, \textit{anji},          \textit{king},          \textit{god},          
  \textit{radio},         \textit{proverbs},      \textit{live},          \textit{dalampasigan}, 
  \textit{life},          \textit{esv}.  
  
\paragraph{K-based clusterings top words}
 Cluster  1: 
  \textit{king},     \textit{proverbs}, \textit{god},      \textit{luke},     \textit{lord},    
  \textit{psalm},    \textit{christ},   \textit{man},      \textit{jesus},    \textit{hath}.
 Cluster  2:
  \textit{android},   \textit{app},       \textit{apple},     \textit{available}, \textit{download}, 
  \textit{hand},      \textit{listen},    \textit{live},      \textit{radio},     \textit{store}.
 Cluster  3: 
  \textit{anjisalvacion}, \textit{anji},          \textit{god},           \textit{dalampasigan}, 
  \textit{esv},           \textit{life},          \textit{love},          \textit{people},       
  \textit{happy},         \textit{go}.           
 
\paragraph{N-based clusterings top words}
Cluster  1:
  \textit{radio},     \textit{unknown},   \textit{app},       \textit{live},      \textit{listen},   
  \textit{android},   \textit{apple},     \textit{available}, \textit{download},  \textit{hand}.
Cluster  2: 
  \textit{anjisalvacion},   \textit{anji},            \textit{dalampasigan},   
  \textit{life},            \textit{happy},           \textit{feelstheconcert},
 \textit{birthday},        \textit{people},          \textit{salvacion},      
 \textit{mv}.             
Cluster  3:
  \textit{king},     \textit{god},      \textit{proverbs}, \textit{esv},      \textit{lord},    
  \textit{christ},   \textit{love},     \textit{good},     \textit{man},      \textit{let}.     

\paragraph{B-based clusterings top words}
Cluster 1:
  \textit{anjisalvacion}, \textit{anji},          \textit{king},          \textit{god},          
  \textit{radio},         \textit{proverbs},      \textit{live},          \textit{dalampasigan}, 
  \textit{life},          \textit{esv}.

\subsubsection{After}

\paragraph{True clusters top words}
Cluster \texttt{\#anjisalvacion}:
 \textit{anjisalvacion},  \textit{anji},           \textit{dalampasigan},   
 \textit{happy},          \textit{life},           \textit{feelstheconcert}, 
 \textit{birthday},       \textit{people},         \textit{salvacion},      
 \textit{mv}.             
Cluster  \texttt{\#nowplaying}:
 \textit{radio},    \textit{unknown},  \textit{app},      \textit{listen},   \textit{live},     
 \textit{android},  \textit{apple},    \textit{available}, \textit{download}, \textit{hand}.     
Cluster \texttt{\#puredoctrinesofchrist}: 
 \textit{king},     \textit{god},      \textit{proverbs}, \textit{christ},   \textit{lord},     
 \textit{jesus},    \textit{words},    \textit{esv},      \textit{accepting}, \textit{good}.     

\paragraph{L-based clusterings top words}

Cluster  1:
 \textit{anjisalvacion},  \textit{anji},           \textit{dalampasigan},   
 \textit{happy},          \textit{life},           \textit{feelstheconcert}, 
 \textit{birthday},       \textit{people},         \textit{salvacion},      
 \textit{mv}.             
Cluster  2: 
 \textit{king},     \textit{god},      \textit{proverbs}, \textit{christ},   \textit{lord},     
 \textit{jesus},    \textit{words},    \textit{esv},      \textit{accepting}, \textit{good}.     
Cluster  3: 
 \textit{radio},    \textit{unknown},  \textit{app},      \textit{listen},   \textit{live},     
 \textit{android},  \textit{apple},    \textit{available}, \textit{download}, \textit{hand}.     

\paragraph{K-based clusterings top words}
Cluster  1:
 \textit{anjisalvacion},  \textit{anji},           \textit{dalampasigan},   
 \textit{happy},          \textit{life},           \textit{feelstheconcert}, 
 \textit{birthday},       \textit{people},         \textit{salvacion},      
 \textit{mv}.             
Cluster  2: 
 \textit{airplay},           \textit{com},               \textit{email},             
 \textit{global},            \textit{go},                \textit{goglobalradio},     
 \textit{goglobalradiogmail}, \textit{info},              \textit{radio},             
 \textit{unknown}.           
Cluster  3: 
 \textit{king},    \textit{god},     \textit{proverbs}, \textit{live},    \textit{app},     
 \textit{listen},  \textit{radio},   \textit{christ},  \textit{hand},    \textit{download}.

\paragraph{$N$-based clusterings top words}
Cluster  1: 
 \textit{anjisalvacion},  \textit{anji},           \textit{dalampasigan},   
 \textit{happy},          \textit{life},           \textit{feelstheconcert},
 \textit{birthday},       \textit{people},         \textit{salvacion},      
 \textit{mv}.             
Cluster  2:
 \textit{king},     \textit{god},      \textit{proverbs}, \textit{christ},   \textit{lord},     
 \textit{jesus},    \textit{words},    \textit{esv},      \textit{accepting}, \textit{good}.     
Cluster  3:
 \textit{radio},    \textit{unknown},  \textit{app},      \textit{listen},   \textit{live},     
 \textit{android},  \textit{apple},    \textit{available}, \textit{download}, \textit{hand}.     

\paragraph{$\mathcal{B}$-based clusterings top words}
Cluster  1:
 \textit{anjisalvacion},  \textit{anji},           \textit{radio},          
 \textit{dalampasigan},   \textit{unknown},        \textit{happy},          
 \textit{live},           \textit{life},           \textit{feelstheconcert},
 \textit{birthday}.       
Cluster  2: 
 \textit{afrobeats},  \textit{app},        \textit{artist},     \textit{caribbean},  
 \textit{family},     \textit{favorite},   \textit{hits},       \textit{mainstream}, 
 \textit{myradiolive}, \textit{o}.          
Cluster 3: 
 \textit{king},     \textit{proverbs}, \textit{god},      \textit{christ},   \textit{lord},     
 \textit{jesus},    \textit{words},    \textit{esv},      \textit{accepting}, \textit{good}.

\begin{table}
\centering
    \begin{tabular}{|l|p{10cm}|}
    \hline
    Method Id & Method Name \\
        \hline
        0. &Combinatorial spectral clustering -- unit length rows \\
        \hline
        1. &Combinatorial spectral clustering -- unit length rows and one additional dimension used \\
        \hline
        2. &Kamvar spectral clustering,\\
        \hline 
3. & Kamvar spectral clustering with additional dimension used,\\
        \hline
4. & Normalized spectral clustering,\\
        \hline
5. &Normalized spectral clustering -- unit length rows, \\
        \hline
6. &Normalized spectral clustering -- unit length rows and one additional dimension used, \\
        \hline
        7. &Normalized spectral clustering -- unit length rows and one additional dimension used PLUS prior SVD analysis \\
        \hline
        8. &Normalized spectral clustering -- unit length rows  PLUS prior SVD analysisis.  \\ \hline
    \end{tabular}
    \caption{Identifiers and names of spectral clustering methods analysed for their performance after rough set treatment}
    \label{tab:GSCmethods}
\end{table}

\begin{table}
\centering
    \begin{tabular}{|l|l|}
    \hline
    Hashtag  Id & Hashtag Name\\
    \hline 
  0 & \texttt{\#1} \\
    \hline 
  1 & \texttt{\#90dayfiance} \\
    \hline 
  2 & \texttt{\#anjisalvacion} \\
    \hline 
  3 & \texttt{\#bbnaija} \\
    \hline 
  4 & \texttt{\#lolinginlove} \\
    \hline 
  5 & \texttt{\#nowplaying} \\
    \hline 
  6 & \texttt{\#puredoctrinesofchrist} \\
    \hline 
  7 & \texttt{\#tejasswiprakash} \\
    \hline 
  8 & \texttt{\#tejran} \\
    \hline 
  9 & \texttt{\#ukraine}  \\ \hline
   \end{tabular}
    \caption{Identifiers and names of hashtags   analysed for GSC algorithm performance after rough set treatment}
    \label{tab:hashtagnames}
\end{table}

\subsection{Real Data Experiments - Multiple Clustering Methods}\label{sec:experimentsvarioushashtags}

In order not just to limit our experiments to a specific set of GSC clustering methods and a specific set of hashtags, we performed the experiments described in this section. 
We took the TWT.10 data described above, and sampled 5 groups of hashtags consisting of 3,4,5,6,7,8,9 hashtags and one group containing all 10 hashtags. The hashtags defined the datasets to be clustered (tweets with these hashtags). Meaning of hashtag identifiers is explained in Table~\ref{tab:hashtagnames}. 

We performed clustering using 9 different GSC clustering methods (listed in Table~\ref{tab:GSCmethods}) on the original data as well as on data derived from the original ones by thresholding on differences between top and low similarities to amount to 0.1 and 0.2. 

There exists no direct measure to state whether or not the explainability was improved. So we proceeded using the following substitute: 
we consider that the explainability was improved if the clustering method clustered the tweets closer to the assigned hashtags. 

The Tables \ref{tab:clustrering10hashcutoff0.0F1}, \dots, \ref{tab:clustrering10hashcutoff0.2relativeerror} 
present results of our experiments about the impact on explainability when performing rough-set based removal of  boundary data.


 \begin{table}
 \footnotesize
  \begin{tabular}{|c|c|c|c|c|c|c|c|c|c|}
\hline
 & \multicolumn{9}{c|}{threshold 0.0}\\
\hline
Method&0&  1 & 2 & 3 & 4 & 5 & 6 & 7 & 8\\
\hline
{Hashtag} set & &  &  &  &  &  &  &  &  \\

0-1-2-3-4-5-6-7-8-9 & 0.161 & 0.157 & 0.161 & 0.157 & 0.502 & 0.537 & 0.538 & 0.538 & 0.537 \\ 
0-1-2-3-4-5-6-7-9 & 0.185 & 0.185 & 0.185 & 0.185 & 0.467 & 0.662 & 0.625 & 0.625 & 0.662 \\ 
0-1-2-3-4-5-7-8-9 & 0.196 & 0.203 & 0.196 & 0.203 & 0.503 & 0.533 & 0.535 & 0.535 & 0.533 \\ 
0-1-2-3-4-6-7-8-9 & 0.187 & 0.188 & 0.187 & 0.188 & 0.517 & 0.546 & 0.638 & 0.638 & 0.546 \\ 
0-1-2-3-4-6-8-9 & 0.224 & 0.226 & 0.224 & 0.226 & 0.616 & 0.647 & 0.606 & 0.606 & 0.647 \\ 
0-1-2-3-7-8-9 & 0.204 & 0.202 & 0.204 & 0.202 & 0.584 & 0.699 & 0.699 & 0.699 & 0.699 \\ 
0-1-2-4-5-6-7-8-9 & 0.188 & 0.184 & 0.188 & 0.184 & 0.471 & 0.631 & 0.593 & 0.593 & 0.631 \\ 
0-1-2-4-5-7-8-9 & 0.233 & 0.232 & 0.233 & 0.232 & 0.41 & 0.593 & 0.68 & 0.68 & 0.593 \\ 
0-1-2-4-5-8-9 & 0.296 & 0.296 & 0.296 & 0.296 & 0.501 & 0.658 & 0.67 & 0.67 & 0.658 \\ 
0-1-3-4-6-7-8 & 0.302 & 0.297 & 0.302 & 0.297 & 0.702 & 0.72 & 0.72 & 0.72 & 0.72 \\ 
0-1-4-5-7-8-9 & 0.274 & 0.268 & 0.274 & 0.268 & 0.325 & 0.348 & 0.517 & 0.517 & 0.348 \\ 
0-1-4-7-8 & 0.259 & 0.259 & 0.259 & 0.259 & 0.561 & 0.557 & 0.558 & 0.558 & 0.557 \\ 
0-2-3-4-5-6-7-9 & 0.185 & 0.185 & 0.185 & 0.185 & 0.52 & 0.662 & 0.625 & 0.625 & 0.662 \\ 
0-2-3-4-5-6-8-9 & 0.185 & 0.186 & 0.185 & 0.186 & 0.47 & 0.663 & 0.625 & 0.625 & 0.663 \\ 
0-2-4-6-8 & 0.377 & 0.367 & 0.377 & 0.367 & 0.697 & 0.692 & 0.692 & 0.692 & 0.692 \\ 
0-4-5-6-7-9 & 0.327 & 0.332 & 0.327 & 0.332 & 0.412 & 0.512 & 0.615 & 0.615 & 0.512 \\ 
0-4-5-7-8 & 0.33 & 0.339 & 0.33 & 0.339 & 0.565 & 0.625 & 0.626 & 0.626 & 0.625 \\ 
0-4-5-8 & 0.254 & 0.508 & 0.254 & 0.508 & 0.533 & 0.793 & 0.864 & 0.864 & 0.793 \\ 
1-2-3-4-5-6-7-8-9 & 0.161 & 0.157 & 0.161 & 0.157 & 0.405 & 0.537 & 0.538 & 0.538 & 0.537 \\ 
1-2-7-9 & 0.31 & 0.323 & 0.31 & 0.323 & 0.648 & 0.642 & 0.962 & 0.962 & 0.642 \\ 
1-3-4-5-6-7-8-9 & 0.193 & 0.194 & 0.193 & 0.194 & 0.473 & 0.515 & 0.478 & 0.478 & 0.515 \\ 
1-3-4-5-6-8-9 & 0.234 & 0.234 & 0.234 & 0.234 & 0.388 & 0.533 & 0.57 & 0.57 & 0.533 \\ 
1-3-4-6 & 0.502 & 0.513 & 0.502 & 0.513 & 0.938 & 0.958 & 0.958 & 0.958 & 0.958 \\ 
1-3-4-6-8-9 & 0.292 & 0.295 & 0.292 & 0.295 & 0.561 & 0.584 & 0.534 & 0.534 & 0.584 \\ 
1-3-7-9 & 0.347 & 0.361 & 0.347 & 0.361 & 0.673 & 0.672 & 0.896 & 0.896 & 0.672 \\ 
2-3-4 & 0.224 & 0.543 & 0.224 & 0.543 & 0.978 & 0.986 & 0.987 & 0.987 & 0.986 \\ 
2-3-4-5-6-9 & 0.223 & 0.225 & 0.223 & 0.225 & 0.402 & 0.671 & 0.742 & 0.742 & 0.671 \\ 
2-3-6-7-8-9 & 0.155 & 0.155 & 0.155 & 0.155 & 0.624 & 0.588 & 0.706 & 0.706 & 0.588 \\ 
2-5-6 & 0.199 & 0.2 & 0.199 & 0.2 & 0.886 & 0.937 & 0.939 & 0.939 & 0.937 \\ 
2-5-6-7-9 & 0.237 & 0.152 & 0.15 & 0.179 & 0.477 & 0.664 & 0.664 & 0.664 & 0.664 \\ 
3-4-5-6-8-9 & 0.234 & 0.234 & 0.234 & 0.234 & 0.386 & 0.533 & 0.57 & 0.57 & 0.533 \\ 
3-4-7 & 0.265 & 0.571 & 0.265 & 0.571 & 0.948 & 0.95 & 0.951 & 0.951 & 0.95 \\ 
3-4-7-8 & 0.407 & 0.407 & 0.407 & 0.407 & 0.675 & 0.674 & 0.678 & 0.678 & 0.674 \\ 
3-6-7-8-9 & 0.178 & 0.199 & 0.178 & 0.199 & 0.556 & 0.51 & 0.51 & 0.51 & 0.51 \\ 
3-7-9 & 0.347 & 0.361 & 0.347 & 0.361 & 0.673 & 0.672 & 0.896 & 0.896 & 0.672 \\ 
4-7-9 & 0.34 & 0.34 & 0.34 & 0.34 & 0.685 & 0.646 & 0.948 & 0.948 & 0.646 \\ 

\hline
  \end{tabular}%
  \caption{F1 measure.  Methods (columns): see Table~\ref{tab:GSCmethods}. 
Hashtags (rows): see Table~\ref{tab:hashtagnames}.
}
\label{tab:clustrering10hashcutoff0.0F1}
\end{table}



 \begin{table}
 \footnotesize
  \begin{tabular}{|c|c|c|c|c|c|c|c|c|c|}
\hline
 & \multicolumn{9}{c|}{threshold 0.0}\\
\hline
Method&0&  1 & 2 & 3 & 4 & 5 & 6 & 7 & 8\\
\hline
{Hashtag} set & &  &  &  &  &  &  &  &  \\

0-1-2-3-4-5-6-7-8-9 & 0.598 & 0.605 & 0.598 & 0.605 & 0.346 & 0.241 & 0.243 & 0.243 & 0.241 \\ 
0-1-2-3-4-5-6-7-9 & 0.577 & 0.577 & 0.577 & 0.577 & 0.366 & 0.189 & 0.195 & 0.195 & 0.189 \\ 
0-1-2-3-4-5-7-8-9 & 0.524 & 0.508 & 0.524 & 0.507 & 0.283 & 0.243 & 0.239 & 0.239 & 0.243 \\ 
0-1-2-3-4-6-7-8-9 & 0.568 & 0.566 & 0.568 & 0.566 & 0.275 & 0.223 & 0.176 & 0.176 & 0.223 \\ 
0-1-2-3-4-6-8-9 & 0.534 & 0.531 & 0.534 & 0.531 & 0.218 & 0.167 & 0.176 & 0.176 & 0.167 \\ 
0-1-2-3-7-8-9 & 0.535 & 0.536 & 0.535 & 0.536 & 0.233 & 0.171 & 0.169 & 0.169 & 0.171 \\ 
0-1-2-4-5-6-7-8-9 & 0.566 & 0.576 & 0.566 & 0.576 & 0.334 & 0.208 & 0.216 & 0.216 & 0.208 \\ 
0-1-2-4-5-7-8-9 & 0.477 & 0.479 & 0.477 & 0.479 & 0.359 & 0.222 & 0.179 & 0.179 & 0.222 \\ 
0-1-2-4-5-8-9 & 0.406 & 0.406 & 0.406 & 0.406 & 0.282 & 0.175 & 0.165 & 0.165 & 0.175 \\ 
0-1-3-4-6-7-8 & 0.422 & 0.433 & 0.422 & 0.433 & 0.167 & 0.132 & 0.131 & 0.131 & 0.132 \\ 
0-1-4-5-7-8-9 & 0.43 & 0.449 & 0.43 & 0.449 & 0.44 & 0.414 & 0.275 & 0.275 & 0.414 \\ 
0-1-4-7-8 & 0.363 & 0.363 & 0.363 & 0.363 & 0.175 & 0.181 & 0.18 & 0.18 & 0.181 \\ 
0-2-3-4-5-6-7-9 & 0.577 & 0.577 & 0.577 & 0.577 & 0.334 & 0.189 & 0.195 & 0.195 & 0.189 \\ 
0-2-3-4-5-6-8-9 & 0.575 & 0.574 & 0.575 & 0.574 & 0.44 & 0.183 & 0.19 & 0.19 & 0.183 \\ 
0-2-4-6-8 & 0.39 & 0.408 & 0.39 & 0.408 & 0.123 & 0.119 & 0.118 & 0.118 & 0.119 \\ 
0-4-5-6-7-9 & 0.436 & 0.426 & 0.436 & 0.426 & 0.336 & 0.306 & 0.25 & 0.25 & 0.306 \\ 
0-4-5-7-8 & 0.425 & 0.405 & 0.425 & 0.405 & 0.232 & 0.187 & 0.186 & 0.186 & 0.187 \\ 
0-4-5-8 & 0.383 & 0.27 & 0.383 & 0.27 & 0.25 & 0.125 & 0.085 & 0.085 & 0.125 \\ 
1-2-3-4-5-6-7-8-9 & 0.598 & 0.605 & 0.598 & 0.605 & 0.406 & 0.241 & 0.243 & 0.243 & 0.241 \\ 
1-2-7-9 & 0.464 & 0.457 & 0.464 & 0.457 & 0.252 & 0.218 & 0.03 & 0.03 & 0.218 \\ 
1-3-4-5-6-7-8-9 & 0.536 & 0.536 & 0.537 & 0.536 & 0.321 & 0.255 & 0.261 & 0.261 & 0.255 \\ 
1-3-4-5-6-8-9 & 0.494 & 0.494 & 0.494 & 0.494 & 0.503 & 0.265 & 0.204 & 0.204 & 0.265 \\ 
1-3-4-6 & 0.353 & 0.339 & 0.353 & 0.339 & 0.058 & 0.039 & 0.039 & 0.039 & 0.039 \\ 
1-3-4-6-8-9 & 0.44 & 0.435 & 0.44 & 0.435 & 0.225 & 0.188 & 0.199 & 0.199 & 0.188 \\ 
1-3-7-9 & 0.318 & 0.314 & 0.318 & 0.314 & 0.175 & 0.177 & 0.078 & 0.078 & 0.177 \\ 
2-3-4 & 0.571 & 0.29 & 0.571 & 0.29 & 0.022 & 0.014 & 0.013 & 0.013 & 0.014 \\ 
2-3-4-5-6-9 & 0.541 & 0.538 & 0.542 & 0.538 & 0.527 & 0.229 & 0.134 & 0.134 & 0.229 \\ 
2-3-6-7-8-9 & 0.631 & 0.63 & 0.631 & 0.63 & 0.224 & 0.218 & 0.181 & 0.181 & 0.218 \\ 
2-5-6 & 0.584 & 0.583 & 0.584 & 0.583 & 0.089 & 0.052 & 0.051 & 0.051 & 0.052 \\ 
2-5-6-7-9 & 0.625 & 0.674 & 0.675 & 0.662 & 0.377 & 0.207 & 0.207 & 0.207 & 0.207 \\ 
3-4-5-6-8-9 & 0.494 & 0.494 & 0.494 & 0.494 & 0.504 & 0.265 & 0.204 & 0.204 & 0.265 \\ 
3-4-7 & 0.504 & 0.202 & 0.504 & 0.202 & 0.036 & 0.033 & 0.031 & 0.031 & 0.033 \\ 
3-4-7-8 & 0.289 & 0.289 & 0.289 & 0.289 & 0.146 & 0.132 & 0.13 & 0.13 & 0.132 \\ 
3-6-7-8-9 & 0.558 & 0.55 & 0.558 & 0.55 & 0.261 & 0.258 & 0.258 & 0.258 & 0.258 \\ 
3-7-9 & 0.318 & 0.314 & 0.318 & 0.314 & 0.175 & 0.177 & 0.078 & 0.078 & 0.177 \\ 
4-7-9 & 0.343 & 0.343 & 0.343 & 0.343 & 0.167 & 0.159 & 0.036 & 0.036 & 0.159 \\ 

\hline
  \end{tabular}%

  \caption{Relative error.  Methods (columns): see Table~\ref{tab:GSCmethods}. 
Hashtags (rows): see Table~\ref{tab:hashtagnames}.
}
\label{tab:clustrering10hashcutoff0.0relativeerror}
\end{table}


 \begin{table}
 \footnotesize
  \begin{tabular}{|c|c|c|c|c|c|c|c|c|c|}
\hline
 & \multicolumn{9}{c|}{threshold 0.1}\\
\hline
Method&0&  1 & 2 & 3 & 4 & 5 & 6 & 7 & 8\\
\hline
{Hashtag} set & &  &  &  &  &  &  &  &  \\

0-1-2-3-4-5-6-7-8-9 & 0.225 & 0.229 & 0.225 & 0.217 & 0.428 & 0.457 & 0.431 & 0.431 & 0.457 \\ 
0-1-2-3-4-5-6-7-9 & 0.239 & 0.239 & 0.239 & 0.208 & 0.372 & 0.533 & 0.53 & 0.532 & 0.533 \\ 
0-1-2-3-4-5-7-8-9 & 0.235 & 0.292 & 0.235 & 0.292 & 0.385 & 0.4 & 0.454 & 0.454 & 0.4 \\ 
0-1-2-3-4-6-7-8-9 & 0.211 & 0.211 & 0.211 & 0.211 & 0.541 & 0.618 & 0.603 & 0.603 & 0.618 \\ 
0-1-2-3-4-6-8-9 & 0.238 & 0.206 & 0.238 & 0.206 & 0.488 & 0.574 & 0.558 & 0.558 & 0.574 \\ 
0-1-2-3-7-8-9 & 0.288 & 0.288 & 0.288 & 0.288 & 0.589 & 0.692 & 0.653 & 0.653 & 0.692 \\ 
0-1-2-4-5-6-7-8-9 & 0.307 & 0.307 & 0.307 & 0.307 & 0.448 & 0.501 & 0.618 & 0.618 & 0.501 \\ 
0-1-2-4-5-7-8-9 & 0.297 & 0.384 & 0.297 & 0.384 & 0.428 & 0.593 & 0.596 & 0.596 & 0.593 \\ 
0-1-2-4-5-8-9 & 0.385 & 0.385 & 0.281 & 0.385 & 0.359 & 0.38 & 0.382 & 0.382 & 0.38 \\ 
0-1-3-4-6-7-8 & 0.226 & 0.427 & 0.226 & 0.427 & 0.521 & 0.527 & 0.525 & 0.525 & 0.527 \\ 
0-1-4-5-7-8-9 & 0.364 & 0.363 & 0.364 & 0.363 & 0.342 & 0.358 & 0.361 & 0.361 & 0.358 \\ 
0-1-4-7-8 & 0.773 & 0.59 & 0.407 & 0.59 & 0.87 & 0.87 & 0.888 & 0.888 & 0.87 \\ 
0-2-3-4-5-6-7-9 & 0.239 & 0.239 & 0.239 & 0.239 & 0.481 & 0.533 & 0.531 & 0.532 & 0.533 \\ 
0-2-3-4-5-6-8-9 & 0.238 & 0.28 & 0.238 & 0.223 & 0.437 & 0.531 & 0.525 & 0.525 & 0.531 \\ 
0-2-4-6-8 & 0.654 & 0.657 & 0.654 & 0.657 & 0.604 & 0.7 & 0.71 & 0.71 & 0.7 \\ 
0-4-5-6-7-9 & 0.384 & 0.384 & 0.384 & 0.384 & 0.359 & 0.529 & 0.669 & 0.669 & 0.529 \\ 
0-4-5-7-8 & 0.349 & 0.56 & 0.349 & 0.56 & 0.495 & 0.554 & 0.554 & 0.554 & 0.554 \\ 
0-4-5-8 & 0.408 & 0.408 & 0.408 & 0.408 & 0.732 & 0.732 & 0.884 & 0.884 & 0.732 \\ 
1-2-3-4-5-6-7-8-9 & 0.225 & 0.224 & 0.225 & 0.202 & 0.428 & 0.457 & 0.431 & 0.431 & 0.457 \\ 
1-2-7-9 & 0.58 & 0.585 & 0.58 & 0.585 & 0.893 & 0.891 & 0.969 & 0.969 & 0.891 \\ 
1-3-4-5-6-7-8-9 & 0.247 & 0.291 & 0.247 & 0.291 & 0.362 & 0.385 & 0.433 & 0.433 & 0.385 \\ 
1-3-4-5-6-8-9 & 0.298 & 0.298 & 0.298 & 0.298 & 0.435 & 0.454 & 0.466 & 0.466 & 0.454 \\ 
1-3-4-6 & 0.634 & 0.634 & 0.634 & 0.634 & 0.624 & 0.641 & 0.641 & 0.641 & 0.641 \\ 
1-3-4-6-8-9 & 0.28 & 0.29 & 0.28 & 0.29 & 0.395 & 0.497 & 0.477 & 0.477 & 0.497 \\ 
1-3-7-9 & 0.477 & 0.537 & 0.477 & 0.537 & 0.708 & 0.709 & 0.869 & 0.869 & 0.709 \\ 
2-3-4 & 0.264 & 0.264 & 0.264 & 0.264 & 0.653 & 0.656 & 0.65 & 0.65 & 0.656 \\ 
2-3-4-5-6-9 & 0.22 & 0.22 & 0.22 & 0.22 & 0.441 & 0.622 & 0.635 & 0.635 & 0.622 \\ 
2-3-6-7-8-9 & 0.208 & 0.215 & 0.208 & 0.215 & 0.627 & 0.589 & 0.7 & 0.7 & 0.589 \\ 
2-5-6 & 0.361 & 0.368 & 0.361 & 0.368 & 0.937 & 0.956 & 0.955 & 0.955 & 0.956 \\ 
2-5-6-7-9 & 0.246 & 0.341 & 0.246 & 0.341 & 0.504 & 0.673 & 0.673 & 0.673 & 0.673 \\ 
3-4-5-6-8-9 & 0.298 & 0.298 & 0.298 & 0.298 & 0.435 & 0.454 & 0.466 & 0.466 & 0.454 \\ 
3-4-7 & 0.294 & 0.294 & 0.294 & 0.294 & 0.62 & 0.623 & 0.545 & 0.545 & 0.623 \\ 
3-4-7-8 & 0.335 & 0.335 & 0.335 & 0.335 & 0.563 & 0.601 & 0.426 & 0.426 & 0.601 \\ 
3-6-7-8-9 & 0.265 & 0.276 & 0.265 & 0.276 & 0.561 & 0.512 & 0.677 & 0.677 & 0.512 \\ 
3-7-9 & 0.477 & 0.537 & 0.477 & 0.537 & 0.708 & 0.709 & 0.869 & 0.869 & 0.709 \\ 
4-7-9 & 0.717 & 0.921 & 0.421 & 0.921 & 0.717 & 0.717 & 0.964 & 0.964 & 0.717 \\

\hline
  \end{tabular}%
  \caption{F1 measure.  Methods (columns): see Table~\ref{tab:GSCmethods}. 
Hashtags (rows): see Table~\ref{tab:hashtagnames}.
}
\label{tab:clustrering10hashcutoff0.1F1}
\end{table}

Table~\ref{tab:clustrering10hashcutoff0.0F1} contains clustering evaluation via F-measure for original data, while Table~\ref{tab:clustrering10hashcutoff0.0relativeerror} presents corresponding relative errors.  
Table~\ref{tab:clustrering10hashcutoff0.1F1} contains clustering evaluation via F-measure for removing boundary data for threshold 0.1, while Table~\ref{tab:clustrering10hashcutoff0.1relativeerror} presents corresponding relative errors.  
Table~\ref{tab:clustrering10hashcutoff0.2F1} contains clustering evaluation via F-measure for removing boundary data for threshold 0.2, while Table~\ref{tab:clustrering10hashcutoff0.2relativeerror} presents corresponding relative errors.


 \begin{table}
 \footnotesize
  \begin{tabular}{|c|c|c|c|c|c|c|c|c|c|}
\hline
 & \multicolumn{9}{c|}{threshold 0.1}\\
\hline
Method&0&  1 & 2 & 3 & 4 & 5 & 6 & 7 & 8\\
\hline
{Hashtag} set & &  &  &  &  &  &  &  &  \\
 
0-1-2-3-4-5-6-7-8-9 & 0.603 & 0.603 & 0.603 & 0.604 & 0.34 & 0.283 & 0.318 & 0.318 & 0.283 \\ 
0-1-2-3-4-5-6-7-9 & 0.57 & 0.57 & 0.57 & 0.604 & 0.417 & 0.222 & 0.226 & 0.223 & 0.222 \\ 
0-1-2-3-4-5-7-8-9 & 0.538 & 0.488 & 0.538 & 0.488 & 0.335 & 0.315 & 0.315 & 0.315 & 0.315 \\ 
0-1-2-3-4-6-7-8-9 & 0.612 & 0.613 & 0.612 & 0.613 & 0.225 & 0.171 & 0.191 & 0.191 & 0.171 \\ 
0-1-2-3-4-6-8-9 & 0.573 & 0.571 & 0.573 & 0.571 & 0.225 & 0.172 & 0.179 & 0.179 & 0.172 \\ 
0-1-2-3-7-8-9 & 0.489 & 0.489 & 0.489 & 0.489 & 0.226 & 0.171 & 0.217 & 0.217 & 0.171 \\ 
0-1-2-4-5-6-7-8-9 & 0.551 & 0.551 & 0.551 & 0.551 & 0.373 & 0.273 & 0.207 & 0.207 & 0.273 \\ 
0-1-2-4-5-7-8-9 & 0.48 & 0.413 & 0.48 & 0.413 & 0.347 & 0.209 & 0.205 & 0.205 & 0.209 \\ 
0-1-2-4-5-8-9 & 0.347 & 0.351 & 0.47 & 0.351 & 0.39 & 0.351 & 0.347 & 0.347 & 0.351 \\ 
0-1-3-4-6-7-8 & 0.523 & 0.286 & 0.523 & 0.286 & 0.197 & 0.175 & 0.18 & 0.18 & 0.175 \\ 
0-1-4-5-7-8-9 & 0.429 & 0.428 & 0.429 & 0.428 & 0.474 & 0.436 & 0.43 & 0.43 & 0.436 \\ 
0-1-4-7-8 & 0.165 & 0.254 & 0.312 & 0.254 & 0.119 & 0.119 & 0.101 & 0.101 & 0.119 \\ 
0-2-3-4-5-6-7-9 & 0.57 & 0.57 & 0.57 & 0.57 & 0.3 & 0.222 & 0.224 & 0.223 & 0.222 \\ 
0-2-3-4-5-6-8-9 & 0.567 & 0.548 & 0.567 & 0.593 & 0.322 & 0.231 & 0.224 & 0.224 & 0.231 \\ 
0-2-4-6-8 & 0.109 & 0.107 & 0.109 & 0.107 & 0.177 & 0.056 & 0.046 & 0.046 & 0.056 \\ 
0-4-5-6-7-9 & 0.356 & 0.357 & 0.356 & 0.357 & 0.391 & 0.254 & 0.142 & 0.142 & 0.254 \\ 
0-4-5-7-8 & 0.522 & 0.298 & 0.522 & 0.298 & 0.382 & 0.305 & 0.305 & 0.305 & 0.305 \\ 
0-4-5-8 & 0.31 & 0.31 & 0.31 & 0.31 & 0.263 & 0.263 & 0.106 & 0.106 & 0.263 \\ 
1-2-3-4-5-6-7-8-9 & 0.603 & 0.604 & 0.603 & 0.605 & 0.34 & 0.283 & 0.318 & 0.318 & 0.283 \\ 
1-2-7-9 & 0.296 & 0.294 & 0.296 & 0.294 & 0.041 & 0.033 & 0.012 & 0.012 & 0.033 \\ 
1-3-4-5-6-7-8-9 & 0.522 & 0.501 & 0.522 & 0.501 & 0.377 & 0.326 & 0.301 & 0.301 & 0.326 \\ 
1-3-4-5-6-8-9 & 0.466 & 0.466 & 0.466 & 0.466 & 0.3 & 0.256 & 0.246 & 0.246 & 0.256 \\ 
1-3-4-6 & 0.048 & 0.047 & 0.048 & 0.047 & 0.06 & 0.038 & 0.037 & 0.037 & 0.038 \\ 
1-3-4-6-8-9 & 0.459 & 0.448 & 0.459 & 0.448 & 0.261 & 0.2 & 0.21 & 0.21 & 0.2 \\ 
1-3-7-9 & 0.269 & 0.241 & 0.269 & 0.241 & 0.142 & 0.141 & 0.086 & 0.086 & 0.141 \\ 
2-3-4 & 0.345 & 0.345 & 0.345 & 0.345 & 0.02 & 0.015 & 0.023 & 0.023 & 0.015 \\ 
2-3-4-5-6-9 & 0.575 & 0.575 & 0.575 & 0.575 & 0.359 & 0.163 & 0.153 & 0.153 & 0.163 \\ 
2-3-6-7-8-9 & 0.603 & 0.599 & 0.603 & 0.599 & 0.219 & 0.214 & 0.185 & 0.185 & 0.214 \\ 
2-5-6 & 0.514 & 0.511 & 0.514 & 0.511 & 0.047 & 0.035 & 0.036 & 0.036 & 0.035 \\ 
2-5-6-7-9 & 0.622 & 0.545 & 0.622 & 0.545 & 0.347 & 0.19 & 0.191 & 0.191 & 0.19 \\ 
3-4-5-6-8-9 & 0.466 & 0.466 & 0.466 & 0.466 & 0.3 & 0.256 & 0.246 & 0.246 & 0.256 \\ 
3-4-7 & 0.211 & 0.211 & 0.211 & 0.211 & 0.044 & 0.042 & 0.135 & 0.135 & 0.042 \\ 
3-4-7-8 & 0.338 & 0.338 & 0.338 & 0.338 & 0.148 & 0.147 & 0.209 & 0.209 & 0.147 \\ 
3-6-7-8-9 & 0.517 & 0.509 & 0.517 & 0.509 & 0.254 & 0.253 & 0.187 & 0.187 & 0.253 \\ 
3-7-9 & 0.269 & 0.241 & 0.269 & 0.241 & 0.142 & 0.141 & 0.086 & 0.086 & 0.141 \\ 
4-7-9 & 0.174 & 0.059 & 0.273 & 0.059 & 0.174 & 0.174 & 0.028 & 0.028 & 0.174 \\ 

\hline
  \end{tabular}%

  \caption{Relative error.  Methods (columns): see Table~\ref{tab:GSCmethods}. 
Hashtags (rows): see Table~\ref{tab:hashtagnames}.
}
\label{tab:clustrering10hashcutoff0.1relativeerror}
\end{table}


 \begin{table}
 \footnotesize
  \begin{tabular}{|c|c|c|c|c|c|c|c|c|c|}
\hline
 & \multicolumn{9}{c|}{threshold 0.2}\\
\hline
Method&0&  1 & 2 & 3 & 4 & 5 & 6 & 7 & 8\\
\hline
{Hashtag} set & &  &  &  &  &  &  &  &  \\

0-1-2-3-4-5-6-7-8-9 & 0.288 & 0.316 & 0.288 & 0.316 & 0.545 & 0.528 & 0.505 & 0.505 & 0.528 \\ 
0-1-2-3-4-5-6-7-9 & 0.323 & 0.378 & 0.323 & 0.378 & 0.654 & 0.653 & 0.626 & 0.626 & 0.653 \\ 
0-1-2-3-4-5-7-8-9 & 0.38 & 0.402 & 0.38 & 0.402 & 0.534 & 0.504 & 0.46 & 0.46 & 0.504 \\ 
0-1-2-3-4-6-7-8-9 & 0.308 & 0.307 & 0.308 & 0.307 & 0.542 & 0.497 & 0.495 & 0.495 & 0.497 \\ 
0-1-2-3-4-6-8-9 & 0.222 & 0.371 & 0.222 & 0.22 & 0.508 & 0.515 & 0.443 & 0.443 & 0.515 \\ 
0-1-2-3-7-8-9 & 0.721 & 0.715 & 0.721 & 0.715 & 0.725 & 0.719 & 0.656 & 0.656 & 0.719 \\ 
0-1-2-4-5-6-7-8-9 & 0.445 & 0.446 & 0.445 & 0.446 & 0.581 & 0.649 & 0.602 & 0.603 & 0.648 \\ 
0-1-2-4-5-7-8-9 & 0.536 & 0.537 & 0.536 & 0.537 & 0.749 & 0.748 & 0.707 & 0.707 & 0.748 \\ 
0-1-2-4-5-8-9 & 0.733 & 0.733 & 0.733 & 0.733 & 0.721 & 0.724 & 0.672 & 0.672 & 0.724 \\ 
0-1-3-4-6-7-8 & 0.345 & 0.344 & 0.345 & 0.344 & 0.538 & 0.549 & 0.543 & 0.543 & 0.549 \\ 
0-1-4-5-7-8-9 & 0.731 & 0.731 & 0.731 & 0.731 & 0.718 & 0.724 & 0.682 & 0.682 & 0.724 \\ 
0-1-4-7-8 & -- & -- & 0.491 & 0.491 & 0.491 & 0.491 & 0.491 & 0.491 & 0.491 \\ 
0-2-3-4-5-6-7-9 & 0.323 & 0.378 & 0.323 & 0.378 & 0.654 & 0.653 & 0.626 & 0.626 & 0.653 \\ 
0-2-3-4-5-6-8-9 & 0.383 & 0.369 & 0.383 & 0.369 & 0.504 & 0.489 & 0.453 & 0.453 & 0.489 \\ 
0-2-4-6-8 & 0.725 & 0.724 & 0.725 & 0.724 & 0.49 & 0.491 & 0.49 & 0.49 & 0.491 \\ 
0-4-5-6-7-9 & 0.549 & 0.549 & 0.549 & 0.549 & 0.774 & 0.775 & 0.722 & 0.722 & 0.775 \\ 
0-4-5-7-8 & 0.645 & 0.654 & 0.645 & 0.654 & 0.631 & 0.64 & 0.654 & 0.654 & 0.64 \\ 
0-4-5-8 & 0.493 & 0.493 & 0.493 & 0.493 & 0.493 & 0.493 & 0.493 & 0.493 & 0.493 \\ 
1-2-3-4-5-6-7-8-9 & 0.288 & 0.316 & 0.288 & 0.316 & 0.545 & 0.528 & 0.505 & 0.505 & 0.528 \\ 
1-2-7-9 & -- & -- & -- & -- & -- & -- & 0.884 & 0.884 & 0.884 \\ 
1-3-4-5-6-7-8-9 & 0.321 & 0.258 & 0.259 & 0.258 & 0.508 & 0.477 & 0.443 & 0.443 & 0.477 \\ 
1-3-4-5-6-8-9 & 0.321 & 0.321 & 0.321 & 0.321 & 0.469 & 0.433 & 0.393 & 0.393 & 0.433 \\ 
1-3-4-6 & 0.653 & 0.653 & 0.653 & 0.653 & 0.647 & 0.647 & 0.649 & 0.649 & 0.647 \\ 
1-3-4-6-8-9 & 0.502 & 0.479 & 0.502 & 0.479 & 0.467 & 0.474 & 0.354 & 0.354 & 0.474 \\ 
1-3-7-9 & 0.979 & 0.979 & 0.979 & 0.979 & 0.659 & 0.659 & 0.659 & 0.659 & 0.659 \\ 
2-3-4 & 0.995 & 0.992 & 0.707 & 0.992 & 0.96 & 0.96 & 0.989 & 0.989 & 0.96 \\ 
2-3-4-5-6-9 & 0.428 & 0.428 & 0.428 & 0.428 & 0.627 & 0.628 & 0.592 & 0.592 & 0.628 \\ 
2-3-6-7-8-9 & 0.195 & 0.197 & 0.195 & 0.195 & 0.689 & 0.695 & 0.608 & 0.608 & 0.695 \\ 
2-5-6 & 0.576 & 0.943 & 0.576 & 0.943 & 0.971 & 0.98 & 0.98 & 0.98 & 0.98 \\ 
2-5-6-7-9 & 0.506 & 0.719 & 0.506 & 0.719 & 0.971 & 0.969 & 0.927 & 0.927 & 0.969 \\ 
3-4-5-6-8-9 & 0.321 & 0.321 & 0.321 & 0.321 & 0.469 & 0.433 & 0.393 & 0.393 & 0.433 \\ 
3-4-7 & 0.987 & 0.985 & 0.459 & 0.985 & 0.961 & 0.961 & 0.978 & 0.978 & 0.961 \\ 
3-4-7-8 & 0.612 & 0.862 & 0.612 & 0.862 & 0.854 & 0.85 & 0.594 & 0.594 & 0.85 \\ 
3-6-7-8-9 & 0.493 & 0.609 & 0.493 & 0.609 & 0.72 & 0.722 & 0.55 & 0.55 & 0.722 \\ 
3-7-9 & 0.979 & 0.979 & 0.979 & 0.979 & 0.659 & 0.659 & 0.659 & 0.659 & 0.659 \\ 
4-7-9 & -- & 0.815 & 0.721 & 0.815 & 0.815 & 0.815 & 0.815 & 0.815 & 0.815 \\

\hline
  \end{tabular}%
  \caption{F1 measure. Methods (columns): see Table~\ref{tab:GSCmethods}. 
Hashtags (rows): see Table~\ref{tab:hashtagnames}.
}
\label{tab:clustrering10hashcutoff0.2F1}
\end{table}


 \begin{table}
 \footnotesize
  \begin{tabular}{|c|c|c|c|c|c|c|c|c|c|}
\hline
 & \multicolumn{9}{c|}{threshold 0.2}\\
\hline
Method&0&  1 & 2 & 3 & 4 & 5 & 6 & 7 & 8\\
\hline
{Hashtag} set & &  &  &  &  &  &  &  &  \\

0-1-2-3-4-5-6-7-8-9 & 0.575 & 0.507 & 0.575 & 0.507 & 0.253 & 0.237 & 0.224 & 0.224 & 0.237 \\ 
0-1-2-3-4-5-6-7-9 & 0.563 & 0.458 & 0.563 & 0.458 & 0.168 & 0.155 & 0.165 & 0.165 & 0.155 \\ 
0-1-2-3-4-5-7-8-9 & 0.358 & 0.341 & 0.358 & 0.341 & 0.207 & 0.219 & 0.241 & 0.241 & 0.219 \\ 
0-1-2-3-4-6-7-8-9 & 0.452 & 0.451 & 0.452 & 0.451 & 0.21 & 0.209 & 0.21 & 0.21 & 0.209 \\ 
0-1-2-3-4-6-8-9 & 0.57 & 0.39 & 0.57 & 0.57 & 0.184 & 0.163 & 0.191 & 0.191 & 0.163 \\ 
0-1-2-3-7-8-9 & 0.11 & 0.112 & 0.11 & 0.112 & 0.113 & 0.125 & 0.137 & 0.137 & 0.125 \\ 
0-1-2-4-5-6-7-8-9 & 0.334 & 0.332 & 0.334 & 0.332 & 0.158 & 0.146 & 0.12 & 0.119 & 0.147 \\ 
0-1-2-4-5-7-8-9 & 0.289 & 0.287 & 0.289 & 0.287 & 0.086 & 0.072 & 0.086 & 0.086 & 0.072 \\ 
0-1-2-4-5-8-9 & 0.044 & 0.044 & 0.044 & 0.044 & 0.058 & 0.051 & 0.068 & 0.068 & 0.051 \\ 
0-1-3-4-6-7-8 & 0.251 & 0.255 & 0.251 & 0.255 & 0.149 & 0.123 & 0.135 & 0.135 & 0.123 \\ 
0-1-4-5-7-8-9 & 0.059 & 0.059 & 0.059 & 0.059 & 0.082 & 0.072 & 0.086 & 0.086 & 0.072 \\ 
0-1-4-7-8 & -- & -- & 0.037 & 0.037 & 0.037 & 0.037 & 0.037 & 0.037 & 0.037 \\ 
0-2-3-4-5-6-7-9 & 0.563 & 0.458 & 0.563 & 0.458 & 0.168 & 0.155 & 0.165 & 0.165 & 0.155 \\ 
0-2-3-4-5-6-8-9 & 0.445 & 0.465 & 0.445 & 0.465 & 0.252 & 0.222 & 0.223 & 0.223 & 0.222 \\ 
0-2-4-6-8 & 0.01 & 0.012 & 0.01 & 0.012 & 0.034 & 0.032 & 0.033 & 0.033 & 0.032 \\ 
0-4-5-6-7-9 & 0.241 & 0.241 & 0.241 & 0.241 & 0.027 & 0.024 & 0.056 & 0.056 & 0.024 \\ 
0-4-5-7-8 & 0.052 & 0.039 & 0.052 & 0.039 & 0.073 & 0.06 & 0.039 & 0.039 & 0.06 \\ 
0-4-5-8 & 0.029 & 0.029 & 0.029 & 0.029 & 0.029 & 0.029 & 0.029 & 0.029 & 0.029 \\ 
1-2-3-4-5-6-7-8-9 & 0.575 & 0.507 & 0.575 & 0.507 & 0.253 & 0.237 & 0.224 & 0.224 & 0.237 \\ 
1-2-7-9 & -- & -- & -- & -- & -- & -- & 0.068 & 0.068 & 0.068 \\ 
1-3-4-5-6-7-8-9 & 0.482 & 0.518 & 0.518 & 0.518 & 0.263 & 0.264 & 0.25 & 0.25 & 0.264 \\ 
1-3-4-5-6-8-9 & 0.463 & 0.463 & 0.463 & 0.463 & 0.244 & 0.238 & 0.239 & 0.239 & 0.238 \\ 
1-3-4-6 & 0.02 & 0.02 & 0.02 & 0.02 & 0.028 & 0.028 & 0.025 & 0.025 & 0.028 \\ 
1-3-4-6-8-9 & 0.137 & 0.153 & 0.137 & 0.153 & 0.174 & 0.151 & 0.179 & 0.179 & 0.151 \\ 
1-3-7-9 & 0.004 & 0.004 & 0.004 & 0.004 & 0.035 & 0.035 & 0.035 & 0.035 & 0.035 \\ 
2-3-4 & 0.005 & 0.007 & 0.258 & 0.007 & 0.037 & 0.037 & 0.01 & 0.01 & 0.037 \\ 
2-3-4-5-6-9 & 0.413 & 0.412 & 0.413 & 0.412 & 0.156 & 0.141 & 0.154 & 0.154 & 0.141 \\ 
2-3-6-7-8-9 & 0.669 & 0.669 & 0.669 & 0.669 & 0.143 & 0.125 & 0.141 & 0.141 & 0.125 \\ 
2-5-6 & 0.329 & 0.057 & 0.329 & 0.057 & 0.021 & 0.014 & 0.014 & 0.014 & 0.014 \\ 
2-5-6-7-9 & 0.485 & 0.211 & 0.485 & 0.211 & 0.025 & 0.022 & 0.033 & 0.033 & 0.022 \\ 
3-4-5-6-8-9 & 0.463 & 0.463 & 0.463 & 0.463 & 0.244 & 0.238 & 0.239 & 0.239 & 0.238 \\ 
3-4-7 & 0.013 & 0.015 & 0.445 & 0.015 & 0.039 & 0.039 & 0.022 & 0.022 & 0.039 \\ 
3-4-7-8 & 0.129 & 0.074 & 0.129 & 0.074 & 0.082 & 0.087 & 0.161 & 0.161 & 0.087 \\ 
3-6-7-8-9 & 0.29 & 0.236 & 0.29 & 0.236 & 0.107 & 0.101 & 0.128 & 0.128 & 0.101 \\ 
3-7-9 & 0.004 & 0.004 & 0.004 & 0.004 & 0.035 & 0.035 & 0.035 & 0.035 & 0.035 \\ 
4-7-9 & -- & 0.16 & 0.15 & 0.16 & 0.16 & 0.16 & 0.16 & 0.16 & 0.16 \\ 

\hline
  \end{tabular}%

  \caption{Relative error. Methods (columns): see Table~\ref{tab:GSCmethods}. 
Hashtags (rows): see Table~\ref{tab:hashtagnames}.
}
\label{tab:clustrering10hashcutoff0.2relativeerror}
\end{table}

Upon increase of threshold from 0.0 to 0.1,  176  out of  324  cases (54\%) experience an error reduction.
Upon increase of threshold from 0.0 to 0.2,  274  out of  315  cases (87\%) experience an error reduction, confirming that the applied boundary removal improves the clustering capability for a range of clustering methods. 

In the cases of individual clustering methods, these improvements differ (see Table~\ref{tab:clustreringRelErrImprovementMethod}).

 \begin{table}
 \footnotesize
  \begin{tabular}{|c|c|c|c|c|c|c|c|c|c|}
\hline
Method&0&  1 & 2 & 3 & 4 & 5 & 6 & 7 & 8\\
\hline
Threshold & &  &  &  &  &  &  &  &  \\
\hline
0.1  & 69\%& 78\% &67\% & 72\%  & 53\% & 47\% &  28\% & 28\% &  47\%  \\
0.2  & 94\%& 97\% &94\% & 94\%  & 94\% & 86\% &  69\% & 69\% &  86\%  \\
\hline
  \end{tabular}%
  \caption{Percentage of cases of improvement of relative error for a given threshold, compared to the original data.  Methods (columns): see Table~\ref{tab:GSCmethods}. 
}
\label{tab:clustreringRelErrImprovementMethod}
\end{table}

In case of the method ``Combinatorial spectral clustering -- unit length rows and one additional dimension used'',   
upon increase of threshold from 0.0 to 0.1,  28  out of  36  cases (78\%) experience an error reduction.
Upon increase of threshold from 0.0 to 0.2,  33  out of  34  cases (97\%) experience an error reduction. It was the best result.

In case of the method ``Normalized spectral clustering -- unit length rows and one additional dimension used'', upon increase of threshold from 0.0 to 0.1,  10  out of  36  cases (28\%) experience an error reduction.
Upon increase of threshold from 0.0 to 0.2,  25  out of  36  cases (69\%) experience an error reduction. It was the worst result. Note, however, that the method performed very well for original data and there was not too much space left for improvements. 

We have also investigated whether or not the improvements are impacted by the number of hashtags considered (see Table~\ref{tab:clustreringRelErrImprovementNumHash})

 \begin{table}
 \footnotesize
  \begin{tabular}{|c|c|c|c|c|c|c|c|}
\hline
No of Hashtags&3&  4 & 5 & 6 & 7 & 8 & 9 \\
\hline
Threshold & &  &  &  &  &  &    \\
\hline
0.1  & 60\%&58\%& 82\%    & 56\% & 51\% & 40\% &  38\% \\
0.2  & 77\%&90\%& 100\%    & 82\% & 91\% & 76\%  & 91\%   \\
\hline
  \end{tabular}%
  \caption{Percentage of cases of improvement of relative error for a given threshold, compared to the original data.  Columns state how many hashtags were used in the dataset. 
}
\label{tab:clustreringRelErrImprovementNumHash}
\end{table}

\begin{table}
 \footnotesize
  \begin{tabular}{|c|c|c|c|}
    \hline
     Initial hashtag set & \multicolumn{3}{c|}{hashtag set included in the experiment}  \\
     \hline
     & threshold 0.0 &  threshold 0.1 & threshold 0.2 \\
    \hline
0-1-2-3-4-5-6-7-8-9	&	2-3-4-5-6-7-8-9	&	2-3-4-5-6-7-8-9	&	2-3-4-5-6-7-8-9	\\
0-1-2-3-4-5-6-7-9	&	2-3-4-5-6-7-9	&	2-3-4-5-6-7-9	&	2-3-4-5-6-7-9	\\
0-1-2-3-4-5-7-8-9	&	2-3-4-5-7-8-9	&	2-3-4-5-7-8-9	&	2-3-4-5-7-8-9	\\
0-1-2-3-4-6-7-8-9	&	2-3-4-6-7-8-9	&	2-3-4-6-7-8-9	&	2-3-4-6-7-8-9	\\
0-1-2-3-4-6-8-9	&	2-3-4-6-8-9	&	2-3-4-6-8-9	&	2-3-4-6-8-9	\\
0-1-2-3-7-8-9	&	2-3-7-8-9	&	2-3-7-8-9	&	2-3-7-8-9	\\
0-1-2-4-5-6-7-8-9	&	2-4-5-6-7-8-9	&	2-4-5-6-7-8-9	&	2-4-5-6-7-8-9	\\
0-1-2-4-5-7-8-9	&	2-4-5-7-8-9	&	2-4-5-7-8-9	&	2-5-7-8-9	\\
0-1-2-4-5-8-9	&	2-4-5-8-9	&	2-4-5-8-9	&	2-5-8-9	\\
0-1-3-4-6-7-8	&	3-4-6-7-8	&	3-4-6-7-8	&	3-4-6-7-8	\\
0-1-4-5-7-8-9	&	4-5-7-8-9	&	4-5-7-8-9	&	5-7-8-9	\\
0-1-4-7-8	&	4-7-8	&	7-8	&	7-8	\\
0-2-3-4-5-6-7-9	&	2-3-4-5-6-7-9	&	2-3-4-5-6-7-9	&	2-3-4-5-6-7-9	\\
0-2-3-4-5-6-8-9	&	2-3-4-5-6-8-9	&	2-3-4-5-6-8-9	&	2-3-4-5-6-8-9	\\
0-2-4-6-8	&	2-4-6-8	&	2-4-6-8	&	2-4-6-8	\\
0-4-5-6-7-9	&	4-5-6-7-9	&	4-5-6-7-9	&	4-5-6-7-9	\\
0-4-5-7-8	&	4-5-7-8	&	5-7-8	&	5-7-8	\\
0-4-5-8	&	4-5-8	&	5-8	&	5-8	\\
1-2-3-4-5-6-7-8-9	&	2-3-4-5-6-7-8-9	&	2-3-4-5-6-7-8-9	&	2-3-4-5-6-7-8-9	\\
1-2-7-9	&	2-7-9	&	2-7-9	&	2-7-9	\\
1-3-4-5-6-7-8-9	&	3-4-5-6-7-8-9	&	3-4-5-6-7-8-9	&	3-4-5-6-7-8-9	\\
1-3-4-5-6-8-9	&	3-4-5-6-8-9	&	3-4-5-6-8-9	&	3-4-5-6-8-9	\\
1-3-4-6-8-9	&	3-4-6-8-9	&	3-4-6-8-9	&	3-4-6-8-9	\\
1-3-4-6	&	3-4-6	&	3-4-6	&	3-4-6	\\
1-3-7-9	&	3-7-9	&	3-7-9	&	3-7-9	\\
2-3-4-5-6-9	&	all	&	all	&	all	\\
2-3-4	&	all	&	all	&	2-3	\\
2-3-6-7-8-9	&	all	&	all	&	all	\\
2-5-6-7-9	&	all	&	all	&	all	\\
2-5-6	&	all	&	all	&	all	\\
3-4-5-6-8-9	&	all	&	all	&	all	\\
3-4-7-8	&	all	&	all	&	3-7-8	\\
3-4-7	&	all	&	all	&	3-7	\\
3-6-7-8-9	&	all	&	all	&	all	\\
3-7-9	&	all	&	all	&	all	\\
4-7-9	&	all	&	 7-9	&	 7-9	\\

    \hline
  \end{tabular}%
  \caption{Hashtags included in the experiment.} 
\label{tab:hashtagreduction}
\end{table}


One can see that starting with 5 hashtags, for threshold 0.1, the improvement is systematically decreasing, indicating that more hashtags require more sharpness in differentiation between hashtags. 
Threshold 0.2 seems to be quite effective.

It shall be noted, however, that the core extraction led in the experiments also to  removal of some original clusters. This is summarized in Table~\ref{tab:hashtagreduction}. The entry ``all'' means that all original clusters are represented after data reduction. Otherwise, the clusters (hashtags) that are represented after reduction, are listed. Clusters 0 and 1 (hashtags  \texttt{\#1} and \texttt{\#90dayfiance}) seem to be removed most frequently. A visual inspection of the tweet texts seems to confirm that they do not have a clear content. 
One can say that in this case, originally  there existed clusters that consist only of the upper approximation, but their lower approximation was in fact empty.

\subsection{Discussion}\label{sec:discexperiments}
The experiments that have been performed appear to support the idea that the explainability of clusters can be improved using the idea of rough set theory inspired removal of boundary documents. It appears to be valid for various brands of GSC algorithms and for various number of clusters.
\FloatBarrier

\section{Conclusions}\label{sec:conclusions}

The paper contributes to the field of data clustering by combining Rough Set Theory with Spectral Graph Clustering to enhance the explainability of the clustering results, making it easier to interpret and apply these results in practical scenarios.
This is achieved by differentiating between core and boundary elements in clusters in a  way independent of a particular clustering, relying only on general properties of GSC.
This leads to clearer explanations for cluster membership by filtering out  ambiguous cases from cluster description. 
    A  more transparent and interpretable clustering process is achieved. 
This approach effectively bridges the gap between the mathematical rigor of Spectral Graph Clustering and the need for understandable, actionable results in real-world applications.

The broad set of methods of Graph Spectral Clustering, until recently, has been considered as black-box clustering methods, without any possibility of explanation of clustering results. 
Only recently, some methods of explanation have been proposed, for example in  {\cite{Plosone2025}}. 
However, under practical settings, the GSC methods produced high discrepancy between the results. 
In this paper, we proposed a new approach to explainability, or rather an amendment to the just mentioned proposals. The amendment consists in removing a part of data with low discernability of cluster membership versus non-membership and concentrate only on the high discernability elements. The explanation to these clusters will be the explainability foundation. In this way, various clustering methods get closer to one another in their results so that finally the intersection between them can be considered a final explanation.

The proposal has several crucial advantages over the previous methodologies. 
It is not a new idea to to distinguish between cluster core elements and boundary elements. Many publications adhere to this distinction. However, this split in those papers is closely related to the clustering algorithm itself. As a consequence, a document is rejected as non-core element not due to some objective criterion, but rather as non-fitting some clustering methodology. We propose instead to use a clustering method independent approach, that results from common sense: documents that are to the same extent similar to all other objects, do not constitute a reasonable cluster element.
The next problem, inherent with the methodologies proposed in the literature, is the neglecting of the need for cluster explanations. In many cases, even explainable algorithms, when adapted to the problem of boundary elements, loose their capability of explaining clearly why elements belong to a given cluster. In our methodology, this problem is avoided in a natural way. Explainable algorithms remain explainable when amended to deal with cluster core elements and boundary elements. 

Last but not least the issue of algorithm convergence was studied. Frequently, noise in the data is addressed by seeking intersection of results of various clustering methods. A simplistic application of such an approach may be disastrous as various algorithms have different built-in clustering target functions. Only those algorithms can be applied in this context that are able to detect same clustering when the data is ideally separated, that is without boundary elements. We have demonstrated that a number of GSC algorithms have this convergence property, and  separate publications ({\cite{Plosone2025}}) demonstrate proofs of equivalence or near equivalence.

The aforementioned convergence property could be used to create an ensemble clustering method consisting of multiple algorithms converging to a common clustering solution. We do not, however, use such an approach, because the explainability property can be easily lost in the process. From explainability point of view, a better idea is to use a couple of convergent algorithms to obtain several similar clusterings, then to get their intersection (if large enough), to reject documents not shared by the clusters and to re-cluster the remaining documents by \textit{one} explainable clustering algorithm to have a valid explanation of clustering of these remaining documents, being the core of clustering.  

{
This paper not only provides with an improvement of clustering and cluster explanation n the concrete case of GSA, but has also some general conceptual consequences. 
Many methods of improvement of clustering results of a given clustering algorithm, in particular based on dropping or correcting data points, are based on the clustering criterion of this algorithm itself. Although the clustering is improved then, but it becomes simply questionable whether or not the cluster revealed real clusters in the data. In fact, such algorithms impose predefined clusters - clusters immanent to the algorithm definition. We withstand here this tendency in that we refer to the general concept of clusters themselves - we remove documents that will most probably not have the property of belonging to any cluster without referring to the result of clustering via a particular clustering algorithm. 
A second contribution is the explicit reference to the clustering model for which GSA are defined. GSA aims generally at discovering clusters that form a block-diagonal similarity matrix. This concept is the foundation of element removal and not the result of clustering via any concrete algorithm. 
Third, we demonstrated in a qualitative manner that such a sharpening of the data set in fact improves the explainability of the clustering.
}

Further research is clearly necessary to check the correctness of some of our assumptions. We took the 5\% limits on the percentage of similarities that are compared. This assumption can be associated with typical approaches to outlier identification. One can still ask whether to take an average or the median of those intervals, whether or not the interval boundaries should depend on the number of envisaged clusters and how such a dependence should look like.



\bibliographystyle{elsarticle-num}
\bibliography{nasza_bib,cudza_bib}

@article{Masson:2008:ECM,
author = {Marie-Hélène Masson and T. Denœux},
title = {{ECM}: An evidential version of the fuzzy c-means algorithm},
journal = {Pattern Recognition},
volume = {41},
number = {4},
pages = {1384-1397},
year = {2008},
issn = {0031-3203},
doi = {https://doi.org/10.1016/j.patcog.2007.08.014},
}

@ARTICLE{Jain2010,
  AUTHOR =       {A. Jain},
  TITLE =        {Data clustering: 50 years beyond $k$-means},
  JOURNAL =      {Pattern Recognition Letters},
  YEAR =         {2010},
  volume =       {31},
  number =       {8},
  pages =        {651-666},
  doi =          {https://doi.org/10.1016/j.patrec.2009.09.011}
}

@article{Maji2007,
  title={Rough set based generalized fuzzy $ C $-means algorithm and quantitative indices},
  author={Maji, Pradipta and Pal, Sankar K},
  journal={IEEE Transactions on Systems, Man, and Cybernetics, Part B (Cybernetics)},
  volume={37},
  number={6},
  pages={1529-1540},
  year={2007},
  doi={https://doi.org/10.1109/TSMCB.2007.906578},
  publisher={IEEE}
}

@article{Pawlak1985,
title = {Rough sets and fuzzy sets},
journal = {Fuzzy Sets and Systems},
volume = {17},
number = {1},
pages = {99-102},
year = {1985},
doi = {https://doi.org/10.1016/S0165-0114(85)80029-4},
author = {Zdzisław Pawlak},
}

@ARTICLE{Dubois1990,
  author =       {D. Dubois and H. Prade},
  title =        {Rough fuzzy sets and fuzzy rough sets},
  journal =      {International Journal of General Systems},
  year =         {1990},
  volume =       {17},
  number =       {2},
  pages =        {91-209},
  doi =          {https://doi.org/10.1080/03081079008935107},
}

@Article{Antoine:2012:CECM, 
  author={Antoine, V. and Quost, B. and Masson, M.-H. and Den{\oe}ux, T.},
  title={{CECM: Constrained evidential C-means algorithm}},
  journal={Computational Statistics \& Data Analysis},
  year=2012,
  volume={56},
  number={4},
  pages={894-914},
  month={},
  doi={10.1016/j.csda.2010.09.02},
  myurl={https://ideas.repec.org/a/eee/csdana/v56y2012i4p894-914.html},
myurl2={
https://www.sciencedirect.com/science/article/pii/S0167947310003646
},
otherpotientallyinterestinfgreferences={
V. Antoine, B. Quost, M.-H. Masson and T. Denoeux. CECM: Constrained Evidential C-Means algorithm. Computational Statistics and Data Analysis, Vol. 56, Issue 4, pages 894–914, 2012.

T. Denoeux and M.-H. Masson. EVCLUS: Evidential Clustering of Proximity Data. IEEE Transactions on Systems, Man and Cybernetics B, Vol. 34, Issue 1, 95–109, 2004.

T. Denoeux, O. Kanjanatarakul and S. Sriboonchitta. EK-NNclus: a clustering procedure based on the evidential K-nearest neighbor rule. Knowledge-Based Systems, Vol. 88, pages 57–69, 2015.

T. Denoeux, S. Sriboonchitta and O. Kanjanatarakul. Evidential clustering of large dissimilarity data. Knowledge-Based Systems, vol. 106, pages 179-195, 2016.

M.-H. Masson and T. Denoeux. ECM: An evidential version of the fuzzy c-means algorithm. Pattern Recognition, Vol. 41, Issue 4, pages 1384–1397, 2008.

M.-H. Masson and T. Denoeux. RECM: Relational Evidential c-means algorithm. Pattern Recognition Letters, Vol. 30, pages 1015–1026, 2009.

V. Antoine, B. Quost, M.-H. Masson and T. Denoeux. CEVCLUS: Evidential clustering with instance-level constraints for relational data. Soft Computing 18(7):1321-1335, 2014.

F. Li, S. Li and T. Denoeux. k-CEVCLUS: Constrained evidential clustering of large dissimilarity data. Knowledge-Based Systems 142:29-44, 2018.

Z.-G. Su and T. Denoeux. BPEC: Belief-Peaks Evidential Clustering. IEEE Transactions on Fuzzy Systems, 27(1):111-123, 2019.

T. Denoeux. NN-EVCLUS: Neural Network-based Evidential Clustering. arXiv:2009.12795, 2020a.

T. Denoeux. Calibrated model-based evidential clustering using bootstrapping. Information Sciences, Vol. 528, pages 17-45, 2020b. 
}
}

@inproceedings{Whang:2015:nonexhaus,
  author = "Joyce Whang AND Inderjit S. Dhillon AND David Gleich",
  title = "Non-exhaustive, Overlapping k-means",
  booktitle = "SIAM International Conference on Data Mining (SDM)",
  pages = "936—-944",
  year = "2015",
  month = "may",
  doi = {https://doi.org/10.1109/TPAMI.2018.2863278}
, myurl={https://www.cs.utexas.edu/~inderjit/public_papers/neo_kmeans_sdm15.pdf}
}

@article{vonLuxburg:2007
, title={A Tutorial on Spectral Clustering}
, author={U. von Luxburg},
  JOURNAL =      {Statistics and Computing},
  YEAR =         {2007},
  volume =       {17},
  number =       {4},
  pages =        {395-416},
  doi =          {https://doi.org/10.1007/s11222-007-9033-z}
}

@article{TU:2022:3673,
title = {An improved {N}ystr\"{o}m spectral graph clustering using k-core decomposition as a sampling strategy for large networks},
journal = {Journal of King Saud University - Computer and Information Sciences},
volume = {34},
number = {6, Part B},
pages = {3673-3684},
year = {2022},
issn = {1319-1578},
doi = {https://doi.org/10.1016/j.jksuci.2022.04.009},
Xurl = {https://www.sciencedirect.com/science/article/pii/S1319157822001379},
author = {Jingzhi Tu and Gang Mei and Francesco Piccialli},
}

@article{Gower:1966
, author={J.C. Gower}
, title={Some distance properties of latent root and vector methods used in multivariate analysis}
, journal={Biometrika}
, volume={53(3-4)}
, pages={325–338}
, year= 1966
, doi={https://doi.org/10.1093/biomet/53.3-4.325}
}

@article{Schaeffer,
author = {Satu Elisa Schaeffer},
title = {Graph clustering},
journal = {Computer Science Review},
volume = {1},
number = {1},
pages = {27-64},
year = {2007},
doi = {https://doi.org/10.1016/j.cosrev.2007.05.001},
}

@BOOK{Gol96,
  author =        {Golub, Gene H. and Van Loan, Charles F.},
  TITLE =        {Matrix Computation. Third edition},
  PUBLISHER =    {Hindustan Book Agency},
  YEAR =         {1996},
  address =      {New Delhi, India},
}

@ARTICLE{Paw82,
  author =       {Z. Pawlak},
  title =        {Rough sets},
  journal =      {International Journal of Computer and Information Sciences},
  year =         {1982},
  volume =       {11},
  number =       {},
  pages =        {341-356},
  doi =          {https://doi.org/10.1007/BF01001956},
}

@INCOLLECTION{Bello17,
  author =       {Rafael Bello and Rafael Falcon},
  title =        {Rough Sets in Machine Learning: A Review},
  booktitle =    {Thriving Rough Sets},
  publisher =    {Springer, Cham},
  year =         {2017},
  editor =       {G. Wang and A. Skowron and Y. Yao and D. \'{S}l{\c{e}}zak and L. Polkowski},
  volume =       {708},
  series =       {Studies in Computational Intelligence},
  chapter =      {},
  pages =        {87-118},
doi={10.1007/978-3-319-54966-8\_5}
}

@ARTICLE{Pet2015,
  author =       {G. Peters},
  title =        {Is there any need for rough clustering?},
  journal =      {Pattern Recognition Letters},
  year =         {2015},
  volume =       {53},
  number =       {},
  pages =        {31-37},
  doi =          {https://doi.org/10.1016/j.patrec.2014.11.003},
}

@article{Lingras2011,
  title={Rough clustering},
  author={Lingras, Pawan and Peters, Georg},
  journal={Wiley Interdisciplinary Reviews: Data Mining and Knowledge Discovery},
  volume={1},
  number={1},
  pages={64--72},
  year={2011},
  publisher={Wiley Online Library},
  doi= {https://doi.org/10.1002/widm.16}
}

@ARTICLE{Peters2006,
  author =       {G. Peters},
  title =        {Some refinements of rough $k$-means clustering},
  journal =      {Pattern Recognition},
  year =         {2006},
  volume =       {39},
  number =       {8},
  pages =        {1481-1491},
  doi =          {https://doi.org/10.1016/j.patcog.2006.02.002},
}

@InProceedings{Peters2013,
author="Peters, Georg
and Crespo, Fernando",
editor="Ciucci, Davide
and Inuiguchi, Masahiro
and Yao, Yiyu
and {\'{S}}l{\k{e}}zak, Dominik
and Wang, Guoyin",
title="An Illustrative Comparison of Rough k-Means to Classical Clustering Approaches",
booktitle="Rough Sets, Fuzzy Sets, Data Mining, and Granular Computing",
year="2013",
publisher="Springer Berlin Heidelberg",
address="Berlin, Heidelberg",
pages="337--344",
doi={https://doi.org/10.1007/978-3-642-41218-9\_36}
}

@InProceedings{Manish2010,
author="Joshi, Manish
and Lingras, Pawan
and Rao, C. Raghavendra",
editor="Yu, Jian
and Greco, Salvatore
and Lingras, Pawan
and Wang, Guoyin
and Skowron, Andrzej",
title="Analysis of Rough and Fuzzy Clustering",
booktitle="Rough Set and Knowledge Technology",
year="2010",
publisher="Springer",
address="Berlin, Heidelberg",
pages="679--686",
doi = "https://doi.org/10.1007/978-3-642-16248-0\_92",
}

@BOOK{Bezdek81,
  author =       {J.C. Bezdek},
  title =        {Pattern recognition with fuzzy objective function algorithms},
  publisher =    {Plenum Press},
  year =         {1981},
  series =       {Advanced Applications in Pattern Recognition},
  address =      {New York and London},
  doi =          {https://doi.org/10.1007/978-1-4757-0450-1},
}

@ARTICLE{Lingras2004,
  author =       {P. Lingras and C. West},
  title =        {Interval Set Clustering of {W}eb Users with Rough K-Means},
  journal =      {Journal of Intelligent Information Systems, 23:1, 5–16, 2004},
  year =         {2004},
  volume =       {23},
  number =       {1},
  pages =        {5-16},
  doi =          {https://doi.org/10.1023/B:JIIS.0000029668.88665.1a},
}

@article{YU2024,
title = {{FRCM}: A fuzzy rough c-means clustering method},
journal = {Fuzzy Sets and Systems},
volume = {480},
pages = {108860},
year = {2024},
doi = {https://doi.org/10.1016/j.fss.2024.108860},
author = {Bin Yu and Zijian Zheng and Mingjie Cai and Witold Pedrycz and Weiping Ding},
}

@ARTICLE{Ubukata2017,
  author =       {Seiki Ubukata and Akira Notsu and Katsuhiro Honda},
  title =        {General Formulation of Rough C-Means Clustering},
  journal =      {Int. J. Computer Sci. Network Security},
  year =         {2017},
  volume =       {17},
  number =       {9},
  pages =        {29-38},
  note =         {URL: \url{http://paper.ijcsns.org/07_book/201709/20170905.pdf}},
}

@article{MITRA2004,
title = {An evolutionary rough partitive clustering},
journal = {Pattern Recognition Letters},
volume = {25},
number = {12},
pages = {1439-1449},
year = {2004},
doi = {https://doi.org/10.1016/j.patrec.2004.05.007},
author = {Sushmita Mitra},
}

@ARTICLE{Mitra2006,
  author =       {S. Mitra and H. Banka and W. Pedrycz},
  title =        {Rough–Fuzzy Collaborative Clustering},
  journal =      {IEEE Transactions on Systems, Man, and Cybernetics, Part B (Cybernetics) },
  year =         {2006},
  volume =       {36},
  number =       {4},
  pages =        {795 - 805},
  doi =          {https://doi.org/10.1109/TSMCB.2005.863371},
}

@article{Cornuejols2018,
title = {Collaborative clustering: Why, when, what and how},
journal = {Information Fusion},
volume = {39},
pages = {81-95},
year = {2018},
issn = {1566-2535},
doi = {https://doi.org/10.1016/j.inffus.2017.04.008},
author = {Antoine Cornuéjols and Cédric Wemmert and Pierre Gançarski and Younès Bennani},
}

@article{Pedrycz2002,
title = {Collaborative fuzzy clustering},
journal = {Pattern Recognition Letters},
volume = {23},
number = {14},
pages = {1675-1686},
year = {2002},
issn = {0167-8655},
doi = {https://doi.org/10.1016/S0167-8655(02)00130-7},
author = {Witold Pedrycz},
}

@ARTICLE{GP10,
  AUTHOR =       {Graves, D. and Pedrycz, W.},
  TITLE =        {Kernel-based fuzzy clustering and fuzzy clustering: A comparative experimental study},
  JOURNAL =      {Fuzzy Sets and Systems},
  YEAR =         {2010},
  volume =       {161},
  number =       {4},
  pages =        {522-543},
  doi =          {https://doi.org/10.1016/j.fss.2009.10.021}
}

@ARTICLE{Zhang2016,
  author =       {Zhang, T. and Ma, F. },
  title =        {Improved rough $k$-means clustering algorithm based on weighted distance measure with {G}aussian function},
  journal =      {International Journal of Computer Mathematics},
  year =         {2016},
  volume =       {94},
  number =       {4},
  pages =        {663-675},
  doi =          {https://doi.org/10.1080/00207160.2015.1124099},
}

@ARTICLE{Singh2017,
  author =       {Singh, G. K. and Mandal, S. },
  title =        {Cluster Analysis using Rough Set Theory},
  journal =      {Journal of Informatics and Mathematical Sciences},
  year =         {2017},
  volume =       {9},
  number =       {3},
  pages =        {509-520},
  doi =          {https://doi.org/10.26713/jims.v9i3.754},
}

@Article{axioms2024,
AUTHOR = {Pérez-Ortega, Joaquín and Moreno-Calderón, Carlos Fernando and Sandra Silvia and Roblero-Aguilar and Almanza-Ortega and Nelva Nely and Frausto-Solís, Juan and Pazos-Rangel, Rodolfo and Rodríguez-Lelis, José María},
TITLE = {A New Criterion for Improving Convergence of Fuzzy C-Means Clustering},
JOURNAL = {Axioms},
VOLUME = {13},
YEAR = {2024},
NUMBER = {1},
ARTICLE-NUMBER = {35},
DOI = {https://doi.org/10.3390/axioms13010035}
}

@INPROCEEDINGS{Paul2014,
author="Paul, Sushmita
and Maji, Pradipta",
editor="Babu, B. V.
and Nagar, Atulya
and Deep, Kusum
and Pant, Millie
and Bansal, Jagdish Chand
and Ray, Kanad
and Gupta, Umesh",
title="A New Rough-Fuzzy Clustering Algorithm and its Applications",
booktitle="Proceedings of the Second International Conference on Soft Computing for Problem Solving (SocProS 2012), December 28-30, 2012",
year="2014",
publisher="Springer India",
address="New Delhi",
pages="1245--1251",
  doi =          {https://doi.org/10.1007/978-81-322-1602-5\_130},
}

@ARTICLE{Pieta2021,
  author =       {Piotr Pi\c{e}ta and Tomasz Szmuc},
  title =        {Applications of Rough Sets in Big Data Analysis: An Overview},
  journal =      {International Journal of Applied Mathematics and Computer Science},
  year =         {2021},
  volume =       {31},
  number =       {4},
  pages =        {659-683},
  doi =          {https://doi.org/10.34768/amcs-2021-0046},
}

@ARTICLE{Zhao2023,
  author =       {F. Zhao and C. Wang and H. Liu},
  title =        {Differential evolution-based transfer rough clustering algorithm},
  journal =      {Complex Intell. Syst.},
  year =         {2023},
  volume =       {9},
  number =       {},
  pages =        {5033-5047},
  doi =          {https://doi.org/10.1007/s40747-023-00987-8},
}

@ARTICLE{Zhuang2021,
  author =       {Fuzhen Zhuang and Zhiyuan Qi and Keyu Duan and Dongbo Xi and Yongchun Zhu and Hengshu Zhu and Hui Xio and Qing He},
  title =        {A Comprehensive Survey on Transfer Learning},
  journal =      {Proceedings of the IEEE},
  year =         {2021},
  volume =       {109},
  number =       {1},
  pages =        {43-76},
  doi =          {https://doi.org/10.1109/JPROC.2020.3004555},
}

@ARTICLE{Cao:2022,
  author={Cao, Bin and Zhao, Jianwei and Liu, Xin and Arabas, Jarosław and Tanveer, Mohammad and Singh, Amit Kumar and Lv, Zhihan},

  journal={IEEE Transactions on Fuzzy Systems}, 

  title={Multiobjective Evolution of the Explainable Fuzzy Rough Neural Network With Gene Expression Programming}, 

  year={2022},

  volume={30},

  number={10},

  pages={4190-4200},

  doi={10.1109/TFUZZ.2022.3141761}}

@inproceedings{GrzegorowskiJSMS:2023,
  author       = {Marek Grzegorowski and
                  Andrzej Janusz and
                  Grzegorz Sliwa and
                  Lukasz Marcinowski and
                  Andrzej Skowron},
  editor       = {Andrea Campagner and
                  Oliver Urs Lenz and
                  Shuyin Xia and
                  Dominik Slezak and
                  Jaroslaw Was and
                  JingTao Yao},
  title        = {Towards {ML} Explainability with Rough Sets, Clustering, and Dimensionality
                  Reduction},
  booktitle    = {Rough Sets - International Joint Conference, {IJCRS} 2023, Krakow,
                  Poland, October 5-8, 2023, Proceedings},
  series       = {Lecture Notes in Computer Science},
  volume       = {14481},
  pages        = {371--386},
  publisher    = {Springer},
  year         = {2023},
  doi          = {https://doi.org/10.1007/978-3-031-50959-9\_26},
}

@InProceedings{Zhu-2008,
author="Zhou, Tao
and Zhang, Yanning
and Lu, Huiling
and Deng, Fang'an
and Wang, Fengxiao",
editor="Wang, Guoyin
and Li, Tianrui
and Grzymala-Busse, Jerzy W.
and Miao, Duoqian
and Skowron, Andrzej
and Yao, Yiyu",
title="Rough Cluster Algorithm Based on Kernel Function",
booktitle="Rough Sets and Knowledge Technology",
year="2008",
publisher="Springer Berlin Heidelberg",
address="Berlin, Heidelberg",
pages="172--179",
doi={https://doi.org/10.1007/978-3-540-79721-0\_27}
}

@INBOOK{Das-2009,
  author =       {Das, S. and Abraham, A. and Konar, A.},
  title =        {Metaheuristic Clustering},
  chapter =      {5},
  pages =        {175-21},
  publisher =    {Springer, Berlin, Heidelberg},
  year =         {2009},
  volume =       {178},
  series =       {Studies in Computational Intelligence},
  doi =          {https://doi.org/10.1007/978-3-540-93964-1\_5},
}

@article{hu2010kernelized,
  title={Kernelized fuzzy rough sets and their applications},
  author={Hu, Qinghua and Yu, Daren and Pedrycz, Witold and Chen, Degang},
  journal={IEEE Transactions on Knowledge and Data Engineering},
  volume={23},
  number={11},
  pages={1649--1667},
  year={2010},
  doi={https://doi.ieeecomputersociety.org/10.1109/TKDE.2010.260},
  publisher={IEEE}
}

@article{Tripathy-2012,
author = {Tripathy, B.K. and Ghosh, Adhir and Panda, G K},
year = {2012},
month = {01},
pages = {},
title = {Kernel based K-means clustering using rough set},
journal = {2012 International Conference on Computer Communication and Informatics, ICCCI 2012},
doi = {10.1109/ICCCI.2012.6158827}
}

@book{Peres:2012,
    author = {Peters, Georg and Lingras, Pawan and Ślęzak, Dominik and Yao, Yiyu},
    title = {Rough Sets: Selected Methods and Applications in Management and Engineering},
    year = {2012},
    doi = {https://doi.org/10.1007/978-1-4471-2760-4},
    publisher = {Springer Publishing Company, Incorporated}
    }

@PROCEEDINGS{Babu2014,
  title =        {Proceedings of the Second International Conference on Soft Computing for Problem Solving (SocProS 2012), December 28-30, 2012},
  year =         {2014},
  editor =       {Babu, B.V. and Nagar, Atulya and Deep, Kusum and Pant, Millie and Bansal, Jagdish Chand and Ray, Kanad and Gupta, Umesh},
  publisher =    {Springer India},
  volume =       {236},
  number =       {},
  series =       {Advances in Intelligent Systems and Computing},
  address =      {New Delhi},
  doi =          {https://doi.org/10.1007/978-81-322-1602-5},
}

@BOOK{STC04,
  author =    {J. Shawe-Taylor and N. Cristianini},
  title =     {Kernel Methods for Pattern Analysis},
  publisher = {Cambridge University Press},
  year =      {2004},
  doi =       {https://doi.org/10.1017/CBO9780511809682}
}

@article{HU:2017,
title = {Incremental fuzzy cluster ensemble learning based on rough set theory},
journal = {Knowledge-Based Systems},
volume = {132},
pages = {144-155},
year = {2017},
issn = {0950-7051},
doi = {https://doi.org/10.1016/j.knosys.2017.06.020},
author = {Jie Hu and Tianrui Li and Chuan Luo and Hamido Fujita and Yan Yang}
}

@article{Yue:2023,
author = {Yue, Guanli and Deng, Ansheng and Qu, Yanpeng and Cui, Hui and Liu, Jiahui},
title = {Fuzzy-Rough induced spectral ensemble clustering},
year = {2023},
issue_date = {2023},
publisher = {IOS Press},
address = {NLD},
volume = {45},
number = {1},
issn = {1064-1246},
url = {https://doi.org/10.3233/JIFS-223897},
doi = {10.3233/JIFS-223897},
journal = {J. Intell. Fuzzy Syst.},
month = {jan},
pages = {1757–1774},
numpages = {18}
}

@article{WANG:201859,
title = {A spectral clustering method with semantic interpretation based on axiomatic fuzzy set theory},
journal = {Applied Soft Computing},
volume = {64},
pages = {59-74},
year = {2018},
issn = {1568-4946},
doi = {https://doi.org/10.1016/j.asoc.2017.12.004},
url = {https://www.sciencedirect.com/science/article/pii/S1568494617307184},
author = {Yuangang Wang and Xiaodong Duan and Xiaodong Liu and Cunrui Wang and Zedong Li}
}

@article{CEKIK2020113691,
title = {A novel filter feature selection method using rough set for short text data},
journal = {Expert Systems with Applications},
volume = {160},
pages = {113691},
year = {2020},
issn = {0957-4174},
doi = {https://doi.org/10.1016/j.eswa.2020.113691},
author = {Rasim Cekik and Alper Kursat Uysal}
}

@inproceedings{Sabbatini:2022,
    title     = {{Symbolic Knowledge Extraction from Opaque Machine Learning Predictors: GridREx \& PEDRO}},
    author    = {Sabbatini, Federico and Calegari, Roberta},
    booktitle = {{Proceedings of the 19th International Conference on Principles of Knowledge Representation and Reasoning}},
    pages     = {554--563},
    year      = {2022},
    month     = {8},
    doi       = {https://doi.org/10.24963/kr.2022/57},
  }

@techreport{Fischer:2005,
    author = {Fischer, I. and Poland, J.},
    title = { Amplifying the block matrix structure for spectral clustering},
    institution = {Technical Report No. IDSIA0305, Dalle Molle Institute for Artificial Intelligence, Switzerland},
    year = 2005,
    url = {https://repository.supsi.ch/5451/}
}

@ARTICLE{Su:2020,

  author={Su, Liangjun and Wang, Wuyi and Zhang, Yichong},

  journal={IEEE Transactions on Information Theory}, 

  title={Strong Consistency of Spectral Clustering for Stochastic Block Models}, 

  year={2020},

  volume={66},

  number={1},

  pages={324-338},

  doi={10.1109/TIT.2019.2934157}}

@INPROCEEDINGS{Gustafson:1978, author={Gustafson, Donald E. and Kessel, William C.}, booktitle={1978 IEEE Conference on Decision and Control including the 17th Symposium on Adaptive Processes}, title={Fuzzy clustering with a fuzzy covariance matrix}, year={1978}, volume={}, number={}, pages={761-766}, doi={10.1109/CDC.1978.268028}}

@inproceedings{Ng:2001, 
  author       = {Andrew Y. Ng and
                  Michael I. Jordan and
                  Yair Weiss},
  editor       = {Thomas G. Dietterich and
                  Suzanna Becker and
                  Zoubin Ghahramani},
  title        = {On Spectral Clustering: Analysis and an algorithm},
  booktitle    = {Advances in Neural Information Processing Systems 14 [Neural Information
                  Processing Systems: Natural and Synthetic, {NIPS} 2001, December 3-8,
                  2001, Vancouver, British Columbia, Canada]},
  pages        = {849--856},
  publisher    = {{MIT} Press},
  year         = {2001},
  url          = {https://proceedings.neurips.cc/paper/2001/hash/801272ee79cfde7fa5960571fee36b9b-Abstract.html},
}

@misc{Macgregor:2022,
      title={A Tighter Analysis of Spectral Clustering, and Beyond}, 
      author={Peter Macgregor and He Sun},
      year={2022},
      eprint={2208.01724},
      archivePrefix={arXiv},
      primaryClass={cs.DS},
      url={https://arxiv.org/abs/2208.01724}, 
}

@article{Peng:2017,
  author       = {Richard Peng and
                  He Sun and
                  Luca Zanetti},
  title        = {Partitioning Well-Clustered Graphs: Spectral Clustering Works!},
  journal      = {{SIAM} J. Comput.},
  volume       = {46},
  number       = {2},
  pages        = {710--743},
  year         = {2017},
  url          = {https://doi.org/10.1137/15M1047209},
  myurl={https://homepages.inf.ed.ac.uk/hsun4/SICOMP.pdf},
  doi          = {10.1137/15M1047209},
  }

@book{STWMAKSpringer:2018,
 author={S.T. Wierzcho\'{n} and   M.A. K{\l}opotek}
, title={Modern Algorithms of Cluster Analysis}
, pages={421}
, publisher={Springer Verlag}
, series={Studies in Big Data}
, volume=34
, year=2018
, doi={https://doi.org/10.1007/978-3-319-69308-8}
, isbn={978-3-319-69307-1}
, eisbn={978-3-319-69308-8}
}

@article{RAKMAKSTW:2020:trick
, author ={
K{\l}opotek, R. and K{\l}opotek, M.A. and Wierzcho\'{n}, S.T}
, year= 2020
, title={ A feasible k-means kernel trick under non-Euclidean feature space}
, journal={ International Journal of Applied Mathematics and Computer Science}
, volume=30
, number= 4
, pages={703--715}
, note={Online publication date: 1-Dec-2020}
, doi={https://doi.org/10.34768/amcs-2020-0052}
}

@article{Plosone2025,
    title={Explainable Graph Spectral Clustering of Text Documents},
    author={Bart{\l}omiej Starosta and 
            Mieczys{\l}aw A. K{\l}opotek  and 
            S{\l}awomir T. Wierzcho{\'n} and
            Dariusz Czerski and 
            Marcin Sydow and  
            Piotr Borkowski},
    year={2025},
    note={https://journals.plos.org/plosone/article?id=10.1371/journal.pone.0313238},
journal={PLoS One}, 
month={February},
date={4},
volume={20(2):e0313238},
doi= {10.1371/journal.pone.0313238},
xnote={  eCollection 2025} 
}

@incollection{RozdzialvMonografia:2025, 
    title={Approaches to {Explain}ability of Output of Graph Spectral Clustering Methods},
    author={Bart{\l}omiej Starosta and 
            Mieczys{\l}aw A. K{\l}opotek  and 
            S{\l}awomir T. Wierzcho{\'n} },
    year={2025},
publisher={University of Siedlce},
booktitle={Design and Implementation of Artificial Intelligence Systems},
series={Intelligent Systems and Information Technology},
editor={Dariusz Mikułowski and Artur Niewiadomski},
isbn={ISBN 978-83-68355-16-1},
pages={5--31},
    note={}
}

@article{MAK:2019:FI:Gower ,
    author = {K\l{}opotek, Mieczys\l{}aw A.},
    title = {On the Existence of Kernel Function for Kernel-Trick of $k$-Means in the Light of {G}ower Theorem},
    year = {2019},
    issue_date = {2019},
    publisher = {IOS Press},
    address = {NLD},
    volume = {168},
    number = {1},
    issn = {0169-2968},
    doi = {https://doi.org/10.3233/FI-2019-1822},
    journal = {Fundam. Inf.},
    month = {jan},
    pages = {25–43},
    numpages = {19}
    }





\end{document}